\documentclass[pdflatex,sn-mathphys]{sn-jnl}

\usepackage{float}
\usepackage[frozencache,cachedir=minted-cache]{minted} 

\usepackage{url}
\usepackage{hyperref}

\jyear{2025}%

\theoremstyle{thmstyleone}%
\theoremstyle{thmstyletwo}%

\theoremstyle{thmstylethree}%

\begin{document}

\title[LLM-guided automated algorithmic discovery]{Automated Algorithmic Discovery for Scientific Computing through LLM-Guided Evolutionary Search: A Case Study in Gravitational-Wave Detection}

\author*[1,2]{\fnm{He} \sur{Wang}}\email{hewang@ucas.ac.cn}
\author*[3]{\fnm{Liang} \sur{Zeng}}\email{zengliangcs@gmail.com}

\affil[1]{\orgdiv{International Centre for Theoretical Physics Asia-Pacific}, \orgname{University of Chinese Academy of Sciences}, \orgaddress{\postcode{100190}, \state{Beijing}, \country{China}}}
\affil[2]{\orgdiv{Taiji Laboratory for Gravitational Wave Universe}, \orgname{University of Chinese Academy of Sciences}, \orgaddress{\postcode{100049}, \state{Beijing}, \country{China}}}
\affil[3]{\orgname{Tsinghua University}, \orgaddress{\postcode{100084}, \state{Beijing}, \country{China}}}


\abstract{
Automated algorithm discovery in scientific computing faces fundamental challenges: vast design spaces with expensive evaluations, domain-specific physical constraints requiring expert knowledge, and the necessity for interpretable solutions that scientists can validate and understand.
We present the Evo-MCTS (Evolutionary Monte Carlo Tree Search) framework, integrating large language models (LLMs) with tree-structured evolutionary search for interpretable algorithm discovery. 
Evo-MCTS combines reflective code synthesis leveraging LLM domain knowledge, multi-scale evolutionary operations on structured code representations, and interpretable algorithmic pathways emerging from tree-guided exploration. 
When applied to gravitational wave detection—a challenging domain with continuous parameter spaces and strict physical constraints—Evo-MCTS achieves 20.2\% improvement over domain-specific methods and 59.1\% over LLM-based optimization frameworks.
This improvement arises from its ability to consistently converge toward interpretable algorithmic structures that integrate multiple functional components.
Our domain-agnostic architecture establishes a generalizable methodology for automated algorithm discovery in scientific computing, where algorithmic transparency and physical validity are as essential as performance optimization.
}
\keywords{Large Language Models, Monte Carlo Tree Search, Automated Algorithmic Discovery, Gravitational Wave Detection, Scientific Computing}



\maketitle

\section{Introduction}\label{sec:intro}

The pursuit of scientific discovery increasingly demands computational approaches that can navigate complex, high-dimensional data spaces while preserving physical interpretability~\cite{zheng2023Large,2023WangScientificdiscoveryage,karniadakis2021physics,baker2019workshop}. Across diverse scientific domains—from molecular dynamics to astrophysical signal detection—algorithm design represents a critical bottleneck where manual engineering approaches struggle to navigate vast combinatorial spaces of possible computational strategies while maintaining domain validity. Gravitational wave astronomy exemplifies this computational challenge: detection systems must identify faint astrophysical signals buried in detector noise while leveraging theoretical predictions from general relativity~\cite{abbott2016observation,abbott2019gwtc1}, yet remain adaptable to unexpected signal morphologies that could reveal new physics beyond current theoretical models.

Despite the impressive progress in gravitational wave detection~\cite{abbott2020guide}, current approaches face core limitations stemming from restrictive assumptions. Matched filtering techniques~\cite{owen1996search,cutler1994gravitational} exhibit optimal sensitivity under Gaussian stationary noise but fundamentally depend on accurate prior knowledge of signal morphologies. Non-template methods~\cite{klimenko2016method} attempt model-independent detection, but typically accept decreased sensitivity in exchange for generality. More recently, deep neural network approaches~\cite{george2018deep,gabbard2018matching,2021HuertaAccelerated} bring computational efficiency, yet operate as black-box systems that conceal decision-making logic and introduce potential hidden biases~\cite{Nagarajan_2025}. These strategies all manifest trade-offs between sensitivity, generality, and interpretability; thus far, none fully resolve the challenge of discovering innovative detection algorithms beyond traditional boundaries.

While advances in computational power and data availability continue to accelerate scientific progress, the design of effective algorithms has become the central bottleneck in modern scientific computing~\cite{baker2019workshop,karniadakis2021physics}. Current gravitational wave detection pipelines represent decades of manual engineering yet still miss potentially discoverable signals due to three critical limitations: (i) \textit{combinatorial explosion} of possible signal processing combinations that exceeds human exploration capacity~\cite{elsken2019neural,eiben2015evolutionary}, (ii) \textit{cognitive biases} that constrain designers to familiar paradigms~\cite{kahneman2011thinking}, and (iii) \textit{local optimization traps} where manual refinement leads to incremental improvements while missing global optima~\cite{wolpert2002no}. These factors highlight the need for approaches that can automate the systematic exploration of algorithm design spaces without sacrificing scientific rigor.

Automated algorithm discovery in scientific domains confronts three intertwined demands: physical constraint adherence, efficient exploration of vast design spaces, and interpretable solutions that remain scientifically transparent. Meeting these requirements in unison is essential for credible scientific progress. Large Language Models (LLMs) embed domain knowledge and physical constraints directly into the generative process~\cite{chen2021evaluating,lewkowycz2022solving}, while Monte Carlo Tree Search (MCTS) drives efficient, structured navigation through the algorithmic landscape~\cite{browne2012Survey,wang2020alphax,li2025hunyuanprover}. Their integration enables an evolutionary search process that systematically uncovers interpretable, high-performing, and physically valid scientific algorithms.

Existing approaches to algorithm discovery span multiple paradigms. Traditional approaches including genetic programming~\cite{koza1994Genetic}, neural architecture search~\cite{elsken2019neural}, and evolutionary computation~\cite{eiben2015evolutionary} suffer from critical limitations: generating syntactically invalid code, lacking domain knowledge integration, or focusing on network topology rather than algorithmic logic. Recent advances have integrated LLMs with structured search and evolutionary methods, including FunSearch~\cite{romera2023mathematical}, EoH~\cite{EoH}, AEL~\cite{AEL}, ReEvo~\cite{ReEvo}, and MCTS-AHD~\cite{MCTS_AHD} for algorithm discovery, which combine LLMs' code generation capabilities with systematic optimization frameworks. However, the existing LLM-based frameworks focus on combinatorial optimization tasks with discrete decision sequences and well-defined mathematical formulations. Scientific signal processing problems like gravitational wave detection present fundamentally different challenges, requiring continuous parameter optimization, domain-specific physical constraints, and interpretable algorithmic pathways that can be validated against theoretical predictions.

We propose Evo-MCTS (Evolutionary Monte Carlo Tree Search), a framework that realizes synergistic integration through three key innovations: (i) \textit{Reflective Code Synthesis} that leverages LLM domain knowledge for performance-driven algorithm generation, adapting to optimization landscapes while maintaining scientific validity, (ii) \textit{Multi-Scale Evolutionary Operations} (Parent/Sibling/Path-wise Crossover, Point Mutation) that operate on structured code representations through MCTS tree traversal, enabling semantic-aware algorithmic transformations, and (iii) \textit{Interpretable Algorithm Pathways} that emerge naturally from the tree structure, enabling post-hoc analysis of algorithmic evolution while providing cumulative learning for future discoveries. The framework addresses interdependent challenges through architectural design—each element enhances others' capabilities while compensating for individual limitations.

We demonstrate the effectiveness of Evo-MCTS through application to gravitational wave detection—a challenging scientific computing domain with continuous parameter spaces and strict physical constraints. Comprehensive evaluation shows the framework achieves 20.2\% performance improvement over domain-specific state-of-the-art algorithms and a remarkable 59.1\% improvement over other LLM-based algorithm optimization frameworks, validating the methodological innovations through systematic algorithmic exploration. The framework demonstrates consistent breakthrough discoveries across independent executions, generating human-interpretable algorithmic pathways that reveal distinct evolutionary patterns organized by functional categories. These interpretable pathways illustrate the framework's capability to navigate complex algorithmic design spaces while maintaining scientific validity and transparency, identifying multi-component integration strategies through performance-driven optimization.

While this study focuses on gravitational wave detection as a challenging testbed, Evo-MCTS offers a transferable framework for interpretable algorithmic discovery that generalizes to broader domains of scientific computing. Its capacity to generate interpretable and high-performing algorithmic pathways makes it especially valuable for scientific applications where understanding algorithmic reasoning is as important as achieving optimal performance. Our approach opens new avenues for automated scientific discovery across physics, chemistry, biology, and engineering, providing a principled methodology that respects domain-specific constraints while systematically exploring algorithmic possibilities beyond human intuition.

\section{Results}

\subsection{Framework Design and Implementation}

\begin{figure*}
\centering
\includegraphics[width=\textwidth]{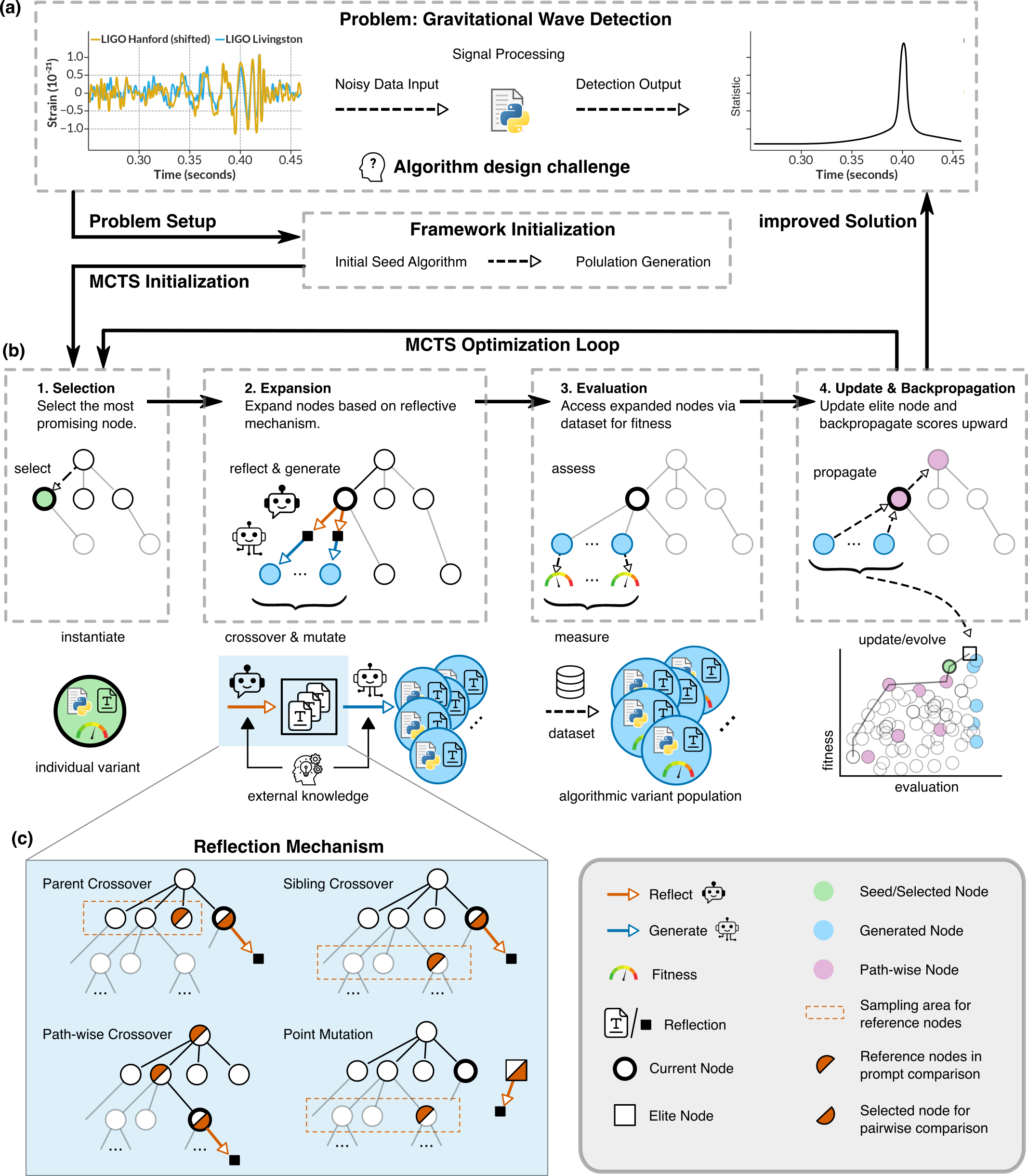}
\caption{\textbf{LLM-Guided Evolutionary Monte Carlo Tree Search Framework for Automated Algorithm Discovery.} \textbf{(a)} Overview of the algorithm discovery pipeline. Starting from raw gravitational wave strain data (left), the framework applies automated algorithmic transformations through LLM-generated code synthesis (center) to produce optimized detection statistics (right). \textbf{(b)} Core architectural components showing the integration of MCTS exploration with evolutionary optimization through dual perspectives of tree search and population evolution. \textbf{(b.1)} UCT-based node selection from initial algorithmic variants including seed algorithms and individual variants, each represented as nodes containing baseline signal processing code. \textbf{(b.2)} MCTS expansion phase where new algorithmic variants are generated through evolutionary operations. Each node contains executable Python code implementing specific detection strategies. \textbf{(b.3)} Algorithm evaluation phase where generated variants are tested against benchmark data to compute fitness scores, determining performance-based selection for subsequent iterations. \textbf{(b.4)} MCTS backpropagation and elite node updates after multiple evolutionary cycles, propagating performance feedback through the tree structure and maintaining diverse high-performing detection strategies. \textbf{(c)} Detailed view of the reflection mechanism during MCTS expansion, showing four evolutionary operations: Parent Crossover, Sibling Crossover, Path-wise Crossover, and Point Mutation.
}
\label{fig:framework}
\end{figure*}

\textbf{Framework Overview.} The Evo-MCTS framework systematically transforms raw time-series data into a comprehensive catalog of optimized detection algorithms through an automated discovery pipeline that integrates domain knowledge encoded in LLMs (Figure~\ref{fig:framework}a, see Methods Section \ref{sec:problem_formulation} for formal problem definition).
To understand this automated transformation process, the framework can be conceptualized through two complementary perspectives (Figure~\ref{fig:framework}b): as an MCTS-guided tree search where nodes represent complete algorithmic implementations and edges encode LLM-driven transformations (detailed implementation in Methods Section \ref{sec:llm_integration}), or as an evolutionary algorithm where populations of algorithms undergo sophisticated selection, crossover, and mutation operations guided by domain knowledge (detailed implementation in Methods Section \ref{sec:evolutionary_ops}).

\textbf{Evolutionary Search Operations.} The core innovation of our Evo-MCTS framework lies in reformulating algorithm design as a tree search problem, where each node represents an executable algorithm and edges correspond to code transformations (Figure~\ref{fig:framework}c). Starting from a seed algorithm, the framework employs four specialized evolutionary operations to expand the search tree:
\begin{itemize}
    \item \textit{Parent Crossover (PC):} Combines algorithmic features from parent nodes to generate offspring that inherit successful detection strategies while exploring new combinations.
    \item \textit{Sibling Crossover (SC):} Enables horizontal knowledge transfer between algorithms at the same tree depth, promoting diversity while maintaining comparable complexity levels.
    \item \textit{Path-wise Crossover (PWC):} Synthesizes information across complete root-to-leaf trajectories, capturing long-range dependencies and enabling global optimization strategies.
    \item \textit{Point Mutation (PM):} Introduces targeted modifications to individual algorithms based on performance analysis, leveraging insights from elite algorithms to enable fine-grained optimization of specific components.
    \end{itemize}
These operations are fundamentally different from traditional genetic algorithms because they operate on structured code representations rather than abstract encodings, and modifications are guided by LLM-based reasoning rather than random perturbations (see Methods Section \ref{sec:evolutionary_ops} and Supplementary Information Section S1 for operation details).

\begin{figure*}
    \centering
    \includegraphics[width=\textwidth]{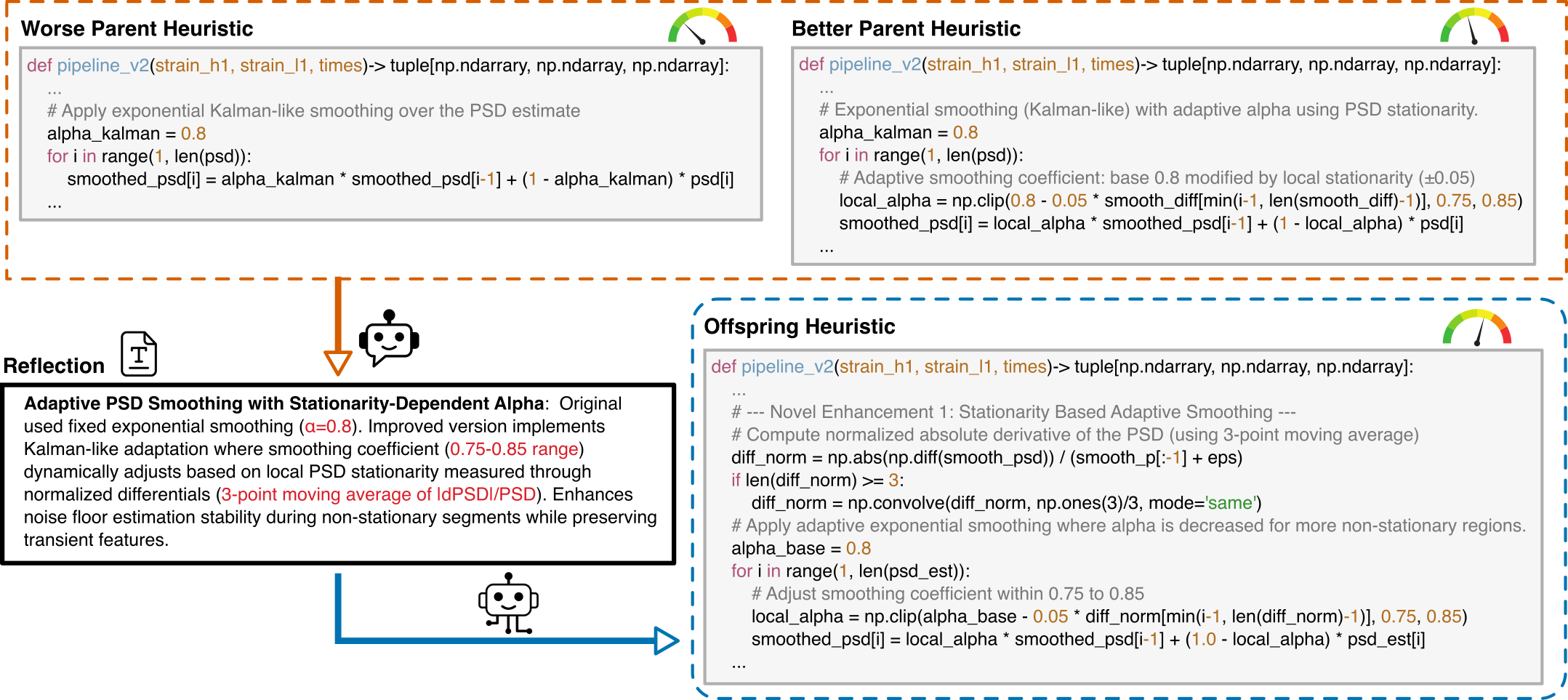}
    \caption{\textbf{LLM-Driven Algorithmic Evolution Through Reflective Code Synthesis.} Demonstration of a single Parent Crossover evolutionary step showing the transformation from two parent algorithms to an enhanced offspring algorithm. \textbf{(Top row)} Code segments from two parent nodes highlighting complementary algorithmic components that will be combined through the crossover operation. \textbf{(Bottom left, black box)} Reflective analysis process showing how the LLM identifies strengths and limitations in the parent algorithms, synthesizing insights about their respective detection strategies and potential synergies. \textbf{(Bottom right)} Generated offspring algorithm code incorporating successful elements from both parents while addressing identified limitations through domain-aware synthesis. This example illustrates the framework's capability to generate physically-motivated algorithmic improvements through automated reasoning, demonstrating how LLM-guided reflection enables discovery of sophisticated signal processing techniques by combining and enhancing existing algorithmic components without manual intervention. The complete reflection prompts and additional evolution examples are provided in Supplementary Information Section S1.}
    \label{fig:reflection_example}
\end{figure*}

\textbf{Reflective Code Generation.} Central to the Evo-MCTS framework's effectiveness is the reflection mechanism that analyzes algorithm performance patterns and guides subsequent explorations (Figure~\ref{fig:framework}b-(2) and Figure~\ref{fig:framework}c). This dual-component system comprises: (i) a performance reflection module that identifies strengths and weaknesses in current algorithms through systematic evaluation across diverse signal conditions, and (ii) a code synthesis module that leverages these insights to generate improved implementations (An example is shown in Figure~\ref{fig:reflection_example}). The reflection process operates at multiple scales—from individual algorithmic components to complete detection pipelines—ensuring both local optimization and global coherence (detailed prompts and examples in Supplementary Information Section S1).

\subsection{Optimization Dynamics and Benchmark Comparison}

We evaluated our Evo-MCTS framework's algorithmic discovery capabilities using the Machine Learning Gravitational-Wave Search Mock Data Challenge (MLGWSC-1) benchmark dataset~\cite{Marlin2023First}—a standardized evaluation environment that provides a controlled testbed for systematic framework validation. The MLGWSC-1 benchmark offers a challenging application domain that combines realistic detector noise characteristics with diverse signal morphologies, enabling demonstration of the framework's effectiveness on scientific computing problems requiring sophisticated signal processing under domain-specific physical constraints. 

\begin{figure*}
\centering
\includegraphics[width=\textwidth]{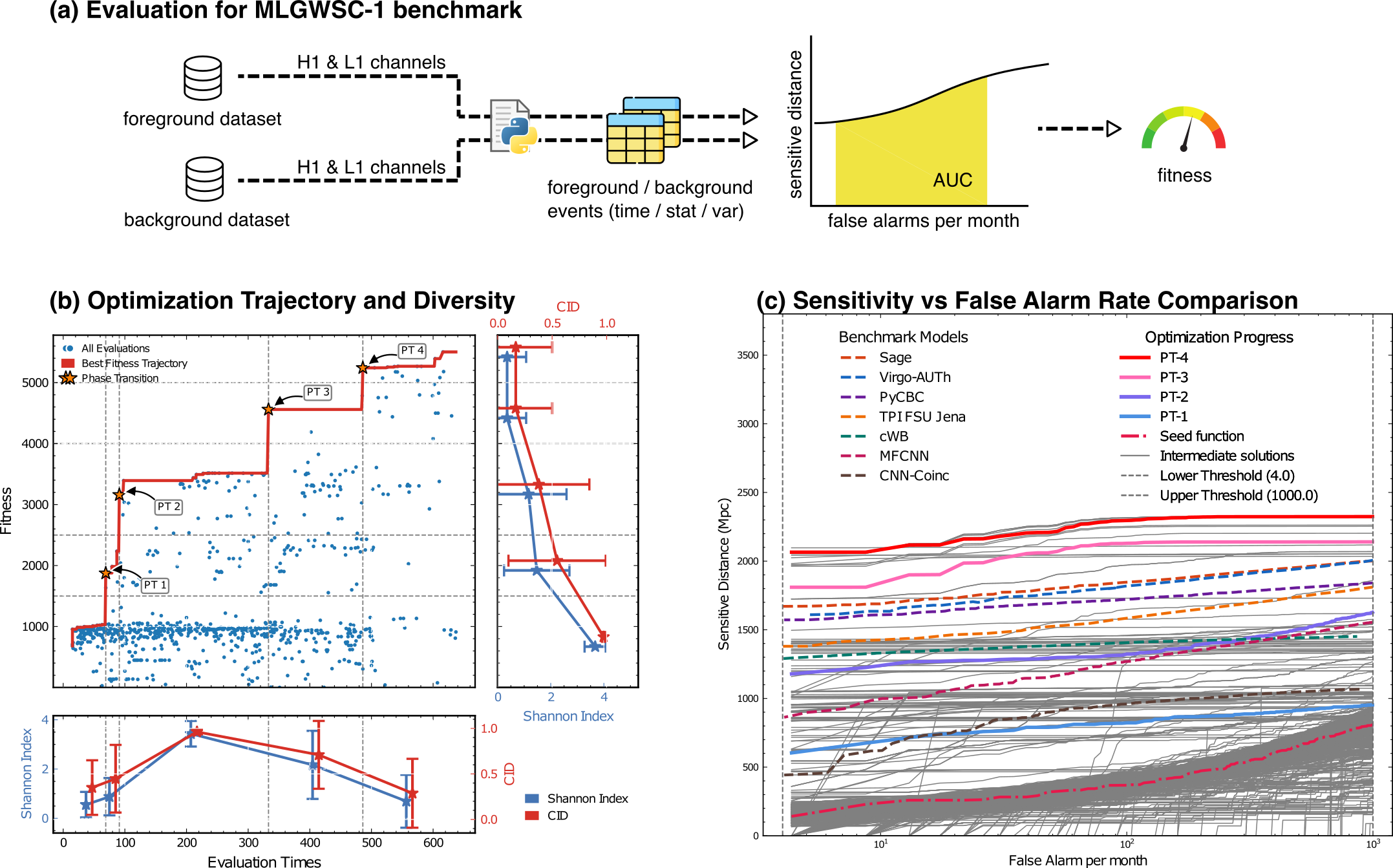}
\caption{\textbf{Framework Optimization Dynamics and Performance Validation on MLGWSC-1 Benchmark.} \textbf{(a)} Evo-MCTS framework adaptation pipeline demonstrating integration with domain-specific evaluation protocols through standardized fitness assessment. The pipeline processes input data through evolved algorithms to produce outputs evaluated against ground truth labels using area under the curve (AUC) metrics. \textbf{(b)} Framework optimization trajectory and diversity analysis across 877 evaluations from 5 independent runs. Combined fitness trajectory (blue dots) with best objective envelope (red line) showing four phase transitions (PT 1-4, orange stars) marking algorithmic breakthroughs with fitness gains $\ge 400$ units. Maximum fitness of 5,241 units achieved, representing a 6-fold improvement from baseline. Diversity metrics include Shannon diversity index (blue, left axis) and Complexity Index of Diversity (CID, red, right axis) with error bars showing standard deviation across runs. Right panel shows fitness-stratified diversity analysis revealing systematic exploration patterns across performance levels. \textbf{(c)} Performance comparison on MLGWSC-1, Set 4 dataset showing framework validation against seven benchmark algorithms (Sage, Virgo-AUTh, PyCBC, TPI FSU Jena, cWB, MFCNN, CNN-Coinc). Optimization milestones PT-1 through PT-4 demonstrate progressive algorithmic refinement, with PT-4 achieving 20.2\% improvement over SOTA baselines. Grey curves represent intermediate solutions explored during optimization, while the red dotted line shows seed function baseline. Vertical dashed lines indicate evaluation range boundaries (4-1000 events per month). Results validate the framework's systematic exploration capabilities, interpretable algorithmic pathways, and effective convergence toward high-performing solutions through progressive complexity enhancement and multi-component integration strategies.}
\label{fig:mlgwsc_benchmark}
\end{figure*}

\textbf{Performance Evaluation Protocol.} Figure~\ref{fig:mlgwsc_benchmark}a illustrates the Evo-MCTS framework's adaptation to the MLGWSC-1 evaluation protocol, demonstrating the framework's capability to integrate with domain-specific assessment procedures. The system transforms raw dual-channel strain data through evolved algorithms, producing detection catalogs evaluated against ground truth labels. The evaluation pipeline employs the area under the curve (AUC) metric as the performance indicator, computed from sensitivity distance versus false alarm rate curves spanning 4-1000 events per month. This metric provides a comprehensive performance measure that balances multiple detection objectives, enabling systematic framework validation through standardized benchmark assessment (detailed experimental configuration provided in Methods Section \ref{sec:experimental_setup}).

\textbf{Evolutionary Optimization Trajectory.} Figure~\ref{fig:mlgwsc_benchmark}b presents the comprehensive optimization trajectory across 877 total evaluations from five independent Evo-MCTS runs, revealing systematic algorithmic discovery with progressive increases in both algorithmic sophistication and detection performance. The optimization process exhibits four distinct phase transitions (PT 1-4), each marking algorithmic breakthroughs that represent the discovery of increasingly sophisticated algorithmic components building upon previous innovations. The combined fitness trajectory demonstrates the framework's capability to navigate the complex algorithmic design space while maintaining consistent improvement patterns across multiple independent runs, systematically exploring and integrating complex detection strategies (detailed analysis of algorithmic evolution patterns provided in Supplementary Information Section S3).

\textbf{Population Diversity Analysis.} The diversity analysis reveals sophisticated exploration patterns that balance algorithmic variety with performance optimization throughout the search process. Peak diversity occurs during the intermediate optimization phase between PT 2 and PT 3, where both Shannon diversity index and Complexity Index of Diversity reach maximum values, indicating the framework maintains diversity in both component selection and structural complexity while systematically exploring combinations of successful components before converging on optimal configurations. Fitness-stratified analysis demonstrates systematic convergence: diversity decreases progressively from broad initial exploration in lower-performing variants to focused refinement in highest-performing configurations, confirming effective balance between exploration and exploitation (diversity metric definitions and computational details provided in Methods Section \ref{sec:experimental_setup}).

\textbf{Experimental Comparison.} Figure~\ref{fig:mlgwsc_benchmark}c compares the Evo-MCTS framework's performance against seven algorithms on MLGWSC-1, Set 4 dataset, demonstrating the framework's effectiveness in navigating complex algorithmic design spaces. The PT-4 configuration achieves 20.2\% performance improvement over competing approaches including Sage~\cite{Nagarajan_2025}, Virgo-AUTh~\cite{Nousi_2023,Marlin2023First}, PyCBC~\cite{Nitz_2023,Marlin2023First}, TPI FSU Jena~\cite{Zelenka_2024,Marlin2023First}, cWB~\cite{DRAGO2021100678,Marlin2023First}, MFCNN~\cite{Gravitational2020Wang,Marlin2023First}, and CNN-Coinc~\cite{Schafer_2022a,Schafer_2022b,Marlin2023First}. Progressive algorithmic sophistication is demonstrated through four milestone configurations (PT-1 through PT-4) across the false alarm rate range of 4-1000 events per month, validating the framework's systematic exploration and refinement capabilities (detailed performance metrics provided in Supplementary Information Section S3).

\textbf{Algorithmic Evolution.} The systematic improvement from PT-1 to PT-4 demonstrates the framework's capability to discover and integrate sophisticated algorithmic components through interpretable evolutionary pathways. The evolved algorithms incorporate nonlinear transformations and adaptive processing strategies that respond to the benchmark's realistic noise characteristics~\cite{abbott2016observation,dal2021gravitational}, illustrating how LLM-guided evolutionary search can systematically explore complex algorithmic design spaces under domain-specific constraints.

The framework's performance relative to existing approaches validates key methodological capabilities: compared to template-based methods like PyCBC~\cite{Nitz_2023} that optimize for specific theoretical assumptions~\cite{finn1992detection,cutler1994gravitational}, our framework demonstrates automated exploration of alternative algorithmic paradigms. Relative to template-free approaches like coherent WaveBurst (cWB)~\cite{klimenko2016method,Klimenko_2016}, the framework shows how systematic LLM-guided search can navigate algorithmic spaces beyond manual heuristic design~\cite{EoH,zhang2024understanding}. Compared to deep learning methods with millions of opaque parameters~\cite{lecun2015deep}, the evolved algorithms maintain interpretability through explicit mathematical formulations~\cite{rudin2019stop,molnar2020interpretable}. These results confirm that automated discovery can achieve competitive performance while preserving transparency, which is essential for scientific applications where algorithmic reasoning must be validated by domain experts (detailed quantitative analysis and algorithm specifications are provided in Supplementary Information Section S3).

\subsection{Generalization Study and Component Analysis}

\begin{figure*}
    \centering
    \includegraphics[width=\textwidth]{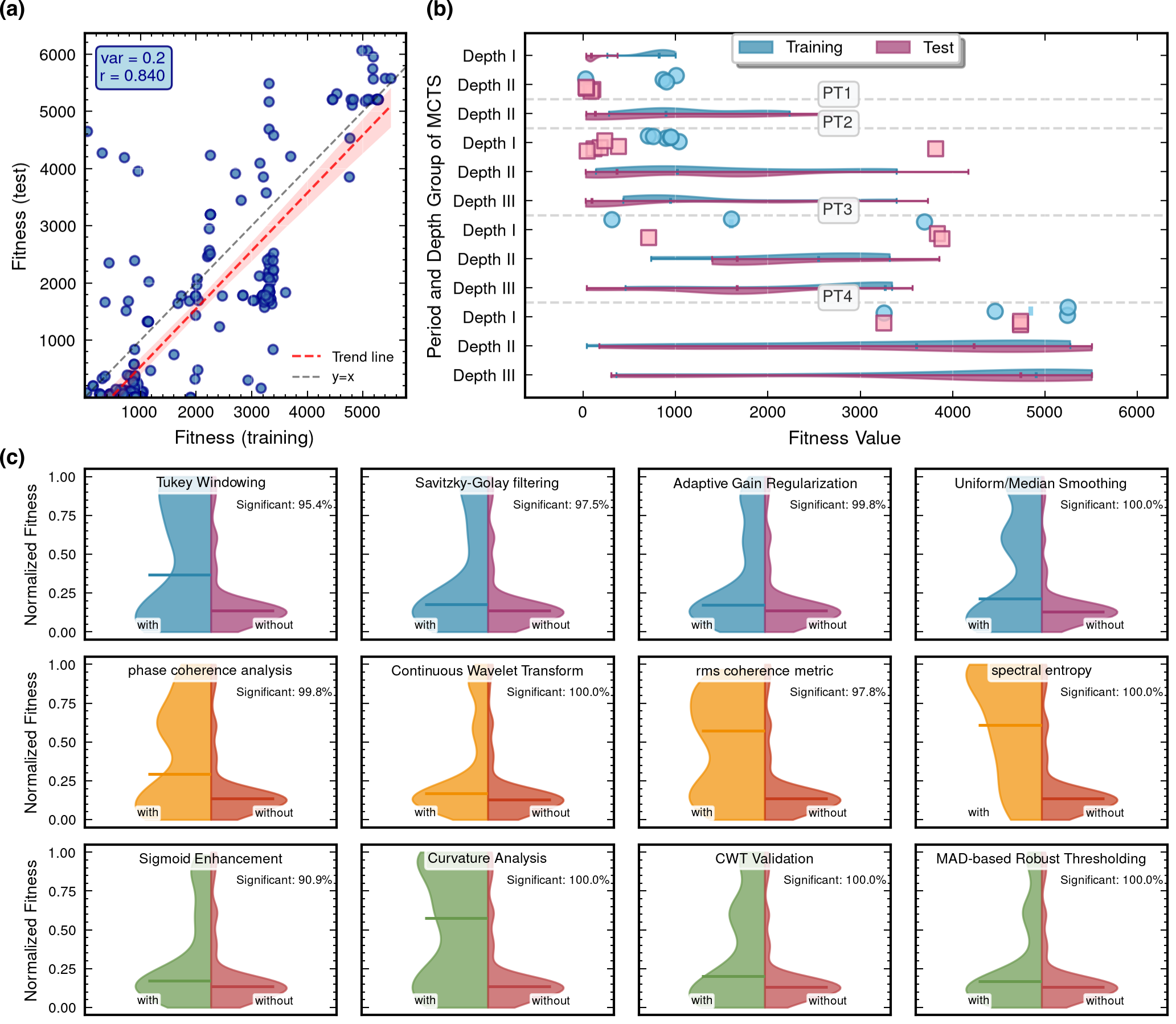}
    \caption{\textbf{Generalization Validation and Component Effectiveness Analysis.} \textbf{(a)} Training versus test performance correlation for 877 algorithmic configurations evaluated under 0.2-second trigger arrival time uncertainty constraint. Each point represents an individual algorithm's fitness scores (computed from AUC metrics) on training (7-day dataset) and test (1-day independent dataset) data. Linear correlation coefficient r = 0.840 indicates strong generalization capability, while variance reflects expected performance variation due to non-stationary, non-Gaussian noise characteristics. Red dashed line shows the empirical trend relationship, while the grey dashed line represents perfect correlation (y=x). High-performing algorithms (fitness $>$ 4000) demonstrate particularly robust generalization across different noise realizations and signal parameters.
    \textbf{(b)} MCTS depth-stratified performance analysis across optimization phases. Fitness distribution of algorithms organized by MCTS tree depth groups (Depth I: depths 1-4, Depth II: depths 5-7, Depth III: depths 8-10) and phase transitions (PT1-PT4). Training performance (teal) and test performance (pink) are shown with violin plots for sample sizes $n \ge 10$ and scatter plots (circles/rectangles) for $n < 10$. The analysis reveals systematic migration of high-fitness algorithms toward deeper tree layers as optimization progresses, with elite algorithms (fitness $>$ 5,000) emerging exclusively in deeper layers during PT4. Enhanced generalization capability is observed in deeper layers during later optimization phases, as evidenced by improved training-test performance alignment in Depth III compared to shallower depth groups.
    \textbf{(c)} Algorithmic component impact analysis. Violin plots comparing normalized fitness distributions between algorithms with specific techniques (left) versus without (right). Techniques categorized as conditioning methods (teal), time-frequency analysis (orange), and trigger detection (green). Technique effectiveness is determined by distributional separation: wider gaps between left and right distributions indicate stronger performance impact. Conditioning techniques (Savitzky-Golay filtering, Adaptive Gain Regularization) and trigger detection methods (Curvature Analysis, Continuous Wavelet Transform Validation) demonstrate the most substantial improvements through clear distributional shifts toward higher fitness values. Statistical validation across 1,000 resampling iterations confirms significance ($p < 0.001$) and practical importance.
    }
    \label{fig:interpretability1}
\end{figure*}

\textbf{Training-Test Correlation.} We evaluated 877 algorithmic configurations on an independent 1-day test dataset, distinct from the 7-day training corpus, under a 0.2-second trigger arrival time uncertainty constraint to assess robustness and interpretability (data specifications are provided in Supplementary Information Section S2).

Figure~\ref{fig:interpretability1}a demonstrates strong training-test performance correlation ($r = 0.840$) across all algorithms, with each point representing fitness scores (computed from AUC metrics) for individual configurations. This correlation validates robust generalization despite significant domain shift between datasets, with performance variation reflecting the non-stationary, non-Gaussian noise characteristics inherent in realistic gravitational wave detection environments. The 0.2-second timing constraint ensures temporal precision essential for astrophysical parameter estimation while preserving detection sensitivity, empirically determined through systematic constraint-correlation analysis (detailed in Supplementary Information Section S4). These results confirm that optimized algorithms achieve genuine performance improvements transferable to independent datasets, validating our evolutionary framework's practical utility for real-world gravitational wave detection applications.

\textbf{Depth-Stratified Analysis.} We analyzed the relationship between MCTS tree depth and algorithm fitness across optimization phases. 
Figure~\ref{fig:interpretability1}b reveals systematic evolution patterns in performance and generalization capability.

The analysis demonstrates clear progression in algorithmic quality through successive phase transitions. During PT1, algorithms are predominantly in shallow depth groups with modest fitness values ($<$ 2,000). As optimization progresses to PT2 and PT3, high-performing algorithms (fitness $>$ 3,000) increasingly emerge in deeper tree layers, indicating successful identification and refinement of promising algorithmic directions.

Elite algorithms (fitness $>$ 5,000) emerge exclusively in deeper tree layers during PT4, suggesting that sophisticated solutions require extensive refinement through multiple decision levels. Training (teal) and test (pink) performance distributions show robust generalization across all depth groups, with algorithms maintaining relative performance rankings regardless of tree depth.

High-performing algorithms become increasingly rare but more consistent in deeper layers, reflecting natural convergence toward superior solutions. Critically, deeper layers exhibit superior generalization capability during later optimization phases, with training-test performance gaps narrowing significantly in Depth III compared to shallower groups. This improved generalization suggests extensive algorithmic refinement enhances both performance and robustness across different observational conditions, confirming the MCTS framework effectively balances exploration breadth with exploitation depth.

\textbf{Technique Impact Analysis.} We conducted comprehensive technique impact analysis using controlled comparative methodology, systematically evaluating algorithms with specific signal processing techniques against matched controls (details provided in Supplementary Information Section S5).

Figure~\ref{fig:interpretability1}c reveals distinct performance impacts across algorithmic components. 
Conditioning techniques demonstrate the most pronounced positive effects, with Savitzky-Golay filtering showing clear distributional separation and asymmetric violin plots shifted toward higher fitness values. This technique has been applied in gravitational wave counterpart studies for smoothing two-dimensional dispersed images from the Hubble Space Telescope, effectively removing high-frequency structure while preserving underlying emission patterns~\cite{troja2017Xray}.
These findings establish quantitative benchmarks for algorithmic component selection in automated gravitational wave detection systems.

\subsection{Evolutionary Pathways and Reproducibility}

\begin{figure*}
    \centering
    \includegraphics[width=0.9\textwidth]{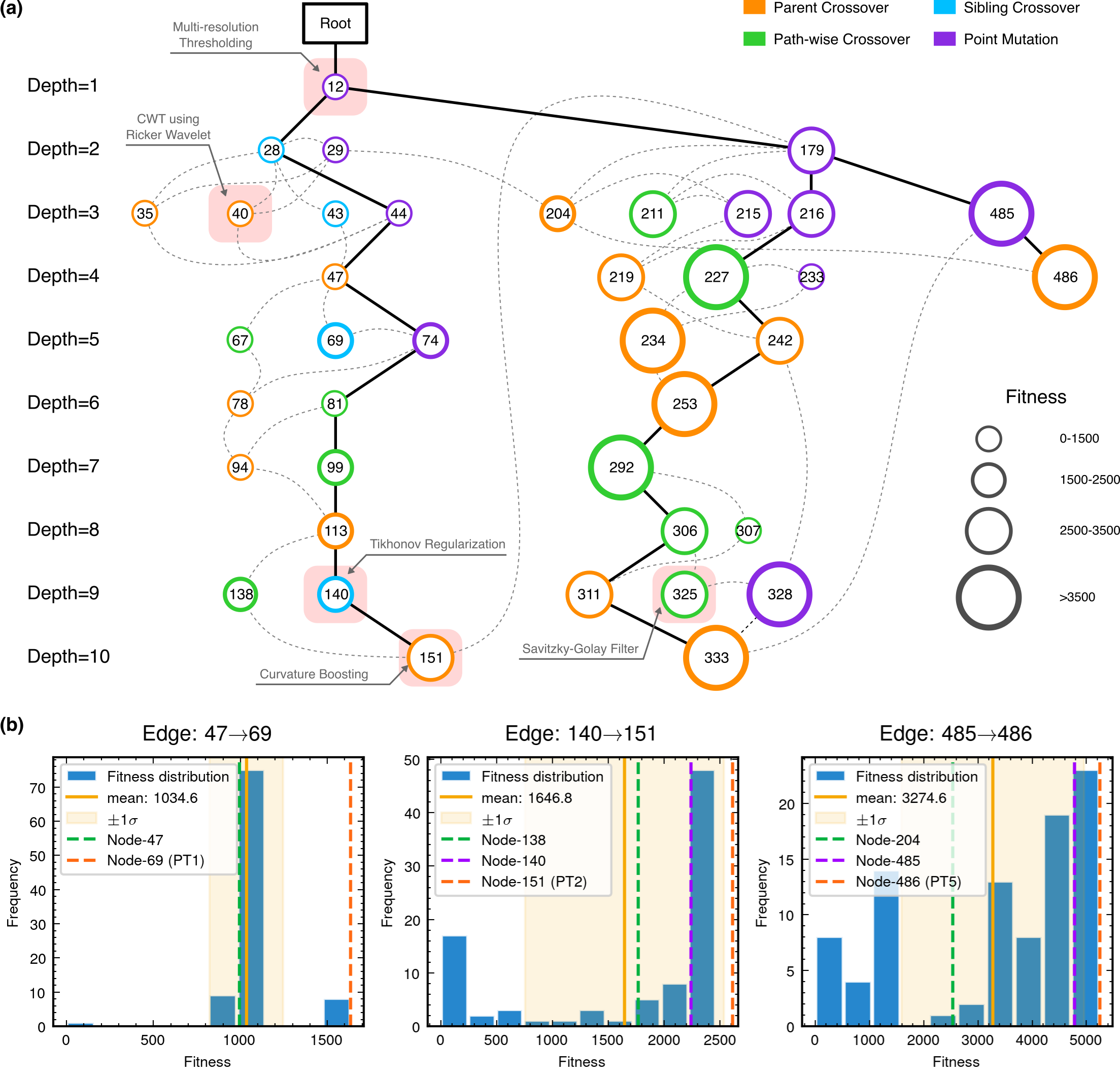}
    \caption{\textbf{Algorithmic Evolutionary Pathway of MCTS and Edge Robustness Analysis.} 
    \textbf{(a)} Complete MCTS tree structure showing all nodes associated with the PT4 algorithm (node 486, fitness=5241.4) discovered in an optimization run. Node sizes encode fitness values (larger circles = higher performance), with evaluation times displayed inside circles. Node colors indicate expansion operation types: Parent Crossover (orange), Sibling Crossover (cyan), Path-wise Crossover (green), and Point Mutation (purple). Solid black lines represent the selected MCTS exploration path, while dashed gray lines indicate nodes referenced in expansion prompts for knowledge synthesis. Five key algorithmic breakthroughs are annotated: Multi-resolution Thresholding (first appearing at node 12), CWT using Ricker Wavelet (node 28), Tikhonov Regularization (node 140), Curvature Boosting (node 151), and Savitzky-Golay Filter (node 333). These techniques propagate through subsequent generations, demonstrating systematic knowledge accumulation and refinement. The tree visualization reveals how sophisticated detection algorithms emerge through progressive technique integration across multiple MCTS depth levels. 
    \textbf{(b)} Edge robustness analysis for three critical evolutionary transitions. Each subplot shows fitness distributions from 100 independent re-executions of specific edges: Edge 47→69 (early breakthrough, mean fitness 1034.6, 89.25\% variants exceeding preceding node performance), Edge 140→151 (intermediate advancement, mean fitness 1646.8, 52.81\% achieving superior fitness with 100\% regularization technique inheritance), and Edge 485→486 (final optimization stage, mean fitness 3274.6, 70.65\% variants outperforming node 204, 25.00\% surpassing node 485). Vertical reference lines indicate the original node fitness values and key ancestral nodes. The distributions demonstrate the stochastic nature of LLM-driven code generation while confirming consistent discovery of high-performance algorithmic variants with robust knowledge transfer across independent executions.}
\label{fig:interpretability2}
\end{figure*}

\textbf{Knowledge Accumulation Pathways.} To understand the mechanistic basis of algorithmic discovery, we conducted comprehensive analysis of the complete MCTS exploration pathway leading to the PT4 algorithm (node 486, fitness = 5241.4). Figure~\ref{fig:interpretability2}a presents the full tree structure encompassing all nodes associated with the selected algorithm, revealing systematic patterns in knowledge accumulation and technique integration across multiple optimization phases (full MCTS tree data and visualization are available in Supplementary Information Section S6).

Critical to understanding the framework's discovery mechanism is the identification of five key algorithmic breakthroughs that emerge at specific nodes and propagate through subsequent generations. These innovations demonstrate systematic knowledge accumulation, with breakthrough techniques subsequently incorporated into descendant algorithms through evolutionary operations. The propagation patterns reveal progressive sophistication through iterative refinement and technique combination, with superior algorithmic components showing robust inheritance and integration capabilities that directly influence fitness improvements across generations. The complete tree structure demonstrates effective LLM-guided exploration that balances exploitation of promising directions with exploration of novel algorithmic territories, leading to high-performance detection strategies significantly exceeding conventional approaches.

\textbf{Stochastic Reproducibility Assessment.} Through comprehensive re-execution analysis of three pivotal evolutionary transitions using 100 independent runs each, we validated the consistency of breakthrough innovations in Figure~\ref{fig:interpretability2}b. 
These analyses confirm that breakthrough algorithmic innovations emerge through systematic discovery processes rather than fortuitous random variations, demonstrating stable technique inheritance and high-probability performance improvements across all re-executions despite increased algorithmic complexity, validating the framework's reliability for automated algorithm discovery and providing confidence in the generalizability of discovered techniques to broader gravitational wave detection challenges.

\subsection{Ablation Studies}

To validate Evo-MCTS's effectiveness, we conducted systematic component analysis across three critical dimensions: integrated optimization architecture, LLM model selection, and domain knowledge incorporation. These analyses reveal that superior performance emerges from synergistic component interactions rather than simple additive effects - addressing the vast search space problem through MCTS-guided reflection structuring, ensuring code generation quality via optimal LLM selection, and maintaining scientific relevance through domain knowledge integration. This multi-faceted approach demonstrates that successful automated discovery requires intelligent integration of search strategy, reasoning capability, and domain expertise beyond mere computational power.

\begin{figure*}
\centering
\includegraphics[width=\textwidth]{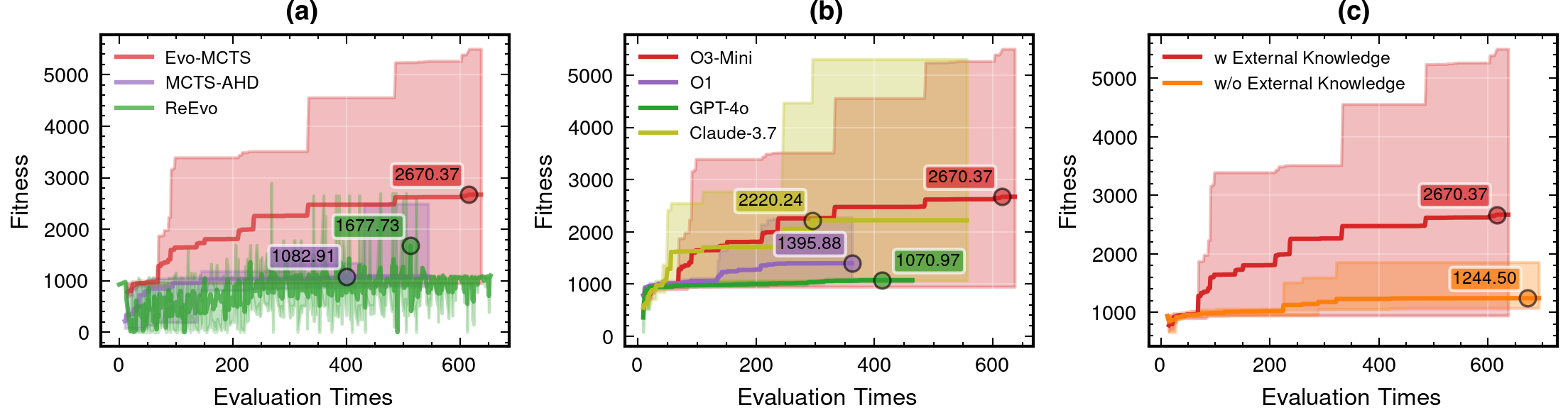}
\caption{\textbf{Framework Ablation Study and Design Validation.} \textbf{(a)} Integrated architecture validation comparing Evo-MCTS (red) against constituent components: MCTS-AHD (purple, 1,677.73 fitness) and ReEvo (green, 1,082.61 fitness). Evo-MCTS achieves superior performance (2,670.37 fitness) through synergistic combination of evolutionary population dynamics and tree-structured search. \textbf{(b)} LLM model selection analysis showing performance variation across foundation models: \texttt{o3-mini-medium} (red, 2,670.37), \texttt{claude-3-7-sonnet-20250219-thinking} (yellow, 2,220.24), \texttt{o1-2024-12-17} (purple, 1,395.88), and \texttt{gpt-4o-2024-11-20} (green, 1,070.97). Results demonstrate the superior performance of reasoning-enhanced models, with \texttt{o3-mini-medium} achieving 150\% improvement over general-purpose models. \textbf{(c)} Domain knowledge integration impact comparing frameworks with external knowledge (red, 2,670.37) versus without domain-specific guidance (orange, 1,244.50). The 115\% performance improvement demonstrates the essential role of scientific domain expertise in automated algorithm discovery. All curves represent averages over at least five independent runs with shaded regions indicating standard deviation. Results validate the framework's three core design principles: integrated optimization architecture, optimal model selection, and domain knowledge incorporation.}
\label{fig:framework_analysis}
\end{figure*}

\textbf{Synergistic Architecture Design.} Figure~\ref{fig:framework_analysis}a demonstrates the substantial benefits of combining evolutionary optimization with MCTS in LLM-guided algorithm discovery. Evo-MCTS achieves superior performance compared to its constituent components operating in isolation - pure MCTS-AHD~\cite{MCTS_AHD} and pure evolutionary optimization (ReEvo~\cite{ReEvo}), all of which leverage LLMs for code generation and algorithmic reasoning.

The performance hierarchy reveals critical insights about LLM-based optimization strategy effectiveness. Pure evolutionary approaches struggle with the vast search space and become trapped in local optima despite LLM guidance, exhibiting high variance with frequent suboptimal exploration in the LLM-generated algorithm space. MCTS-AHD provides improvement through systematic search space organization but with limited diversity in LLM-generated solutions. The full Evo-MCTS combines the best of both worlds: maintaining population diversity through evolutionary mechanisms while focusing computational resources on promising algorithmic directions through tree search guidance, all while maximizing the utilization of LLM reasoning capabilities. This integration achieves a remarkable 59.1\% improvement over MCTS-AHD alone, demonstrating that combining population-based diversity maintenance with tree-structured exploitation creates emergent optimization capabilities that exceed the sum of individual LLM-powered components.

\textbf{LLM Selection.} Figure~\ref{fig:framework_analysis}b investigates the impact of different foundation models on algorithmic discovery performance, revealing significant variations in code generation capability across state-of-the-art language models. The analysis establishes \texttt{o3-mini-medium} as our fiducial model configuration.

The performance hierarchy reveals insights about the relationship between model architecture and scientific code generation capability. Particularly intriguing is the superior performance of \texttt{claude-3-7-sonnet-20250219-thinking} over \texttt{o1-2024-12-17}, despite both being reasoning-enhanced models. This suggests that Claude's specific training methodologies and architectural choices may be better suited for sustained algorithmic reasoning tasks in gravitational wave detection algorithm development. The substantial performance gap between reasoning-enhanced models and general-purpose models underscores the critical importance of model architecture selection for scientific code generation tasks.

This demonstrates that framework effectiveness depends not merely on access to LLMs, but on selecting models with appropriate reasoning architectures for complex scientific applications. The robustness analysis shows consistent performance rankings across multiple runs, validating our model selection strategy while revealing that different LLM architectures exhibit distinct strengths in algorithmic reasoning and code synthesis.

\textbf{Necessity of Domain Knowledge.} Figure~\ref{fig:framework_analysis}c presents one of the most striking results: the dramatic impact of domain knowledge integration on algorithmic discovery performance. The comparison between frameworks with and without external knowledge reveals a performance difference of 115\%, demonstrating that domain-specific guidance is essential for effective automated algorithm discovery in specialized scientific domains.

The framework without external knowledge exhibits relatively flat optimization trajectories, while domain knowledge integration demonstrates sustained improvement by serving as a constraint mechanism that guides the vast search space toward physically meaningful solutions. The emphasis on non-linear processing significantly enhances gravitational wave signal detection in real-world non-Gaussian, non-stationary noise environments, successfully leveraging scientific expertise to accelerate discovery beyond pure computational search (domain knowledge templates and integration methodology detailed in Supplementary Information Section S1).

\section{Discussion}\label{sec:discussion}

The Evo-MCTS framework demonstrates that LLM-guided evolutionary search can effectively navigate complex algorithmic design spaces in scientific computing applications. Our results establish key methodological insights about interpretable algorithm discovery that extend beyond the gravitational wave detection demonstration to broader scientific computing domains.


\textbf{Methodological Validation.} 
The Evo-MCTS framework's performance on gravitational wave detection 
demonstrates key methodological capabilities transferable to broader 
scientific computing domains. The 59.1\% improvement over existing 
LLM-based frameworks (MCTS-AHD, ReEvo) validates the synergistic 
integration of tree-structured search with evolutionary population 
dynamics, while the 20.2\% improvement over domain-specific methods 
confirms that automated discovery can navigate complex design spaces 
effectively when guided by appropriate domain constraints.

The systematic convergence toward similar algorithmic structures 
across independent runs—despite stochastic LLM variations—provides 
empirical evidence for the framework's reliability as a discovery 
mechanism. This convergent behavior contrasts sharply with random 
search or pure evolutionary approaches, demonstrating that MCTS 
guidance combined with LLM reasoning enables focused exploration 
of promising algorithmic regions while maintaining solution diversity.

\textbf{Practical Considerations.} 
We acknowledge that competition-based benchmarks like MLGWSC-1, 
while providing standardized evaluation environments, have inherent 
limitations in capturing the full complexity of operational 
gravitational wave detection. The benchmark's simplified evaluation 
metrics enable controlled framework validation but do not encompass 
real-time processing requirements, comprehensive parameter estimation, 
or integration with multi-messenger astronomy infrastructure required 
for production deployment. Our framework's primary contribution lies 
in the methodology for automated algorithm discovery rather than 
proposing production-ready gravitational wave detection pipelines, 
with the benchmark serving as a challenging testbed that demonstrates 
framework capabilities on realistic scientific computing problems.


\textbf{Future Directions.} The evolutionary MCTS approach demonstrates broad potential for scientific algorithm discovery beyond gravitational waves. As Browne et al. note, MCTS ``can be used with little or no domain knowledge, and has succeeded on difficult problems where other techniques have failed''~\cite{browne2012Survey}. The framework's domain-agnostic architecture suggests applications in molecular optimization for drug discovery, materials design algorithms, and signal detection across astrophysical domains. The demonstrated interpretability advantages enable hybrid human-AI systems where algorithmic discoveries inform theoretical understanding while domain insights guide optimization, potentially accelerating scientific algorithm development across multiple disciplines.

\section{Methods}\label{sec:methods}

\subsection{Problem Formulation}\label{sec:problem_formulation}

This section formalizes automated algorithm discovery as a constrained optimization problem and introduces the LLM-guided evolutionary framework architecture, using gravitational wave detection as a concrete instantiation of the general framework.

\textbf{Algorithm Discovery Problem Instantiation.} 
To demonstrate the framework's capabilities on a challenging scientific computing problem, we instantiate the algorithm discovery framework for gravitational wave detection. Given dual-detector strain data $\mathbf{d}(t) \in \mathbb{R}^{2 \times N}$, where $\mathbf{d}(t) = [d_H(t), d_L(t)]^T$ represents the strain data from Hanford and Livingston interferometers sampled at $f_s = 2048$ Hz over finite observation windows of length $N$ samples, we seek to discover optimal detection algorithms that maximize performance while satisfying operational constraints.

\textbf{Optimization Objective.} 
The algorithm discovery problem is formulated as:
\begin{align}
a^* = \arg\max_{a \in \mathcal{A}} \quad & \mathcal{F}(a, \mathbf{d}) \\
\text{subject to} \quad & \|\Delta t_{\text{arrival}}\| \leq 0.2 \text{ seconds} \\
& T_{\text{comp}}(a) \leq T_{\max} \\
& E(a) \leq E_{\max} \\
& a: \mathbb{R}^{2 \times N} \rightarrow \mathbb{R}^3
\end{align}

where:
\begin{itemize}
\item $\mathcal{A} = \{a \mid a \text{ is executable and satisfies constraints}\}$ represents the space of executable detection algorithms
\item $\mathcal{F}(a, \mathbf{d}) = \int_{FAR_{\min}}^{FAR_{\max}} d_L(FAR; a, \mathbf{d}) \, d(FAR)$ measures detection performance as the area under the sensitive distance versus false alarm rate curve following the MLGWSC-1 protocol~\cite{Marlin2023First}
\item $\Delta t_{\text{arrival}}$ denotes trigger arrival time uncertainty for astrophysical parameter estimation
\item $T_{\text{comp}}(a)$ and $E(a)$ represent computational time and error handling trial count constraints with thresholds $T_{\max}$ and $E_{\max}$
\item Each algorithm $a$ maps dual-detector strain data to a three-column detection catalog table containing peak times, signal ranking statistics, and timing uncertainties
\end{itemize}

The fitness function $\mathcal{F}$ evaluates algorithms against ground truth labels $\mathbf{y}_{\text{true}}$ using the area under the curve (AUC) of sensitive distance versus false alarms per month, where $d_L(FAR; a, \mathbf{d})$ represents the sensitive distance at a given false alarm rate $FAR$, providing more physically meaningful performance assessment than traditional Receiver Operating Characteristic curves for gravitational wave detection applications.

\textbf{Algorithm Discovery Framework.}
The framework implements a general iterative search procedure that combines Monte Carlo Tree Search exploration with evolutionary population dynamics:

\[\mathbf{P}_{t+1} = \text{Evolve}(\mathbf{P}_t, \mathcal{L}, \mathcal{K}_{\text{domain}})\]

where $\mathbf{P}_t = \{a_1^{(t)}, a_2^{(t)}, \ldots, a_k^{(t)}\}$ represents the algorithm population at iteration $t$.

The framework comprises the following \textbf{key components:}
\begin{itemize}
\item \textbf{LLM Code Generation}: $\mathcal{L}: (\text{prompt}, \text{context}) \rightarrow \text{code}$ represents the language model for code generation, with operation-specific prompting strategies $\sigma: \mathcal{O} \rightarrow \mathcal{P}$ mapping evolutionary operations $\mathcal{O} = \{\text{PC}, \text{SC}, \text{PWC}, \text{PM}\}$ to specialized prompt templates $\mathcal{P}$.

\item \textbf{Domain Knowledge}: $\mathcal{K}_{\text{domain}} = (\mathcal{K}_{\text{physics}}, \mathcal{K}_{\text{methods}}, \mathcal{K}_{\text{constraint}})$ encapsulates domain-specific expertise including physical principles, algorithmic techniques, and computational constraints. For the gravitational wave detection instantiation, this includes detector physics, signal processing methods, and operational requirements.

\item \textbf{MCTS Selection}: Node selection follows Upper Confidence bounds applied to Trees (UCT) with adaptive exploration:
\[\pi_{\text{UCT}}(n) = \arg\max_{c \in \text{children}(n)} \left[ \text{normalized\_fitness}(c) + \text{adaptive\_exploration}(c) \right]\]
where the complete implementation with fitness normalization, adaptive exploration constants, and epsilon regularization is detailed in Section~\ref{sec:mcts_details}.
\end{itemize}

\textbf{Evolutionary Operations.}
The framework employs four evolutionary operations for algorithmic transformation, each utilizing operation-specific prompting strategies with the same underlying language model:
\begin{align}
\text{PC}(\mathbf{P}_t): \quad & a_{\text{new}} = \mathcal{L}(\text{prompt}_{\text{PC}}(a_p, a_r), \mathcal{K}_{\text{domain}}) \\
\text{SC}(\mathbf{P}_t): \quad & a_{\text{new}} = \mathcal{L}(\text{prompt}_{\text{SC}}(a_c, \{a_{s_i}\}), \mathcal{K}_{\text{domain}}) \\
\text{PWC}(\mathbf{P}_t): \quad & a_{\text{new}} = \mathcal{L}(\text{prompt}_{\text{PWC}}(\{a_{d_i}\}), \mathcal{K}_{\text{domain}}) \\
\text{PM}(\mathbf{P}_t): \quad & a_{\text{new}} = \mathcal{L}(\text{prompt}_{\text{PM}}(a_c, a_e), \mathcal{K}_{\text{domain}})
\end{align}
where PC = Parent Crossover, SC = Sibling Crossover, PWC = Path-wise Crossover, and PM = Point Mutation. Here, $a_p$ and $a_r$ denote parent and reference algorithms, $a_c$ represents the current algorithm, $\{a_{s_i}\}$ are sibling algorithms, $a_e$ is an elite algorithm, and $\{a_{d_i}\}$ are distant algorithm sets.

Each operation employs tailored prompt templates $\text{prompt}_{\text{op}}(\cdot)$ that guide the language model to generate appropriate algorithmic variants: Parent Crossover prompts emphasize combining successful features from parent algorithms, Sibling Crossover focuses on lateral exploration within similar algorithmic families, Path-wise Crossover promotes paradigm shifts by integrating distant algorithmic approaches, and Point Mutation targets localized refinements of existing implementations. Domain knowledge $\mathcal{K}_{\text{domain}}$ ensures adherence to domain-specific principles and physical constraints across all operations.

\textbf{Reflection and Adaptation.}
The system incorporates analytical reasoning through specialized reflection using the DeepSeek-R1 model:
\[\mathcal{K}_{\text{domain}}^{(t+1)} = \mathcal{K}_{\text{domain}}^{(t)} \cup \{\text{insights}(R(\text{history}_t, \text{performance}_t, \mathcal{K}_{\text{domain}}))\}\]
where $R: \mathcal{H} \times \mathcal{F}^* \times \mathcal{K}_{\text{domain}} \rightarrow \mathcal{I}$ represents the reflection function mapping MCTS history $\mathcal{H}$, fitness evaluations $\mathcal{F}^*$, and domain knowledge to actionable insights $\mathcal{I}$.

This reflection mechanism analyzes performance patterns across the MCTS tree to identify successful algorithmic principles and guide subsequent evolutionary operations toward promising regions of the solution space, enabling cumulative learning that incorporates both performance feedback and domain-specific constraints.

\textbf{Population Management.}
At each iteration, the algorithm population is updated through elite preservation with selection pressure $\beta$:
\begin{equation}
    \mathbf{P}_{t+1} = \text{Elite}(\mathbf{P}_t \cup \{a_{\text{new}}\}, k, \beta)
\end{equation}
where $\text{Elite}(\cdot, k, \beta)$ selects the top-$k$ algorithms based on fitness scores $\mathcal{F}(a_i, \mathbf{d})$ with selection probability: 
$p_{\text{select}}(a_i) = {\exp(\beta \cdot \mathcal{F}(a_i, \mathbf{d}))}/{\sum_{j=1}^{|{P}_t|} \exp(\beta \cdot \mathcal{F}(a_j, \mathbf{d}))}$

This formulation establishes a general framework for automated algorithm discovery as a constrained optimization problem in the space of executable algorithms, solved through LLM-guided evolutionary search with MCTS exploration and domain knowledge integration. The gravitational wave detection application demonstrates this framework's effectiveness on challenging scientific computing problems requiring continuous parameter optimization under strict physical constraints.

\subsection{LLM Integration for Code Generation}\label{sec:llm_integration}

The framework leverages state-of-the-art language models to transform algorithmic concepts into executable code, implementing a multi-model strategy that capitalizes on the complementary strengths of different architectures. This subsection details the model selection, prompting strategies, and error handling mechanisms that enable robust algorithmic discovery.

\textbf{Model Architecture and Task Allocation.} The framework employs a heterogeneous ensemble of four language models: \texttt{o3-mini-medium}, \texttt{o1-2024-12-17}, \texttt{gpt-4o-2024-11-20}, and \texttt{claude-3-7-sonnet-20250219-thinking} for code generation tasks. For reflection operations, we utilize \texttt{deepseek-r1-250120} exclusively due to its analytical reasoning capabilities.

\textbf{Prompting Strategy and Temperature Control.} All models operate with temperature 1.0 to optimize the trade-off between algorithmic diversity and code validity.

The prompting framework employs depth-aware adaptation mechanisms applicable across scientific computing domains. 
Task descriptions clarify optimization objectives, while depth information guides exploration scope: shallow nodes emphasize paradigm shifts and architectural changes, deeper nodes focus on parameter optimization and mathematical refinements. External domain knowledge integration provides optimization directives referencing established algorithmic principles and domain-specific constraints.
This adaptive architecture enables systematic solution space exploration while maintaining domain coherence and physical validity (complete templates in Supplementary Information Section S1, demonstrated through gravitational wave detection application).

\textbf{Error Handling and Iterative Refinement.} The framework implements a robust error recovery mechanism to handle code generation failures. When syntax errors or runtime exceptions occur during algorithm evaluation, the system captures detailed error information including stack traces and execution context. This diagnostic information is then incorporated into subsequent conversation rounds with the LLM to generate corrected implementations. The system attempts up to three correction iterations per failed algorithm. If all correction attempts fail, the corresponding node expansion is skipped to maintain computational efficiency. This approach ensures that the majority of generated algorithms remain executable while preventing infinite correction loops that could stall the evolutionary process (complete templates in Supplementary Information Section S1).

The analysis captures algorithmic innovations, signal processing techniques, performance expectations, and computational characteristics in compressed form. This enables efficient reference to previous discoveries without overwhelming the LLM context, facilitating continued exploration while maintaining algorithmic memory across generations (complete templates in Supplementary Information Section S1).

\textbf{Domain Knowledge Integration.} The framework incorporates domain-specific expertise through three structured prompt categories: initialization prompts defining problem-specific principles and data characteristics, evolution prompts encouraging effective algorithmic strategies and adaptive processing, and reflection prompts evaluating performance and computational efficiency (complete templates in Supplementary Information Section S1).

For the gravitational wave detection demonstration, the knowledge base prioritizes techniques appropriate for transient signal processing in non-stationary noise: adaptive thresholds, multi-scale analysis, and robust statistical estimators. Implementation emphasizes adaptive parameters over fixed values while maintaining computational efficiency.

This approach ensures domain constraint compliance while enabling exploration beyond conventional algorithmic paradigms, demonstrating the framework's capability to integrate scientific expertise with automated discovery.

\subsection{Evolutionary Operations Design and Seed Algorithm}\label{sec:evolutionary_ops}

\textbf{Seed Algorithm Architecture.} The optimization process begins with a deliberately simple seed function that establishes a baseline for algorithmic improvement (detailed implementation in Supplementary Information Section S1). This seed algorithm implements a conventional signal processing pipeline consisting of three sequential operations that represent standard approaches in gravitational wave data analysis.

The first operation performs frequency-domain whitening to normalize the detector noise characteristics:
\begin{equation}
X_{\text{white}}(f) = \frac{X(f)}{\sqrt{S(f)}}
\end{equation}
where $X(f)$ represents the Fourier transform of the input strain data, and $S(f)$ is the power spectral density estimate obtained via Welch's method using a 4096-sample window with 50\% overlap and Hann windowing. This whitening operation ensures that all frequency components contribute equally to subsequent analysis, compensating for the detector's frequency-dependent noise characteristics.

The second operation applies time-frequency decomposition using the short-time Fourier transform (STFT) to capture transient signal characteristics:
\begin{equation}
S_{xx}(f,\tau) = \left\| \sum_{n=0}^{N-1} x(n+\tau)\, w(n)\, e^{-j 2\pi f n / N} \right\|^2
\end{equation}
where $w(n)$ is a window function (256 samples with 128-sample overlap), $\tau$ is the time shift, and $N$ is the window length. The algorithm independently processes both Hanford (H1) and Livingston (L1) detector data, then combines the resulting spectrograms using simple averaging: 
\[
    \mathrm{TF}_{\text{metric}} = \frac{1}{2} \left\langle S_{xx}^{\text{H1}} + S_{xx}^{\text{L1}} \right\rangle_f
\]
where $\langle \cdot \rangle_f$ denotes frequency-bin averaging.

The final operation identifies candidate events through basic peak detection with fixed thresholds. The algorithm estimates background levels using the median and applies simple peak finding with predetermined height and prominence criteria. This approach represents a minimalist detection strategy that lacks the sophistication necessary for robust gravitational wave identification, particularly in the presence of non-Gaussian noise transients and weak signals.

\textbf{Initial Population Generation.} The evolutionary framework initializes with a single seed algorithm, then generates 8 diverse variants through systematic prompting variations (Figure~\ref{fig:framework}a). Each variant maintains identical input-output interfaces while implementing distinct signal processing approaches: alternative whitening schemes, varied time-frequency decomposition methods, and different peak detection strategies. Following initial generation, two Point Mutation operations are applied to create additional variants, resulting in a total population of 10 algorithms that form the depth-1 initial population for MCTS exploration.

\textbf{Elite Preservation Strategy.} The framework maintains an elite individual representing the best-performing algorithm discovered throughout evolution. This elite serves as a performance benchmark for new variants and provides genetic material during Point Mutation operations, which specifically leverage elite characteristics to guide targeted algorithmic improvements (Figure~\ref{fig:framework}c). The elite is updated whenever a new algorithm demonstrates superior performance, ensuring monotonic progress while maintaining access to the current best solution (Figure~\ref{fig:framework}b.4).

\textbf{Operation Sequence and Execution Strategy.} Each MCTS expansion level follows a structured sequence of evolutionary operations: Parent Crossover (PC) executes 5 times, Path-wise Crossover (PWC) executes 2 times, Sibling Crossover (SC) executes once, and Point Mutation (PM) executes twice. As observed in Figure~\ref{fig:interpretability2}a, this sequence balances four distinct algorithmic improvement mechanisms: (i) vertical knowledge transfer through PC operations that combine features from algorithms at different tree depths, (ii) long-range dependency capture via PWC operations that synthesize information along complete evolutionary trajectories, (iii) horizontal exploration using SC operations that facilitate exploration between algorithms of similar complexity based on nodes already generated at the same tree level, and (iv) fine-grained optimization through PM operations that introduce targeted modifications based on performance analysis. The systematic application of these operations ensures comprehensive exploration of the algorithmic solution space while maintaining computational efficiency through controlled expansion rates.

\subsection{Implementation of Monte Carlo Tree Search}\label{sec:mcts_details}

\textbf{MCTS Framework and Tree Policy.} The framework implements a standard Monte Carlo Tree Search algorithm adapted for algorithmic discovery, where each node represents an executable algorithm and tree expansion corresponds to algorithmic evolution (Figure~\ref{fig:framework}b). The MCTS operates through four canonical phases: selection, expansion, simulation, and backpropagation. However, our implementation modifies the traditional simulation phase by replacing random rollouts with direct algorithm evaluation on the gravitational wave detection task.

The selection phase traverses the tree from root to leaf using the Upper Confidence Bound applied to Trees (UCT) policy, balancing exploitation of high-performing algorithms with exploration of under-visited branches (Figure~\ref{fig:framework}c). Unlike traditional MCTS applications where leaf nodes represent terminal game states, our leaf nodes represent algorithms that can be further evolved through the four evolutionary operations (PC, SC, PWC, PM).

\textbf{UCT Score Calculation and Adaptive Exploration.} The UCT score for each node combines exploitation and exploration terms with an adaptive exploration strategy that accounts for the finite evaluation budget:

\begin{equation}
\text{UCT}(n) = \frac{Q(n) - Q_{\min}}{Q_{\max} - Q_{\min} + \epsilon} + c \cdot \sqrt{\frac{\ln(N(p) + 1)}{N(n) + \epsilon}}
\end{equation}

where $Q(n)$ represents the normalized fitness value of node $n$, $Q_{\min}$ and $Q_{\max}$ are the minimum and maximum fitness values observed across all nodes, $N(p)$ and $N(n)$ are the visit counts of the parent and current node respectively, $\epsilon$ is a small constant preventing division by zero, and $c$ is the exploration constant.

The exploration constant $c$ adapts dynamically based on the remaining evaluation budget:
\begin{equation}
c = c_0 \cdot \max\left(1 - \frac{t}{T}, 0\right)
\end{equation}
where $c_0$ is the initial exploration constant, $t$ is the current evaluation count, and $T$ is the maximum evaluation budget. This adaptive mechanism ensures that the algorithm emphasizes exploration early in the search when the budget is abundant, then gradually shifts toward exploitation as evaluations are consumed.

\textbf{Fitness Normalization and Q-Value Management.} The fitness values (corresponding to the objective function in MCTS) are normalized to the range $[0,1]$ to ensure consistent UCT calculations regardless of the absolute scale of performance metrics. The normalization uses running minimum and maximum values:

\begin{equation}
Q_{\text{normalized}} = \frac{Q_{\text{raw}} - Q_{\min}}{Q_{\max} - Q_{\min} + \epsilon}
\end{equation}

This normalization is crucial for maintaining meaningful exploration-exploitation balance, as it prevents algorithms with vastly different performance scales from skewing the selection process.

\textbf{Backpropagation and Value Updates.} The backpropagation phase updates node values along the path from the newly evaluated leaf to the root. Our implementation uses a discount factor approach that balances immediate performance with long-term potential:

\begin{equation}
Q(p) = Q(p) \cdot (1 - \gamma) + \max_{c \in \text{children}(p)} Q(c) \cdot \gamma
\end{equation}
where $Q(p)$ is the parent node's Q-value, $\gamma$ is the discount factor, and the maximum is taken over all child nodes. This update rule ensures that parent nodes reflect the performance of their best children while maintaining some memory of their previous estimates.

The backpropagation also maintains global statistics by updating the minimum and maximum Q-values across the entire tree, enabling consistent normalization for future UCT calculations. Additionally, the algorithm maintains a ranked list of all observed fitness values to support advanced selection strategies and performance analysis.

\textbf{Tree Expansion Strategy.} Node expansion occurs when the UCT selection phase reaches a leaf node that has been visited multiple times, indicating sufficient confidence in its potential. The expansion strategy creates one new child node per expansion operation, chosen through the evolutionary operations framework. This conservative expansion approach prevents explosive tree growth while ensuring thorough exploration of promising regions.

Each newly created node inherits structural information from its parent, including the algorithmic context and performance history. The initial Q-value for new nodes is set to the parent's Q-value, providing a reasonable starting estimate that will be refined through subsequent evaluations.

\textbf{Memory Management and Tree Pruning.} To maintain computational efficiency with limited memory resources, the implementation includes mechanisms for selective tree pruning. Nodes that demonstrate consistently poor performance relative to their siblings are marked for potential removal, while maintaining sufficient diversity to avoid premature convergence.

The tree structure also maintains subtree references that enable efficient traversal and analysis of evolutionary pathways. These references support the PWC operation by providing access to complete root-to-leaf trajectories without requiring expensive tree traversals.

\textbf{Convergence Detection and Termination.} The MCTS continues until either the evaluation budget is exhausted or convergence is detected through analysis of the Q-value distribution. Convergence is identified when the state-of-the-art (SOTA) algorithms show minimal improvement over a specified number of iterations, indicating that the search has reached a local optimum within the current exploration strategy.

This MCTS implementation creates a systematic framework for exploring the algorithmic space while maintaining computational efficiency and ensuring reproducible results. The combination of adaptive exploration, normalized fitness values, and efficient tree management enables the discovery of high-performing algorithms within reasonable computational budgets.

\subsection{Experimental Setup}\label{sec:experimental_setup}

\textbf{Benchmark Selection and Evaluation.} We selected the Machine Learning Gravitational-Wave Search Mock Data Challenge (MLGWSC-1) benchmark~\cite{Marlin2023First} as a challenging testbed for our approach. This benchmark provides a controlled evaluation environment with realistic complexity: actual detector noise, diverse signal morphologies, and standardized assessment protocols. Dataset 4 was selected for evaluation as it incorporates operational detector noise from the O3a observing run, providing a realistic application scenario for showcasing algorithmic discovery on scientific computing problems with domain-specific physical constraints.

Dataset 4 incorporates noise from both Hanford (H1) and Livingston (L1) detectors with quality filtering applied, spanning overlapping segments of minimum 2-hour duration. Signal injection parameters create a challenging evaluation scenario where at least 33\% of injected signals have optimal network SNR $< 4$, testing the framework's ability to discover algorithms that balance sensitivity with false-alarm control under realistic operational conditions.

\textbf{Algorithm Input/Output Interface.} Following the MLGWSC-1 specification, each algorithm processes HDF5 files containing raw detector data from both H1 and L1 detectors. The input files contain grouped datasets organized by integer start times, with each dataset including strain data and metadata attributes (\texttt{start\_time}, \texttt{delta\_t}). Our evolved algorithms output HDF5 files containing exactly three datasets of equal length: (i) \texttt{time} - GPS times of suspected gravitational wave events, (ii) \texttt{stat} - ranking statistics where larger values indicate higher detection confidence, and (iii) \texttt{var} - timing accuracy tolerance representing the maximum allowed separation between predicted and true event times.

\textbf{Evaluation Metrics and Performance Assessment.} The framework employs two primary evaluation metrics consistent with MLGWSC-1 standards. The false-alarm rate (FAR) measures the expected frequency of false-positive events exceeding a given ranking statistic threshold, calculated by applying algorithms to pure noise data and dividing event counts by total analyzed time. The sensitive distance quantifies detection capability at specified false-alarm rates, representing the maximum distance at which sources can be reliably detected. For uniformly distributed signals in volume, this reduces to the fraction of detected signals multiplied by the volume of a sphere with radius equal to the maximum injected distance. The area under the curve (AUC) metric integrates sensitive distance across the false-alarm rate range, providing a comprehensive performance indicator that balances detection sensitivity against false-alarm tolerance.

\textbf{Configuration of Training and Test Sets.} To ensure efficient optimization while maintaining statistical rigor, we partitioned the MLGWSC-1 Dataset 4 into training and test subsets. The training set comprises the first 7 days of data from early injection indices, enabling rapid algorithmic evaluation during the optimization process. This temporal partitioning ensures that the minimum false-alarm rate boundary is set at 4 events per month, corresponding to the statistical requirements for meaningful AUC calculation. The test set consists of 1 day of data selected from later temporal segments, providing independent validation of optimized algorithms (detailed data partitioning specifications provided in Supplementary Information Section S2).

\textbf{Diversity Metrics in Evolutionary Computation.} We implemented sophisticated diversity measurement following established practices in evolutionary computation~\cite{HSEvo}. Population encoding involves three preprocessing steps: (i) removing comments and docstrings using abstract syntax tree parsing to focus on functional algorithmic content, (ii) standardizing code snippets into common coding style following PEP 8 conventions to eliminate stylistic variations, and (iii) converting normalized code snippets to vector representations using the CodeT5+ embedding model to enable quantitative similarity analysis.

We employ two complementary diversity metrics: the Shannon Diversity Index, calculated as $H = -\sum_{i=1}^{n} p_i \log p_i$ where $p_i$ represents the frequency of the $i$-th unique algorithmic variant in the population, and the Complexity Index of Diversity (CID), computed as $CID = \frac{1}{n}\sum_{i=1}^{n} \frac{||x_i - \bar{x}||}{||\bar{x}||+\epsilon}$ where $x_i$ represents the embedding vector of the $i$-th algorithm, $\bar{x}$ is the population centroid, and $\epsilon$ prevents division by zero. These metrics provide complementary perspectives on population diversity: Shannon index captures algorithmic variety while CID measures structural complexity differences.

\textbf{LLM API Configuration and Model Selection.} Our framework integrates multiple state-of-the-art language models through official APIs from OpenAI, Anthropic, and DeepSeek~\cite{OpenAI2023GPT4, OpenAI2024Learning, Anthropic2025Claude, DeepSeek_R1_2025}. For code generation tasks, we employ thinking-enhanced models including \texttt{o3-mini-medium}, \texttt{o1-2024-12-17}, \texttt{gpt-4o-2024-11-20}, and \texttt{claude-3-7-sonnet-20250219-thinking}, selected for their demonstrated capabilities in multi-step reasoning and code synthesis~\cite{chen2021evaluating, Bubeck2023Sparks}. For reflection mechanisms, we exclusively utilize \texttt{deepseek-r1-250120} due to its superior performance in analytical reasoning tasks~\cite{DeepSeek_R1_2025}, particularly its ability to identify subtle performance patterns and propose targeted algorithmic improvements based on empirical observations. We designate \texttt{o3-mini-medium} as our fiducial model configuration, providing the baseline reference for performance comparisons and ensuring consistency across experimental runs. All models operate with temperature parameter set to 1.0 to balance creative exploration with syntactic reliability~\cite{Ouyang2022Training}.

\textbf{Hyperparameter Configuration and Experimental Design.} The Evo-MCTS framework employs carefully tuned hyperparameters optimized for gravitational wave detection algorithm discovery. Initial population size is set to 10 algorithms, balancing computational efficiency with adequate diversity for effective exploration. MCTS depth is limited to 10 levels, providing sufficient hierarchical structure for complex algorithmic development while maintaining computational tractability. Each experimental configuration undergoes 5 independent runs with different random seeds to ensure statistical robustness and enable confidence interval estimation.

The 5-run experimental design serves multiple purposes: (i) quantifying algorithmic discovery reliability across different initialization conditions, (ii) enabling statistical analysis of optimization trajectories and convergence patterns, and (iii) providing robust diversity measurements that account for stochastic variations in the evolutionary process. Comprehensive results from all individual runs are documented in Supplementary Information Section S3, enabling detailed analysis of inter-run variability and optimization consistency.

\textbf{Ablation Studies and Comparative Analysis.} To validate the effectiveness of our integrated Evo-MCTS approach, we conduct comprehensive ablation studies comparing against two established frameworks: ReEvo~\cite{ReEvo} (pure evolutionary mechanism) and MCTS-AHD~\cite{MCTS_AHD} (pure MCTS mechanism). These comparisons isolate the contributions of different algorithmic components while maintaining consistent evaluation protocols.

ReEvo represents the evolutionary optimization baseline, employing genetic algorithms with traditional crossover and mutation operations powered by LLMs but without tree-structured exploration. The framework implements population-based optimization through iterative code generation and selection, focusing on evolutionary diversity maintenance and fitness-driven selection pressure. MCTS-AHD implements pure Monte Carlo Tree Search for automated heuristic design without evolutionary population dynamics or reflection mechanisms. This approach emphasizes tree-based exploration with UCT-guided node selection but lacks the multi-generational insight synthesis and population diversity maintenance that characterizes evolutionary approaches.

By comparing Evo-MCTS against these component frameworks using identical LLM integration protocols and evaluation budgets, we quantify the synergistic benefits of combining evolutionary population dynamics with structured tree search exploration. The experimental design maintains consistent evaluation metrics, hardware configurations, and statistical analysis procedures across all three frameworks, with computational fairness ensured through equivalent LLM evaluation counts as the unified iteration metric. This approach enables direct performance comparison while isolating the specific contributions of different optimization strategies under identical resource constraints. The comprehensive multi-run analysis results across all frameworks are presented in Supplementary Information Section S3, demonstrating the statistical robustness of our comparative evaluation.

\textbf{Protocol for Edge Robustness Analysis.} To validate the consistency of algorithmic breakthroughs, we developed a systematic edge re-execution protocol. For each selected evolutionary transition, we perform 100 independent re-executions using identical prompt templates while maintaining all experimental conditions constant except for the LLM sampling parameters. Each re-execution employs the same parent algorithms, evolutionary operation type, and external knowledge integration templates used in the original optimization run.

The re-execution protocol preserves all deterministic components (fitness evaluation, data preprocessing, statistical analysis) while allowing natural variation in LLM-generated code through different random seeds. This approach enables quantification of discovery mechanism robustness while accounting for the inherent stochasticity in large language model outputs. Statistical analysis employs standard descriptive metrics (mean, standard deviation) combined with confidence interval estimation to characterize the reliability of breakthrough discovery patterns, as illustrated in Figure~\ref{fig:interpretability2}(b).

\textbf{Computational Resources and Parallelization Techniques.} The numerical calculations in this study were carried out on the ORISE Supercomputer, equipped with 32-core x86 processors and 4 GPGPU accelerators per node, enabling efficient parallel execution of LLM API calls and algorithm evaluations. The Evo-MCTS framework employs asynchronous parallelization to maximize resource utilization, allowing multiple LLM requests to be processed concurrently while maintaining synchronization for tree updates and performance analysis. This parallelization strategy significantly reduces overall computational time while ensuring that all evaluations are performed under consistent conditions.

\section{Data availability}

The MLGWSC-1 Dataset 4 used in this study is publicly available at \href{https://github.com/gwastro/ml-mock-data-challenge-1/}{https://github.com/gwastro/ml-mock-data-challenge-1/}. The dataset includes real detector noise from LIGO Hanford (H1) and Livingston (L1) observatories during the O3a observing run. 

The ReEvo framework implementation is available at \href{https://github.com/ai4co/reevo}{https://github.com/ai4co/reevo} with the original codebase and documentation. The MCTS-AHD framework can be accessed through its official repository at \href{https://github.com/zz1358m/MCTS-AHD-master}{https://github.com/zz1358m/MCTS-AHD-master}. Both frameworks were used for comparative analysis following their original specifications and hyperparameter configurations.

Training and test data partitions, along with detailed preprocessing specifications, are provided in the Supplementary Information Section S2. All experimental results and algorithm performance metrics are available upon reasonable request to the corresponding author.

\section{Code availability}

The complete source code for the Evo-MCTS framework is publicly available at \href{https://github.com/iphysresearch/evo-mcts}{https://github.com/iphysresearch/evo-mcts}. The repository includes all implementation details, experimental configurations, and reproducibility instructions.
The code is written in Python and is fully reproducible with the provided environment and dependencies. The repository contains comprehensive documentation, example usage scripts, and all necessary configuration files to replicate the experimental results presented in this study.

\backmatter

\section*{Acknowledgments}

We thank the LIGO Scientific Collaboration for providing the MLGWSC-1 dataset and establishing the evaluation framework that enabled this research. We acknowledge the computational resources provided by the ORISE Supercomputer facility, which were essential for the large-scale LLM-based optimization experiments.

We are grateful to OpenAI, Anthropic, and DeepSeek for providing API access to their language models, which formed the core of our algorithmic discovery framework. We also thank the developers of the ReEvo and MCTS-AHD frameworks for making their implementations publicly available for comparative analysis.

H.W. is supported by the National Key Research and Development Program of China (Grant No. 2021YFC2203004), the National Natural Science Foundation of China (NSFC) (Grant Nos. 12405076, 12247187, 12147103), the National Astronomical Data Center (Grant No. NADC2023YDS-01), and the Fundamental Research Funds for the Central Universities.

\newpage
\begin{center}
{\Huge Supplementary information}\\[1em]
\end{center}

\begin{appendices}

\renewcommand{\thesection}{S\arabic{section}}
\renewcommand{\thesubsection}{S\arabic{section}.\arabic{subsection}}
\renewcommand{\thesubsubsection}{S\arabic{section}.\arabic{subsection}.\arabic{subsubsection}}

\renewcommand{\appendixname}{}
\renewcommand{\appendixpagename}{}
\renewcommand{\appendixtocname}{}

\section{LLM Prompting Templates}\label{sec:S1}

This section provides comprehensive details of the prompting strategies employed across different phases of the algorithmic discovery process. The templates are designed to guide language models through systematic reasoning while incorporating domain-specific knowledge and maintaining consistency across evolutionary operations.

\textbf{System Context and Task Definition.} All interactions with the LLM ensemble begin with a standardized system prompt that establishes the expert role and problem context:
\begin{lstlisting}[basicstyle=\footnotesize\ttfamily, breaklines=true, breakindent=0pt, breakatwhitespace=false, columns=flexible]
You are an expert in gravitational wave signal detection algorithms. Your task is to design heuristics that can effectively solve optimization problems. The task involves constructing a pipeline for gravitational wave signal detection. This pipeline will encompass data conditioning and time-frequency transformations as part of the signal processing workflow. The input will consist of raw, finite-length dual-channel gravitational wave data from the H1 and L1 detectors. The pipeline will be tested on segmented data spanning several weeks, with each segment having variable length (7000s-30000s). Each segment's dual-channel data will be directly used as input. The ultimate goal is to produce a catalog of potential gravitational wave signals, where each trigger includes information such as GPS time, ranking statistic, and the timing accuracy of the prediction. This systematic approach is essential for effectively identifying and cataloging candidate gravitational wave signals.
\end{lstlisting}
This system prompt serves multiple purposes: (i) establishing domain expertise expectations, (ii) defining the specific optimization context, (iii) specifying input data characteristics, and (iv) clarifying the expected output format and evaluation criteria.

\subsection{Initial Algorithm Generation Prompts}

\textbf{Seed Function Template and Analysis Framework.} The initial algorithm generation process begins with a structured analysis of the seed function to establish baseline understanding. The seed function analysis template guides the LLM through systematic examination of the foundational algorithm:
\begin{lstlisting}[basicstyle=\footnotesize\ttfamily, breaklines=true, breakindent=0pt, breakatwhitespace=false, columns=flexible]
## Seed Function Analysis Task
Analyze the foundational algorithm's design strategy to establish baseline understanding for Monte Carlo Tree Search (MCTS) exploration. This first-level analysis will guide subsequent optimization directions.

## Seed Function Implementation
```python
{prompt_seed_func}
```
- **Technical implementation details**: {prompt_other_inf}
- **Performance impact rationale**: {prompt_inout_inf}

## Context for Analysis
This initial analysis at MCTS depth first-level should:
- Identify core algorithmic mechanisms
- Extract fundamental processing stages
- Surface high-level optimization opportunities
- Establish baseline for diversity generation
{external_knowledge}

## Analysis Requirements
1. Characterize the seed's core approach in one sentence containing:
    - Primary computational strategy
    - Key transformation stages
    - Fundamental signal processing techniques
    - Overall optimization philosophy

2. Focus on architectural-level characteristics rather than implementation details

3. Description must fit within single braces and avoid:
    - Code references
    - Parameter-level details
    - Performance assessments
    - Comparative statements

## Output Format Rules
- Return optimization strategies within SINGLE BRACE
- Ensure entire response can be parseable by regex: \\{{(.*?)\\}} with DOTALL flag
\end{lstlisting}

\textbf{Seed Algorithm Specification.} The seed function implements a three-stage linear signal processing pipeline that serves as the evolutionary starting point:

\begin{itemize}
    \item Stage 1: Data Conditioning and Whitening
\begin{minted}[frame=lines,framesep=2mm,baselinestretch=1.2,fontsize=\tiny,linenos,breaklines]{python}
def data_conditioning(strain_h1: np.ndarray, strain_l1: np.ndarray, times: np.ndarray) -> tuple[np.ndarray, np.ndarray, np.ndarray]:
    window_length = 4096
    dt = times[1] - times[0]
    fs = 1.0 / dt
    
    def whiten_strain(strain):
        strain_zeromean = strain - np.mean(strain)
        freqs, psd = signal.welch(strain_zeromean, fs=fs, nperseg=window_length,
                                    window='hann', noverlap=window_length//2)
        smoothed_psd = np.convolve(psd, np.ones(32) / 32, mode='same')
        smoothed_psd = np.maximum(smoothed_psd, np.finfo(float).tiny)
        white_fft = np.fft.rfft(strain_zeromean) / np.sqrt(np.interp(np.fft.rfftfreq(len(strain_zeromean), d=dt), freqs, smoothed_psd))
        return np.fft.irfft(white_fft)

    whitened_h1 = whiten_strain(strain_h1)
    whitened_l1 = whiten_strain(strain_l1)
    
    return whitened_h1, whitened_l1, times
\end{minted}
    \item Stage 2: Time-Frequency Decomposition
\begin{minted}[frame=lines,framesep=2mm,baselinestretch=1.2,fontsize=\tiny,linenos,breaklines]{python}
def compute_metric_series(h1_data: np.ndarray, l1_data: np.ndarray, time_series: np.ndarray) -> tuple[np.ndarray, np.ndarray]:
    fs = 1 / (time_series[1] - time_series[0])
    f_h1, t_h1, Sxx_h1 = signal.spectrogram(h1_data, fs=fs, nperseg=256, noverlap=128, mode='magnitude', detrend=False)
    f_l1, t_l1, Sxx_l1 = signal.spectrogram(l1_data, fs=fs, nperseg=256, noverlap=128, mode='magnitude', detrend=False)
    tf_metric = np.mean((Sxx_h1**2 + Sxx_l1**2) / 2, axis=0)
    gps_mid_time = time_series[0] + (time_series[-1] - time_series[0]) / 2
    metric_times = gps_mid_time + (t_h1 - t_h1[-1] / 2)
    
    return tf_metric, metric_times
\end{minted}
    \item Stage 3: Peak Detection and Trigger Generation
\begin{minted}[frame=lines,framesep=2mm,baselinestretch=1.2,fontsize=\tiny,linenos,breaklines]{python}
def calculate_statistics(tf_metric, t_h1):
    background_level = np.median(tf_metric)
    peaks, _ = signal.find_peaks(tf_metric, height=background_level * 1.0, distance=2, prominence=background_level * 0.3)
    peak_times = t_h1[peaks]
    peak_heights = tf_metric[peaks]
    peak_deltat = np.full(len(peak_times), 10.0)  # Fixed uncertainty value
    return peak_times, peak_heights, peak_deltat
\end{minted}
\end{itemize}

\textbf{Template Variables and Customization.} The prompting template incorporates several customizable variables that enable systematic variation generation:

\begin{itemize}
    \item \texttt{prompt\_seed\_func}: Complete seed function implementation
    \item \texttt{prompt\_other\_inf}: Technical implementation details including sampling rates, window parameters, and algorithmic constraints
    \item \texttt{prompt\_inout\_inf}: Performance impact rationale explaining the relationship between input characteristics and expected output quality
    \item \texttt{external\_knowledge}: Domain-specific knowledge injection including gravitational wave physics, detector characteristics, and signal morphology constraints
\end{itemize}

\textbf{Output Format Requirements.} All generated algorithms must conform to the standardized interface:
\begin{minted}[frame=lines,framesep=2mm,baselinestretch=1.2,fontsize=\tiny,linenos,breaklines]{python}
def pipeline_v{N}(strain_h1: np.ndarray, strain_l1: np.ndarray, times: np.ndarray) -> tuple[np.ndarray, np.ndarray, np.ndarray]:
    # Algorithm implementation for seed function
    # ...
    # Stage 1: Data Conditioning and Whitening
    whitened_h1, whitened_l1, data_times = data_conditioning(strain_h1, strain_l1, times)
    # Stage 2: Time-Frequency Decomposition
    tf_metric, metric_times = compute_metric_series(whitened_h1, whitened_l1, data_times)
    # Stage 3: Peak Detection and Trigger Generation
    peak_times, peak_heights, peak_deltat = calculate_statistics(tf_metric, metric_times)
    return peak_times, peak_heights, peak_deltat
\end{minted}
This interface consistency ensures that all generated algorithms can be evaluated within the same framework while enabling diverse internal implementations.

\subsection{Parent Crossover Implementation}
The Parent Crossover (PC) operation represents a sophisticated evolutionary operation that combines algorithmic components from two reference implementations at different levels of the MCTS hierarchy. This operation is designed to preserve successful characteristics from both parent algorithms while introducing novel enhancements that exceed simple interpolation.

\textbf{Template Structure and Crossover Strategy.} The PC operation employs a structured template that guides the LLM through systematic analysis and synthesis of two parent algorithms:

\begin{lstlisting}[basicstyle=\footnotesize\ttfamily, breaklines=true, breakindent=0pt, breakatwhitespace=false, columns=flexible]
## Task Overview
Develop a novel algorithm that strategically combines components from two reference implementations while introducing innovative enhancements. The solution must demonstrate measurable improvements beyond simple interpolation of existing approaches.
Current Depth Level: [Level {depth}]

## Implementation Analysis
### Code Comparison
1. VERSION A (Baseline Implementation):
```python
{worse_code}
```

2. VERSION B (Enhanced Implementation): 
```python
{better_code}
```

### Strengths to Combine
```text
{reflection}
```

Key Synthesis Requirements:
- Preserve 2 distinct advantages from Version A
- Incorporate 3 critical enhancements from Version B
- Identify 1 synergistic improvement opportunity

## Architecture Strategy
{external_knowledge}

### Depth-Specific Synthesis Guidelines (Depth={depth})
1. Structural Synthesis (Depth 1-2):
    - Create hybrid control flow combining best elements from both versions
    - Example: "Combine Version A's iteration structure with Version B's termination conditions"
    - Forbid direct replication of either version's architecture

2. Implementation Fusion (Depth 3-4):
    - Develop novel parameter hybridization techniques
    - Example: "Blend Version A's exploration mechanism with Version B's exploitation strategy"
    - Require at least one innovative combination per functional module

3. Mathematical Innovation (Depth 5+):
    - Derive new computational operators through version synthesis
    - Example: "Fuse Version A's approximation method with Version B's error correction"
    - Mandate 10-20% computational complexity reduction    
\end{lstlisting}
This template structure ensures that the crossover operation is not merely concatenative but involves intelligent analysis and strategic combination of algorithmic strengths.

\textbf{Depth-Adaptive Synthesis Guidelines.} The PC operation implements depth-specific strategies that adapt the crossover complexity based on the current position in the MCTS tree:
\begin{itemize}
    \item Structural Synthesis (Depth 1-2):
        \begin{itemize}
            \item Focuses on combining high-level architectural elements from both parent algorithms
            \item Creates hybrid control flow structures that merge the best organizational patterns
            \item Example directive: ``Combine Version A's iteration structure with Version B's termination conditions"
            \item Explicitly forbids direct replication of either parent's complete architecture
        \end{itemize}
    \item Implementation Fusion (Depth 3-4):
        \begin{itemize}
            \item Emphasizes parameter hybridization and functional module integration
            \item Develops novel approaches to blend algorithmic strategies
            \item Example directive: ``Blend Version A's exploration mechanism with Version B's exploitation strategy"
            \item Requires at least one innovative combination per functional module
        \end{itemize}
    \item Mathematical Innovation (Depth 5+):
        \begin{itemize}
            \item Derives new computational operators through sophisticated version synthesis
            \item Focuses on mathematical justification for algorithmic improvements
            \item Example directive: ``Fuse Version A's approximation method with Version B's error correction"
            \item Mandates 10-20\% computational complexity reduction alongside performance gains
        \end{itemize}
\end{itemize}

\textbf{Innovation Requirements and Quality Assurance.} The PC operation enforces strict innovation standards to ensure that generated algorithms represent genuine improvements:
\begin{itemize}
    \item Core Innovation Targets:
        \begin{itemize}
            \item Synthesize 3+ novel elements not present in either parent version
            \item Resolve 2 fundamental limitations identified through comparative analysis
            \item Introduce 1 breakthrough enhancement with rigorous mathematical justification
            \item Demonstrate non-trivial performance gains over both parent algorithms
            \item Prohibit direct replication of complete code blocks from either parent
        \end{itemize}
\end{itemize}

\textbf{Reflection Generation Process.} Before conducting the crossover synthesis, the system generates analytical insights through a depth-adaptive reflection template:

\begin{lstlisting}[basicstyle=\footnotesize\ttfamily, breaklines=true, breakindent=0pt, breakatwhitespace=false, columns=flexible]
## Task Objective
Analyze optimization patterns across algorithm versions and generate depth-specific improvement strategies. Current MCTS Depth: depth/max_depth={depth}/{max_depth}

## Depth-Specific Focus
- Shallow (Depth 1-2): Structural patterns & control flow
- Medium (Depth 3-4): Implementation techniques & parameterization
- Deep (Depth 5+): Mathematical formulations & computational primitives

## Algorithm Comparison
- Original (Suboptimal)
```python
{code_worse}
```

- Improved (Optimized) 
```python
{code_better}
```

## Depth-Adaptive Analysis
### 1. Core Pattern Extraction
For {depth}-level analysis:
- Shallow: Compare control structures/algorithmic paradigms
- Medium: Analyze parameter configurations/function compositions
- Deep: Examine mathematical operators/numerical methods

### 2. Optimization Principle Generation
Generate 3-5 transferable rules that:
- Directly address {depth}-specific limitations
- Contain concrete parameter values from improved version
- Maintain functional equivalence

## Output Format Rules
- Return optimization strategies within SINGLE BRACE
- Ensure entire response can be parseable by regex: \\{{(.*?)\\}} with DOTALL flag
\end{lstlisting}

This reflection generation produces the \texttt{reflection} variable used in the main crossover template.

\textbf{Reflection-Guided Analysis.} The crossover process incorporates a reflection component that analyzes the strengths and weaknesses of both parent algorithms:
\begin{lstlisting}[basicstyle=\footnotesize\ttfamily, breaklines=true, breakindent=0pt, breakatwhitespace=false, columns=flexible]
## Requirements
1. Core Innovation Targets:
    - Synthesize 3+ novel elements not present in either version
    - Resolve 2 fundamental limitations identified in analysis
    - Introduce 1 breakthrough enhancement with mathematical justification
    - Demonstrate non-trivial performance gain over both versions
    - Prohibit direct replication of complete code blocks
\end{lstlisting}
This reflection analysis is generated through the \texttt{deepseek-r1-250120} model and provides crucial insights that guide the synthesis process. The reflection identifies:
\begin{itemize}
    \item Computational advantages in each parent algorithm
    \item Structural design patterns that contribute to performance
    \item Potential synergistic combinations that could yield emergent benefits
    \item Limitation patterns that should be addressed in the offspring
\end{itemize}

\textbf{Output Format and Validation.} The PC operation enforces a standardized output format that ensures both human readability and automated processing:
\begin{lstlisting}[basicstyle=\footnotesize\ttfamily, breaklines=true, breakindent=0pt, breakatwhitespace=false, columns=flexible]
2. Output Format:
- Place the core design idea in a sentence within a brace BEFORE the function definition
- For the core design idea format: \\{{A hybrid gravitational wave detection pipeline...}}
- Implement as Python function: {func_name}
- Inputs: {input_count} parameter(s) ({joined_inputs})
- Outputs: {output_count} return value(s) ({joined_outputs})
- Follow: {inout_inf}
- Constraints: {other_inf}
- IMPORTANT: All output code MUST be valid Python syntax. Do not place description text inside curly braces within the function body.
- Example of correct format:
    \\{{Core design description here}}
    ```python
    def pipeline_v2(strain_h1: np.ndarray, strain_l1: np.ndarray, times: np.ndarray) -> tuple[np.ndarray, np.ndarray, np.ndarray]:
        """Core design description can alternatively be placed here as a docstring"""
        # Function implementation...
    ``` 
\end{lstlisting}

\subsection{Sibling Crossover Implementation}
The Sibling Crossover (SC) operation implements a sophisticated two-phase approach that leverages peer algorithm insights to generate improved offspring. Unlike Parent Crossover, which combines algorithms from different hierarchical levels, SC focuses on horizontal knowledge transfer between algorithms at the same MCTS depth, promoting diversity while maintaining comparable complexity levels.

\textbf{Two-Phase Architecture.} The SC operation employs a unique two-stage process: first generating optimization hints through multi-level reflection analysis (Phase 1), then implementing concrete algorithmic improvements based on these insights (Phase 2). This separation enables more targeted optimization by allowing the system to first identify improvement opportunities before implementing solutions.

\textbf{Phase 1: Multi-Level Reflection Analysis.} The first phase generates depth-specific optimization hints by synthesizing insights from sibling algorithms and parent-level analysis:
\begin{lstlisting}[basicstyle=\footnotesize\ttfamily, breaklines=true, breakindent=0pt, breakatwhitespace=false, columns=flexible]
## Task Overview
Generate depth-specific optimization hints for gravitational wave detection algorithms by synthesizing multi-level reflections. 
Current Optimization Depth: {parent_depth}/{max_depth} (shallow: structural patterns, medium: implementation techniques, deep: mathematical details)

## Contextual Insights
1. Peer Algorithm Reflections (Depth {parent_depth}):
    - Formatted as performance-annotated entries: [No.N Brother Reflection | Score: X]<reflection>
    - Time-ordered weighting (newest=highest priority) with objective score-based ranking
    - Includes full technical post-mortems from immediate ancestors
{parent_reflections}

2. Father Algorithm Analysis (Depth {father_depth}):
{father_reflection}

## Hint Generation Requirements
1. Produce 3-5 executable optimization directives that:
    - Integrate cross-depth insights from peer implementations
    - Target {parent_depth}-level (shallow: structural patterns, medium: implementation techniques, deep: mathematical details) components for improvement
    - Formulate mathematically sound enhancements
    - Align with gravitational wave data processing objectives

2. Output Format Rules
    - Return optimization strategies within SINGLE BRACE
    - Ensure entire response can be parseable by regex: \\{{(.*?)\\}} with DOTALL flag
    - Focus on {parent_depth}-appropriate modifications
    - Emphasize time-domain processing optimizations

## Critical Constraints
- Each directive must correspond to concrete code changes
- Explicitly connect to reflection insights where applicable
- Maintain strict {parent_depth}-level focus in all suggestions
- Exclude explanatory text within the hint brace
- Prioritize modifications matching current depth's optimization type
\end{lstlisting}

\textbf{Sibling Selection and Weighting Strategy.} The SC operation employs a sophisticated parent selection mechanism that prioritizes high-performing sibling algorithms:
\begin{minted}[frame=lines,framesep=2mm,baselinestretch=1.2,fontsize=\tiny,linenos,breaklines]{python}
# Select parents based on objective value weights
other = [ind for ind in pop if ind['code'] != father['code']]
weights = [1.0 / (-ind['fitness'] + 1e-10) for ind in other]  # Lower objective = higher weight
normalized_weights = [w / sum(weights) for w in weights]
parents = random.choices(other, weights=normalized_weights, k=min(self.m, len(other)))
\end{minted}
This weighting strategy ensures that successful algorithmic patterns from high-performing siblings are more likely to influence the offspring generation process.

\textbf{Depth-Adaptive Optimization Focus.} The first phase implements depth-specific optimization strategies that adapt to the current position in the MCTS tree:
\begin{itemize}
    \item Shallow Depth (1-2): Focuses on structural patterns and control flow restructuring
    \item Medium Depth (3-4): Emphasizes implementation techniques and numerical optimizations
    \item Deep Depth (5+): Concentrates on mathematical details and advanced computational methods
\end{itemize}

\textbf{Phase 2: Concrete Algorithm Implementation.} The second phase transforms the optimization hints into executable algorithms:
\begin{lstlisting}[basicstyle=\footnotesize\ttfamily, breaklines=true, breakindent=0pt, breakatwhitespace=false, columns=flexible]
## Algorithm Optimization Task
Develop an enhanced gravitational signal processing algorithm for interferometer data analysis by implementing concrete improvements from multi-level code analysis.

## Technical Context
1. Optimization Depth Specifications:
- Current Focus Level: {depth} (max_depth={max_depth})
    (1-2: Control flow restructuring, 3-4: Numerical computation optimizations, 5+: Advanced linear algebra methods)
- Code Analysis Insights from Prior Level:
```text
{reflection}
```

2. Base Implementation Details:
[Functional Purpose] {algorithm_description}
[Core Implementation] 
```python
{algorithm_code}
```

## Implementation Directives (Depth {depth}):
- Shallow (1-2): Restructure control flow using reflection suggestion (e.g., split data conditioning/analysis phases)
- Medium (3-4): Apply numerical optimizations from reflection (e.g., FFT window size optimization)
- Deep (5+): Implement matrix computation improvements from reflection (e.g., regularized inverse covariance)

{external_knowledge}

## Output Format
- Place the core design idea in a sentence within a brace BEFORE the function definition
- For the core design idea format: \\{{A hybrid gravitational wave detection pipeline...}}
- Implement as Python function: {func_name}
- Inputs: {input_count} parameter(s) ({joined_inputs})
- Outputs: {output_count} return value(s) ({joined_outputs})
- Follow: {inout_inf}
- Constraints: {other_inf}
- IMPORTANT: All output code MUST be valid Python syntax. Do not place description text inside curly braces within the function body.
- Example of correct format:
    \\{{Core design description here}}
    ```python
    def pipeline_v2(strain_h1: np.ndarray, strain_l1: np.ndarray, times: np.ndarray) -> tuple[np.ndarray, np.ndarray, np.ndarray]:
        """Core design description can alternatively be placed here as a docstring"""
        # Function implementation...
    ```

## Important Notes
- Focus on algorithmic improvements rather than code style changes
- Ensure the new implementation directly addresses the reflection insights    
\end{lstlisting}

\textbf{Reflection Processing and Template Variables.} The SC operation processes multiple sources of algorithmic insight through structured template variables:
\begin{itemize}
    \item \texttt{parent\_reflections}: Performance-annotated reflections from peer algorithms, formatted as ranked entries with objective scores
    \item \texttt{father\_reflection}: Analysis from the immediate parent algorithm at depth-1
    \item \texttt{reflection}: Synthesized optimization hints generated in Phase 1
    \item \texttt{algorithm\_description}: Functional description of the base algorithm
    \item \texttt{algorithm\_code}: Complete implementation of the parent algorithm
\end{itemize}

\textbf{Quality Assurance and Validation.} The two-phase approach enables comprehensive quality control:
\begin{itemize}
    \item Phase 1 Validation:
        \begin{itemize}
            \item Ensures reflection insights are depth-appropriate
            \item Validates mathematical soundness of optimization suggestions
            \item Confirms alignment with gravitational wave processing objectives
        \end{itemize}
    \item Phase 2 Validation:
        \begin{itemize}
            \item Verifies syntactic correctness of generated code
            \item Confirms interface compliance with standardized function signatures
            \item Tests algorithmic improvements against reflection insights
            \item Validates computational efficiency claims
        \end{itemize}
\end{itemize}

\textbf{Temporal Weighting and Performance Ranking.} The SC operation implements sophisticated temporal weighting that prioritizes recent algorithmic discoveries while maintaining objective score-based ranking:
\begin{itemize}
    \item Time-ordered weighting: Newer algorithms receive higher priority in the reflection synthesis
    \item Performance-based ranking: Algorithms with better objective scores contribute more heavily to the optimization hints
    \item Cross-depth integration: Insights from both peer algorithms and parent-level analysis are systematically combined
\end{itemize}
This comprehensive approach ensures that SC operations generate algorithms that not only improve upon their immediate ancestors but also incorporate the collective intelligence of high-performing siblings, leading to more robust and efficient gravitational wave detection strategies.

\subsection{Point Mutation Implementation}
Point Mutation (PM) operations introduce targeted modifications to individual algorithms based on performance analysis, implementing two distinct approaches that offer different levels of sophistication and computational investment. The framework provides both single-stage direct improvement and two-stage reflection-driven enhancement strategies.

\textbf{Single-Stage Point Mutation: Direct Algorithm Improvement.} The operation implements a straightforward approach that directly compares an original algorithm with a high-performing elite algorithm to generate improvements. This method prioritizes computational efficiency while maintaining effective algorithmic enhancement.

Template Structure:
\begin{lstlisting}[basicstyle=\footnotesize\ttfamily, breaklines=true, breakindent=0pt, breakatwhitespace=false, columns=flexible]
## Task Overview
You will analyze an original algorithm, an improved version of it, and create a new enhanced algorithm. Below are the key components:

## Algorithm Details
1. ORIGINAL ALGORITHM:
    - Description: {original_algorithm_description}
    - Code:
```python
{original_algorithm_code}
```
    - **Objective Value**: {original_objective_value}

2. BETTER ALGORITHM (Reference Implementation):
    - Description: {better_algorithm_description}
    - Code: 
```python
{better_algorithm_code}
```
    - **Objective Value**: {better_objective_value}
    - Improvement Insights: 
```text
{better_algorithm_reflection}
```

## Implementation Requirements
1. Analyze the differences between the original and better algorithms
2. Create a new algorithm that:
    - Incorporates successful elements from the better algorithm
    - Addresses limitations revealed in the improvement insights
    - Produces better results than the original algorithm
3. Output format requirements:
    - Place the core design idea in a sentence within a brace BEFORE the function definition
    - For the core design idea format: \\{{A hybrid gravitational wave detection pipeline...}}
    - Implement as Python function: {func_name}
    - Inputs: {input_count} parameter(s) ({joined_inputs})
    - Outputs: {output_count} return value(s) ({joined_outputs})
    - Follow: {inout_inf}
    - Constraints: {other_inf}
    - IMPORTANT: All output code MUST be valid Python syntax. Do not place description text inside curly braces within the function body.
    - Example of correct format:
        \\{{Core design description here}}
        ```python
        def pipeline_v2(strain_h1: np.ndarray, strain_l1: np.ndarray, times: np.ndarray) -> tuple[np.ndarray, np.ndarray, np.ndarray]:
            """Core design description can alternatively be placed here as a docstring"""
            # Function implementation...
        ```
{external_knowledge}

## Important Notes
- Focus on algorithmic improvements rather than code style changes
- Ensure the new implementation directly addresses the reflection insights
\end{lstlisting}

\textbf{Two-Stage Point Mutation: Reflection-Driven Enhancement.} The operation implements a sophisticated two-phase approach that mirrors the sibling crossover methodology but focuses on individual algorithm improvement rather than horizontal knowledge transfer.

\textbf{Phase 1: Strategic Reflection Generation.} The first phase synthesizes insights from multiple sources to generate comprehensive optimization guidelines:
\begin{lstlisting}[basicstyle=\footnotesize\ttfamily, breaklines=true, breakindent=0pt, breakatwhitespace=false, columns=flexible]
## Task Overview
Generate optimized technical guidelines for gravitational wave detection algorithms through systematic analysis of multi-generational reflection insights. Focus on enhancing data conditioning pipelines, time-frequency analysis methods, noise suppression techniques, and H1-L1 detector coherence optimization. Produce executable directives addressing: waveform recognition precision, computational complexity management, and non-stationary noise differentiation while maintaining strict API compliance.

## Input Context
1. NEW INSIGHTS FROM RECENT ITERATIONS:
    - Formatted as performance-annotated entries: [Parent N Reflection | Score: X]<reflection>
    - Time-ordered weighting (newest=highest priority) with objective score-based ranking
    - Includes full technical post-mortems from immediate ancestors
{parent_reflections}

2. LONG-TERM REFLECTION REPOSITORY:
    - Contains battle-tested insights from top 1% performers
    - 3x weighting factor for architectural-level insights
    - Curated through 3-stage filtration: 
        1. Statistical significance validation
        2. Cross-generational effectiveness verification
        3. Compatibility check with current detector configurations
{elite_reflection}

## Implementation Requirements
1. Perform weighted synthesis of reflections
2. Generate 3-5 technically-grounded optimization directives
3. Prioritize:
    - Mitigation of historical implementation flaws
    - Amplification of proven effective patterns
    - Weighted integration of multi-generational insights

## Output Format
- Return all guidelines within SINGLE BRACE
- Ensure entire response can be parseable by regex: \\{{(.*?)\\}} with DOTALL flag
- Concrete technical directives only
- No explanatory text or formatting
\end{lstlisting}

\textbf{Phase 2: Concrete Algorithm Implementation.} The second phase transforms the strategic insights into executable algorithmic improvements:
\begin{lstlisting}[basicstyle=\footnotesize\ttfamily, breaklines=true, breakindent=0pt, breakatwhitespace=false, columns=flexible]
## Task Overview
Leverage insights from prior strategic reflection to architecturally enhance the gravitational wave detection algorithm. Develop improvements that directly address identified limitations in CRITICAL REFLECTION INSIGHTS while preserving core functionality through:

1. Stage-level architectural modifications informed by reflection analysis
2. Reflection-driven noise reduction and coherence enhancement strategies
3. Time-frequency analysis variations targeting specific weaknesses identified
4. H1-L1 synthesis improvements based on cross-detector insights

Generate architecturally distinct variants that implement reflection-derived concepts through fundamental structural changes.

## Input Context
1. CRITICAL REFLECTION INSIGHTS (Improvement Basis):
```text
{reflection}
```

2. REFERENCE IMPLEMENTATION:
[Description] {elite_algorithm_description}
[Baseline Code]
```python
{elite_algorithm_code}
```

## Implementation Requirements
1. Execute reflection-guided analysis:
    - Map reflection insights to specific code components
    - Identify 2-3 architectural limitations in current implementation
2. Propose improvements that directly convert reflection insights into:
    - Enhanced signal path architecture
    - Novel noise handling structures
    - Optimized computational patterns
    - Advanced detector synergy mechanisms
3. Maintain strict interface compatibility with existing system integration

{external_knowledge}

## Output Format
- Place the core design idea in a sentence within a brace BEFORE the function definition
- For the core design idea format: \\{{A hybrid gravitational wave detection pipeline...}}
- Implement as Python function: {func_name}
- Inputs: {input_count} parameter(s) ({joined_inputs})
- Outputs: {output_count} return value(s) ({joined_outputs})
- Follow: {inout_inf}
- Constraints: {other_inf}
- IMPORTANT: All output code MUST be valid Python syntax. Do not place description text inside curly braces within the function body.
- Example of correct format:
    \\{{Core design description here}}
    ```python
    def pipeline_v2(strain_h1: np.ndarray, strain_l1: np.ndarray, times: np.ndarray) -> tuple[np.ndarray, np.ndarray, np.ndarray]:
        """Core design description can alternatively be placed here as a docstring"""
        # Function implementation...
    ```

## Important Notes
- Focus on algorithmic improvements rather than code style changes
- Ensure the new implementation directly addresses the reflection insights
\end{lstlisting}

\textbf{Selection Strategies and Elite Integration.} Both PM operations leverage the elite offspring as a performance benchmark and source of successful algorithmic patterns. The key distinction lies in their selection strategies:
\begin{itemize}
    \item Single-Stage Selection Strategy:
        \begin{itemize}
            \item Selects a single parent algorithm from the population (excluding the elite)
            \item Directly compares parent performance with elite offspring
            \item Implements immediate improvement through direct analysis
        \end{itemize}
    \item Two-Stage Selection Strategy:
        \begin{itemize}
            \item Selects multiple parent algorithms for comprehensive reflection analysis
            \item Incorporates both recent algorithmic insights and long-term elite patterns
            \item Implements sophisticated multi-generational knowledge synthesis
        \end{itemize}
\end{itemize}

\textbf{Computational Efficiency Considerations.} The two PM approaches offer different computational trade-offs:
\begin{itemize}
    \item Single-Stage Advantages:
        \begin{itemize}
            \item Single-stage processing reduces computational overhead
            \item Direct comparison enables rapid algorithm improvement
            \item Simplified prompting reduces LLM token consumption
        \end{itemize}
    \item Two-Stage Advantages:
        \begin{itemize}
            \item Two-stage processing enables more sophisticated optimization
            \item Multi-generational insight integration leads to more robust improvements
            \item Reflection-driven approach produces more interpretable algorithmic modifications
        \end{itemize}
\end{itemize}

\textbf{Template Variable Integration.} Both PM operations incorporate comprehensive template variables that enable systematic algorithmic improvement:
\begin{itemize}
    \item Common Variables:
        \begin{itemize}
            \item \texttt{func\_name}, \texttt{input\_count}, \texttt{output\_count}: Interface specification
            \item \texttt{joined\_inputs}, \texttt{joined\_outputs}: Parameter documentation
            \item \texttt{better\_algorithm\_description}, \texttt{better\_algorithm\_code}: Elite algorithm details
            \item \texttt{original\_objective\_value}, \texttt{better\_objective\_value}: Performance metrics
            \item \texttt{better\_algorithm\_reflection}: Elite algorithm insights
        \end{itemize}
    \item Two-Stage Variables:
        \begin{itemize}
            \item \texttt{parent\_reflections}: Multi-parent reflection synthesis
            \item \texttt{elite\_reflection}: Long-term elite insights
            \item \texttt{reflection}: Generated optimization guidelines
        \end{itemize}
\end{itemize}

\textbf{Quality Assurance and Validation.} Both PM operations implement rigorous validation procedures:
\begin{itemize}
    \item Single-Stage Validation:
        \begin{itemize}
            \item Direct performance comparison with both parent and elite algorithms
            \item Verification of improvement insight integration
            \item Confirmation of interface compliance
        \end{itemize}
    \item Two-Stage Validation:
        \begin{itemize}
            \item Two-stage validation covering both reflection generation and implementation
            \item Cross-generational consistency checking
            \item Architectural improvement verification
        \end{itemize}
\end{itemize}

The dual PM approach provides flexibility in algorithmic improvement strategies, enabling the framework to adapt to different optimization scenarios while maintaining consistent quality standards and interface compliance.

\subsection{Path-wise Crossover Implementation}
Path-wise Crossover (PWC) operations synthesize information along complete root-to-leaf trajectories in the MCTS tree, capturing long-range dependencies and enabling global optimization strategies. The framework implements two distinct PWC approaches that differ in their analytical methodologies: reflection-based synthesis and comprehensive algorithm analysis.

\textbf{Reflection-Based Path-wise Crossover: Multi-Algorithm Insight Synthesis.} The operation implements a two-stage process that analyzes reflection patterns across multiple algorithms in a complete MCTS path to identify generalizable optimization principles.

\textbf{Phase 1: Cross-Algorithm Pattern Analysis.} The first phase extracts recurring technical strategies from multiple algorithm reflections:
\begin{lstlisting}[basicstyle=\footnotesize\ttfamily, breaklines=true, breakindent=0pt, breakatwhitespace=false, columns=flexible]
## Task Overview
Analyze and synthesize technical reflections from multiple algorithm iterations to identify cross-algorithm optimization patterns and guide next-generation algorithm design. Prioritize extraction of generalizable technical principles over implementation-specific details.
Current Optimization Depth: depth/max_depth={depth}/{max_depth} (shallow: structural patterns, medium: implementation techniques, deep: mathematical details)

## Input Context
Analyzing {num_algorithms} algorithm reflections from MCTS exploration trajectories. Technical reflections follow depth-specific analysis requirements. Structural format: [No.N algorithm's reflection (depth: X)]<reflection>
{algorithm_reflections}

## Reflection Requirements
1. **Pattern Identification** (Key Observed Patterns):
    - Extract 2-3 recurring technical strategies (e.g. "Multi-scale wavelet decomposition" not "used Morlet wavelet")
    - Categorize by analysis level: 
        * Structural: Component architecture (e.g. "Parallel filter banks")
        * Implementation: Algorithmic choices (e.g. "Adaptive thresholding")
        * Mathematical: Core transforms (e.g. "Orthogonal matching pursuit")

2. **Technical Pathway Analysis** (Promising Technical Pathways):
    - Identify under-utilized but theoretically sound approaches (e.g. "Sparse representation in frequency domain")
    - Specify required technical components without code details (e.g. "Requires: Overcomplete basis construction")

3. **Optimization Principles** (Strategic Optimization Principles):
    - Formulate depth-specific guidelines (e.g. "At mathematical level: Maximize time-frequency product $\leq$ 0.5")
    - Relate physical constraints to algorithmic parameters (e.g. "Wavelet duration should match typical glitch durations")

4. **Specificity Balance**:
    - Technical specificity: Name mathematical concepts (e.g. "Gabor uncertainty") and signal processing domains
    - Implementation avoidance: Omit code structures (e.g. "Avoid: 'Use 3 nested loops'")

## Output Format Rules
- Return optimization strategies within SINGLE BRACE
- Ensure entire response can be parseable by regex: \\{{(.*?)\\}} with DOTALL flag
- Do not include markdown formatting or additional explanations
\end{lstlisting}

\textbf{Phase 2: Algorithm Implementation.} The second phase transforms the synthesized insights into concrete algorithmic improvements:
\begin{lstlisting}[basicstyle=\footnotesize\ttfamily, breaklines=true, breakindent=0pt, breakatwhitespace=false, columns=flexible]
## Task Overview
Develop an enhanced gravitational wave detection algorithm through targeted modifications addressing specific technical shortcomings identified in the reflection analysis.

## Input Context
[Critical Reflection Insights]
```text
{reflection}
```

[Baseline Implementation]
[Functional Description] {algorithm_description}
[Current Codebase]
```python
{algorithm_code}
```

{external_knowledge}

## Output Format
- Place the core design idea in a sentence within a brace BEFORE the function definition
- For the core design idea format: \\{{A hybrid gravitational wave detection pipeline...}}
- Implement as Python function: {func_name}
- Inputs: {input_count} parameter(s) ({joined_inputs})
- Outputs: {output_count} return value(s) ({joined_outputs})
- Follow: {inout_inf}
- Constraints: {other_inf}
- IMPORTANT: All output code MUST be valid Python syntax. Do not place description text inside curly braces within the function body.
- Example of correct format:
  \\{{Core design description here}}
  ```python
  def pipeline_v2(strain_h1: np.ndarray, strain_l1: np.ndarray, times: np.ndarray) -> tuple[np.ndarray, np.ndarray, np.ndarray]:
      """Core design description can alternatively be placed here as a docstring"""
      # Function implementation...
  ```

## Important Notes
- Focus on algorithmic improvements rather than code style changes
- Ensure the new implementation directly addresses the reflection insights
\end{lstlisting}

\textbf{Comprehensive Algorithm Analysis Path-wise Crossover: Multi-Level Technical Synthesis.} The operation implements a more sophisticated analytical approach that examines complete algorithm implementations across different depth levels.

\textbf{Phase 1: Multi-Level Technical Analysis.} The first phase conducts comprehensive analysis of algorithms along the complete MCTS path:
\begin{lstlisting}[basicstyle=\footnotesize\ttfamily, breaklines=true, breakindent=0pt, breakatwhitespace=false, columns=flexible]
## Task Objective
Synthesize technical insights from algorithm evolution MCTS path to guide targeted improvements. Current Analysis Level: depth/max_depth={depth}/{max_depth} (1-2: structural, 3-4: implementation, 5+: mathematical)

## Depth-Specific Focus
- Shallow (Depth 1-2): Structural patterns & control flow
- Medium (Depth 3-4): Implementation techniques & parameterization
- Deep (Depth 5+): Mathematical formulations & computational primitives

## Input Context
Analyzing {num_algorithms} algorithm reflections from MCTS exploration trajectories. Technical reflections follow depth-specific analysis requirements. Structural format: [No.N algorithm's reflection (depth: X)]<description><objective><code>
{parent_info}

## Synthesis Process
1. Cross-Level Insight Integration:
    - Identify key recurring technical strategies across abstraction levels
    - Note level-specific constraints affecting current implementations

2. Domain Compliance Verification:
    - Validate approaches against gravitational wave signal characteristics
    - Check numerical reliability across different implementation levels

3. Improvement Planning:
    - Structural: Adjust data processing pipelines
    - Implementation: Optimize critical parameter relationships
    - Mathematical: Enhance core transformation components

## Technical Workflow
### 1. Multi-Level Technical Analysis
Structural -> Compare module composition and interaction patterns
Implementation -> Assess parameter sensitivity and adaptation logic
Mathematical -> Examine transformation kernels and precision handling

### 2. Level-Appropriate Optimization
For current depth={depth}:
    - Select 2-4 improvement focus areas with technical rationale
    - Define implementation requirements for each focus area
    - Establish verification criteria with domain constraints

## Output Format Rules
- Return optimization strategies within SINGLE BRACE
- Ensure entire response can be parseable by regex: \\{{(.*?)\\}} with DOTALL flag
- Do not include markdown formatting or additional explanations
\end{lstlisting}

\textbf{Phase 2: Algorithm Implementation.} The operation shares the same implementation phase as the reflection-based path-wise crossover, utilizing the reflection-based PWC template for consistent output formatting and algorithmic generation.

\textbf{Methodological Distinctions.} The key differences between the reflection-based path-wise crossover and the comprehensive algorithm analysis PWC lie in their analytical strategies:
\begin{itemize}
    \item Reflection-Based PWC:
        \begin{itemize}
            \item Focuses on synthesizing existing reflection insights from multiple algorithms
            \item Emphasizes pattern recognition across previously analyzed algorithmic behaviors
            \item Prioritizes extraction of generalizable technical principles over implementation details
            \item Categorizes insights by structural, implementation, and mathematical analysis levels
        \end{itemize}
    \item Comprehensive Algorithm Analysis PWC:
        \begin{itemize}
            \item Conducts direct analysis of complete algorithm implementations
            \item Examines algorithmic components across multiple depth levels simultaneously
            \item Integrates cross-level insights through systematic technical workflow
            \item Emphasizes domain compliance verification and improvement planning
        \end{itemize}
\end{itemize}

\textbf{Depth-Adaptive Processing.} Both PWC operations implement depth-specific analysis strategies that adapt to the current position in the MCTS tree:
\begin{itemize}
    \item Shallow Depth Focus (1-2):
        \begin{itemize}
            \item Structural patterns and component architecture analysis
            \item Control flow restructuring and module composition optimization
            \item Data processing pipeline adjustments
        \end{itemize}
    \item Medium Depth Focus (3-4):
        \begin{itemize}
            \item Implementation techniques and algorithmic parameter optimization
            \item Critical parameter relationship assessment
            \item Numerical computation enhancement strategies
        \end{itemize}
    \item Deep Depth Focus (5+):
        \begin{itemize}
            \item Mathematical formulation analysis and computational primitive optimization
            \item Transformation kernel examination and precision handling
            \item Advanced linear algebra method integration
        \end{itemize}
\end{itemize}

\textbf{Path Trajectory Analysis.} Both operations process algorithms along complete MCTS paths, with depth tracking that enables comprehensive evolutionary analysis:
\begin{itemize}
    \item Reflection-Based PWC Path Processing:
        \begin{itemize}
            \item Analyzes reflection patterns from algorithms at decreasing depth levels
            \item Tracks depth-specific insights through structured format annotations
            \item Synthesizes cross-depth technical strategies for optimization guidance
        \end{itemize}
    \item Comprehensive Algorithm Analysis PWC Path Processing:
        \begin{itemize}
            \item Examines complete algorithm implementations with performance metrics
            \item Integrates algorithmic descriptions, objective values, and code analysis
            \item Conducts multi-level technical synthesis across the entire path trajectory
        \end{itemize}
\end{itemize}

\textbf{Template Variable Integration.} Both PWC operations incorporate sophisticated template variables that enable comprehensive path analysis:
\begin{itemize}
    \item Common Variables:
        \begin{itemize}
            \item \texttt{depth}, \texttt{max\_depth}: Depth-specific processing parameters
            \item \texttt{num\_algorithms}: Path length and analysis scope
            \item \texttt{func\_name}, \texttt{input\_count}, \texttt{output\_count}: Interface specifications
            \item \texttt{external\_knowledge}: Domain knowledge integration
        \end{itemize}
    \item Reflection-Based PWC Variables:
        \begin{itemize}
            \item \texttt{algorithm\_reflections}: Multi-algorithm reflection synthesis
            \item \texttt{reflection}: Generated optimization insights
        \end{itemize}
    \item Comprehensive Algorithm Analysis PWC Variables:
        \begin{itemize}
            \item \texttt{parent\_info}: Complete algorithm implementation details
            \item \texttt{current\_algorithm\_description}, \texttt{current\_algorithm\_code}: Baseline algorithm specifications
            \item \texttt{current\_objective\_value}: Performance reference metrics
        \end{itemize}
\end{itemize}

\textbf{Quality Assurance and Validation.} Both PWC operations implement comprehensive validation procedures:
\begin{itemize}
    \item Reflection-Based PWC Validation:
        \begin{itemize}
            \item Pattern identification verification across multiple algorithm reflections
            \item Technical pathway analysis consistency checking
            \item Optimization principle formulation validation
        \end{itemize}
    \item Comprehensive Algorithm Analysis PWC Validation:
        \begin{itemize}
            \item Multi-level technical analysis coherence verification
            \item Domain compliance checking across different implementation levels
            \item Cross-level insight integration validation
        \end{itemize}
\end{itemize}

The dual PWC approach provides complementary strategies for capturing long-range dependencies in the MCTS tree, enabling the framework to synthesize insights across complete evolutionary trajectories while maintaining depth-specific optimization focus and domain knowledge integration.

\subsection{Domain Knowledge Integration}
Domain knowledge integration serves as a critical component that ensures generated algorithms remain grounded in gravitational wave detection principles while encouraging exploration beyond traditional linear processing methods. The framework incorporates specialized domain expertise through structured knowledge templates that guide algorithmic development toward physically meaningful and computationally efficient solutions.

\textbf{External Knowledge Template Structure.} The domain knowledge integration employs a comprehensive template that emphasizes non-linear processing approaches and adaptive algorithmic strategies:
\begin{lstlisting}[basicstyle=\footnotesize\ttfamily, breaklines=true, breakindent=0pt, breakatwhitespace=false, columns=flexible]
### External Knowledge Integration
1. **Non-linear** Processing Core Concepts:
    - Signal Transformation: 
        * Non-linear vs linear decomposition
        * Adaptive threshold mechanisms
        * Multi-scale analysis
    
    - Feature Extraction:
        * Phase space reconstruction
        * Topological data analysis
        * Wavelet-based detection
    
    - Statistical Analysis:
        * Robust estimators
        * Non-Gaussian processes
        * Higher-order statistics

2. Implementation Principles:
    - Prioritize adaptive over fixed parameters
    - Consider local vs global characteristics
    - Balance computational cost with accuracy
\end{lstlisting}

\textbf{Non-linear Processing Emphasis.} The domain knowledge framework explicitly prioritizes non-linear algorithmic approaches over traditional linear methods, recognizing that gravitational wave signals exhibit complex, transient characteristics that require sophisticated analysis techniques. This emphasis addresses fundamental limitations in conventional matched filtering approaches that rely heavily on linear processing assumptions.

\textbf{Signal Transformation Guidance.} The domain knowledge provides specific guidance on signal transformation strategies that leverage advanced signal processing concepts:
\begin{itemize}
    \item Non-linear vs Linear Decomposition: The framework encourages exploration of non-linear decomposition methods that can capture complex signal morphologies beyond the capabilities of traditional Fourier-based approaches. This includes techniques such as empirical mode decomposition, intrinsic mode functions, and adaptive basis construction.
    \item Adaptive Threshold Mechanisms: Rather than employing fixed threshold values, the domain knowledge promotes adaptive thresholding strategies that respond to local signal characteristics and noise conditions. This approach enables more robust detection performance across diverse observational scenarios.
    \item Multi-scale Analysis: The framework emphasizes multi-scale signal analysis techniques that can simultaneously capture both short-duration transient signals and longer-duration continuous wave sources. This includes wavelet-based methods, time-frequency analysis, and hierarchical decomposition strategies.
    \item Feature Extraction Methodologies. The domain knowledge incorporates advanced feature extraction approaches that extend beyond traditional signal processing paradigms:
        \begin{itemize}
            \item Phase Space Reconstruction: The framework encourages exploration of phase space reconstruction techniques that can reveal hidden dynamical structures in gravitational wave data. This includes embedding dimension analysis, recurrence analysis, and attractor reconstruction methods.
            \item Topological Data Analysis: The domain knowledge promotes topological data analysis approaches that can identify persistent features and structural patterns in high-dimensional gravitational wave data. This includes persistent homology, Mapper algorithms, and topological feature extraction.
            \item Wavelet-based Detection: The framework emphasizes wavelet-based detection strategies that can provide optimal time-frequency resolution for transient signal analysis. This includes continuous wavelet transforms, discrete wavelet decomposition, and wavelet packet analysis.
        \end{itemize}
    \item Statistical Analysis Enhancement. The domain knowledge integrates sophisticated statistical analysis techniques that account for the complex noise characteristics of gravitational wave detectors:
        \begin{itemize}
            \item Robust Estimators: The framework promotes robust statistical estimators that can maintain performance in the presence of outliers and non-Gaussian noise distributions. This includes median-based estimators, M-estimators, and trimmed mean approaches.
            \item Non-Gaussian Processes: The domain knowledge emphasizes analysis techniques that can handle non-Gaussian noise processes commonly encountered in gravitational wave data. This includes heavy-tailed distributions, skewed probability models, and non-stationary noise characterization.
            \item Higher-order Statistics: The framework encourages exploration of higher-order statistical moments and cumulants that can capture subtle signal characteristics beyond second-order analysis. This includes bispectrum analysis, higher-order moment estimation, and polyspectral techniques.
        \end{itemize}
    \item Implementation Principles and Constraints. The domain knowledge provides specific implementation principles that guide algorithmic development toward practical and efficient solutions:
        \begin{itemize}
            \item Adaptive Parameter Prioritization: The framework emphasizes adaptive parameter selection over fixed parameter values, enabling algorithms to respond dynamically to changing signal and noise conditions. This principle encourages exploration of learning-based parameter adjustment, feedback control mechanisms, and online adaptation strategies.
            \item Local vs Global Characteristics: The domain knowledge promotes consideration of both local signal characteristics and global data patterns, enabling algorithms to balance fine-grained analysis with comprehensive signal understanding. This includes local stationarity analysis, global trend estimation, and multi-resolution processing approaches.
            \item Computational Cost-Accuracy Balance: The framework provides guidance on balancing computational efficiency with detection accuracy, ensuring that generated algorithms remain practical for real-time implementation while maintaining scientific rigor. This includes complexity analysis, algorithmic optimization, and performance benchmarking considerations.
        \end{itemize}
\end{itemize}

\textbf{Integration Across Evolutionary Operations.} The domain knowledge template is systematically integrated across all evolutionary operations (PC, SC, PM, PWC) through the {external\_knowledge} template variable. This ensures consistent application of gravitational wave detection principles regardless of the specific evolutionary strategy employed.

\textbf{Physical Validity Assurance.} The domain knowledge template ensures that all generated algorithms respect fundamental physical constraints related to gravitational wave signal characteristics, detector limitations, and noise properties.

\textbf{Computational Feasibility.} The implementation principles guide algorithmic development toward computationally feasible solutions that can be practically implemented within the constraints of current computational resources and real-time processing requirements.

This comprehensive domain knowledge integration creates a robust framework for scientifically grounded algorithmic discovery, ensuring that the evolutionary process generates algorithms that are both innovative and practically applicable to gravitational wave detection challenges.

\subsection{Error Handling and Iterative Refinement}

The Evo-MCTS framework incorporates a robust error handling mechanism that enables iterative refinement of generated algorithms through automated debugging and correction processes. When generated code encounters execution errors, fails to detect signals, or exceeds computational time limits, the system employs a rechat strategy using advanced reasoning models to diagnose and resolve issues.

\textbf{Error Detection and Classification.} The framework monitors three primary failure modes during algorithm execution: (i) runtime exceptions and syntax errors that prevent code execution, (ii) algorithmic failures where no gravitational wave signals are detected despite their presence in the data, and (iii) computational timeout scenarios where algorithms exceed predefined execution limits. Each failure mode triggers specific diagnostic protocols tailored to the underlying issue type.

\textbf{Iterative Refinement Protocol.} Upon error detection, the system implements a structured refinement process through a carefully designed prompt content structure. The system constructs conversation messages in a specific format to facilitate effective error correction:

Initially, when no system content is provided, the framework creates a message list containing a single user role entry with the original prompt content (\texttt{prompt\_content}):

\texttt{messages = [\{"role": "user", "content": prompt\_content\}]}

When a rechat response is available (indicating a previous failed attempt), the system extends the conversation by first appending the assistant's previous response (\texttt{rechat\_response}):

\texttt{messages.append(\{"role": "assistant", "content": rechat\_response\})}

Subsequently, the system adds a new user message that explicitly requests debugging and issue resolution (\texttt{prompt\_content}):

\texttt{messages.append(\{"role": "user", "content": "Your previous code had execution errors, couldn't find signals, or timed out. Please debug and fix the issues:\textbackslash n\textbackslash n" + prompt\_content\})}

This structured approach maintains the conversational context while providing clear guidance for error correction, ensuring that the assistant understands both the original requirements and the specific issues that need to be addressed.

\textbf{Automated Debugging Integration.} The error handling system leverages advanced reasoning capabilities to analyze failed algorithms and propose targeted corrections. This approach maintains the evolutionary optimization trajectory while addressing immediate technical obstacles that could otherwise terminate the search process. The iterative refinement ensures that promising algorithmic concepts are not discarded due to implementation errors, instead receiving corrective guidance to achieve functional implementations.

\subsection{Post-Generation Analysis and Knowledge Extraction}

The post-generation analysis phase extracts interpretable insights from evolved algorithms through automated knowledge distillation. This process transforms the raw algorithmic implementations into concise, human-readable descriptions that capture the essential design principles and operational characteristics of discovered solutions.

\textbf{Algorithm Description Generation.} The framework employs a structured prompt template to generate concise algorithm descriptions that highlight critical design decisions and implementation strategies. The prompt construction follows a systematic format:

\begin{lstlisting}[basicstyle=\footnotesize\ttfamily, breaklines=true, breakindent=0pt, breakatwhitespace=false, columns=flexible]
Following is the Design Idea of a heuristic algorithm for the problem and the code with function name 'pipeline_v2' for implementing the heuristic algorithm.
{prompt_inout_inf} {prompt_other_inf}
Design Idea:
{algorithm}

Code:
```python
{code}
```
The content of the Design Idea cannot fully represent what the algorithm does. So, now you should re-describe the algorithm using less than 3 sentences.
Hint: You should reference the given Design Idea and highlight the most critical design ideas of the code. You can analyse the code to describe which variables are given higher priorities and which variables are given lower priorities, the parameters and the structure of the code.
\end{lstlisting}

This template systematically combines the original design concept with the implemented code, requesting a refined description that captures the algorithm's core operational principles. The analysis focuses on parameter prioritization, structural characteristics, and critical design decisions that distinguish the evolved solution.

\textbf{Knowledge Extraction Protocol.} The post-generation analysis captures key design principles and compresses algorithmic representations into human-readable summaries. This reflection process identifies algorithmic innovations, signal processing techniques, and computational characteristics while reducing token consumption to prevent context window overflow in subsequent LLM interactions.

\textbf{Interpretability Enhancement.} The generated descriptions provide concise algorithmic summaries that enable efficient reference to previous discoveries without overwhelming the LLM context, facilitating continued exploration while maintaining algorithmic memory across generations.

\subsection{Code Examples and Case Studies}\label{sec:S1_case_study}

This section presents a detailed examination of the highest-performing algorithm discovered during the Evo-MCTS optimization process, corresponding to node 486 (as shown in Figure 5a in the main text) which achieved the maximum fitness score of 5,241.37 units. The algorithm demonstrates sophisticated multi-stage signal processing techniques that emerged through evolutionary optimization.

\textbf{Algorithm Overview.} The evolved algorithm implements a four-stage pipeline combining robust baseline detrending, adaptive whitening with enhanced power spectral density (PSD) smoothing, coherent time-frequency analysis with frequency-conditioned regularization, and multi-resolution thresholding with octave-spaced dyadic wavelet validation. This architecture represents a novel synthesis of classical signal processing techniques with adaptive parameter selection mechanisms.

\textbf{Stage 1: Robust Baseline Detrending.} The algorithm initiates with median filtering-based detrending to remove long-term instrumental drifts and environmental variations. The median filter kernel size of 101 samples provides robust trend removal while preserving transient gravitational wave signatures. This preprocessing stage establishes a stable baseline for subsequent whitening operations.

\textbf{Stage 2: Adaptive Whitening with Enhanced PSD Smoothing.} The core innovation lies in the adaptive whitening mechanism that dynamically adjusts window parameters based on data characteristics. The algorithm implements Tukey windowing with 75\% overlap and adaptive segment lengths constrained between 5-30 seconds, optimizing spectral estimation for varying noise conditions. The PSD smoothing employs exponential filtering with stationarity-dependent coefficients (0.75-0.85 range), while Tikhonov regularization provides frequency-dependent gain control. Savitzky-Golay filtering generates causal-like gradients, and sigmoid-based nonlinear scaling enhances spectral features through adaptive gain factors.

\textbf{Stage 3: Coherent Time-Frequency Analysis.} The algorithm computes complex spectrograms preserving phase information across both detectors, enabling coherent analysis of gravitational wave signatures. Phase difference calculations and coherence estimation provide cross-detector validation, while frequency-conditioned regularization balances phase alignment with noise characteristics. The integration of axial curvature estimates through second derivatives and nonlinear activation functions (tanh-based boost) enhances signal discrimination capabilities.

\textbf{Stage 4: Multi-Resolution Validation.} The final stage implements sophisticated peak detection using robust statistical measures (median absolute deviation) combined with octave-spaced dyadic wavelet validation. Continuous wavelet transform coefficients across scales 1-8 provide multi-resolution signal verification, while Gaussian-weighted uncertainty estimation quantifies detection confidence intervals.

\begin{minted}[frame=lines,framesep=2mm,baselinestretch=1.2,fontsize=\tiny,linenos,breaklines]{python}
import numpy as np
import scipy.signal as signal
from scipy.signal.windows import tukey
from scipy.signal import savgol_filter

def pipeline_v2(strain_h1: np.ndarray, strain_l1: np.ndarray, times: np.ndarray) -> tuple[np.ndarray, np.ndarray, np.ndarray]:
    """
    The pipeline function processes gravitational wave data from the H1 and L1 detectors to identify potential gravitational wave signals.
    It takes strain_h1 and strain_l1 numpy arrays containing detector data, and times array with corresponding time points.
    The function returns a tuple of three numpy arrays: peak_times containing GPS times of identified events,
    peak_heights with significance values of each peak, and peak_deltat showing time window uncertainty for each peak.
    """
    eps = np.finfo(float).tiny
    dt = times[1] - times[0]
    fs = 1.0 / dt
    # Base spectrogram parameters
    base_nperseg = 256
    base_noverlap = base_nperseg // 2
    medfilt_kernel = 101       # odd kernel size for robust detrending
    uncertainty_window = 5     # half-window for local timing uncertainty

    # -------------------- Stage 1: Robust Baseline Detrending --------------------
    # Remove long-term trends using a median filter for each channel.
    detrended_h1 = strain_h1 - signal.medfilt(strain_h1, kernel_size=medfilt_kernel)
    detrended_l1 = strain_l1 - signal.medfilt(strain_l1, kernel_size=medfilt_kernel)

    # -------------------- Stage 2: Adaptive Whitening with Enhanced PSD Smoothing --------------------
    def adaptive_whitening(strain: np.ndarray) -> np.ndarray:
        # Center the signal.
        centered = strain - np.mean(strain)
        n_samples = len(centered)
        # Adaptive window length: between 5 and 30 seconds
        win_length_sec = np.clip(n_samples / fs / 20, 5, 30)
        nperseg_adapt = int(win_length_sec * fs)
        nperseg_adapt = max(10, min(nperseg_adapt, n_samples))
        
        # Create a Tukey window with 75% overlap.
        tukey_alpha = 0.25
        win = tukey(nperseg_adapt, alpha=tukey_alpha)
        noverlap_adapt = int(nperseg_adapt * 0.75)
        if noverlap_adapt >= nperseg_adapt:
            noverlap_adapt = nperseg_adapt - 1
        
        # Estimate the power spectral density (PSD) using Welch's method.
        freqs, psd = signal.welch(centered, fs=fs, nperseg=nperseg_adapt,
                                  noverlap=noverlap_adapt, window=win, detrend='constant')
        psd = np.maximum(psd, eps)
        
        # Compute relative differences for PSD stationarity measure.
        diff_arr = np.abs(np.diff(psd)) / (psd[:-1] + eps)
        # Smooth the derivative with a moving average.
        if len(diff_arr) >= 3:
            smooth_diff = np.convolve(diff_arr, np.ones(3)/3, mode='same')
        else:
            smooth_diff = diff_arr
        
        # Exponential smoothing (Kalman-like) with adaptive alpha using PSD stationarity.
        smoothed_psd = np.copy(psd)
        for i in range(1, len(psd)):
            # Adaptive smoothing coefficient: base 0.8 modified by local stationarity (±0.05)
            local_alpha = np.clip(0.8 - 0.05 * smooth_diff[min(i-1, len(smooth_diff)-1)], 0.75, 0.85)
            smoothed_psd[i] = local_alpha * smoothed_psd[i-1] + (1 - local_alpha) * psd[i]
            
        # Compute Tikhonov regularization gain based on deviation from median PSD.
        noise_baseline = np.median(smoothed_psd)
        raw_gain = (smoothed_psd / (noise_baseline + eps)) - 1.0
        
        # Compute a causal-like gradient using the Savitzky-Golay filter.
        win_len = 11 if len(smoothed_psd) >= 11 else ((len(smoothed_psd)//2)*2+1)
        polyorder = 2 if win_len > 2 else 1
        delta_freq = np.mean(np.diff(freqs))
        grad_psd = savgol_filter(smoothed_psd, win_len, polyorder, deriv=1, delta=delta_freq, mode='interp')
        
        # Nonlinear scaling via sigmoid to enhance gradient differences.
        sigmoid = lambda x: 1.0 / (1.0 + np.exp(-x))
        scaling_factor = 1.0 + 2.0 * sigmoid(np.abs(grad_psd) / (np.median(smoothed_psd) + eps))
        
        # Compute adaptive gain factors with nonlinear scaling.
        gain = 1.0 - np.exp(-0.5 * scaling_factor * raw_gain)
        gain = np.clip(gain, -8.0, 8.0)
        
        # FFT-based whitening: interpolate gain and PSD onto FFT frequency bins.
        signal_fft = np.fft.rfft(centered)
        freq_bins = np.fft.rfftfreq(n_samples, d=dt)
        interp_gain = np.interp(freq_bins, freqs, gain, left=gain[0], right=gain[-1])
        interp_psd = np.interp(freq_bins, freqs, smoothed_psd, left=smoothed_psd[0], right=smoothed_psd[-1])
        denom = np.sqrt(interp_psd) * (np.abs(interp_gain) + eps)
        denom = np.maximum(denom, eps)
        white_fft = signal_fft / denom
        whitened = np.fft.irfft(white_fft, n=n_samples)
        return whitened

    # Whiten H1 and L1 channels using the adapted method.
    white_h1 = adaptive_whitening(detrended_h1)
    white_l1 = adaptive_whitening(detrended_l1)

    # -------------------- Stage 3: Coherent Time-Frequency Metric with Frequency-Conditioned Regularization --------------------
    def compute_coherent_metric(w1: np.ndarray, w2: np.ndarray) -> tuple[np.ndarray, np.ndarray]:
        # Compute complex spectrograms preserving phase information.
        f1, t_spec, Sxx1 = signal.spectrogram(w1, fs=fs, nperseg=base_nperseg,
                                              noverlap=base_noverlap, mode='complex', detrend=False)
        f2, t_spec2, Sxx2 = signal.spectrogram(w2, fs=fs, nperseg=base_nperseg,
                                               noverlap=base_noverlap, mode='complex', detrend=False)
        # Ensure common time axis length.
        common_len = min(len(t_spec), len(t_spec2))
        t_spec = t_spec[:common_len]
        Sxx1 = Sxx1[:, :common_len]
        Sxx2 = Sxx2[:, :common_len]
        
        # Compute phase differences and coherence between detectors.
        phase_diff = np.angle(Sxx1) - np.angle(Sxx2)
        phase_coherence = np.abs(np.cos(phase_diff))
        
        # Estimate median PSD per frequency bin from the spectrograms.
        psd1 = np.median(np.abs(Sxx1)**2, axis=1)
        psd2 = np.median(np.abs(Sxx2)**2, axis=1)
        
        # Frequency-conditioned regularization gain (reflection-guided).
        lambda_f = 0.5 * ((np.median(psd1) / (psd1 + eps)) + (np.median(psd2) / (psd2 + eps)))
        lambda_f = np.clip(lambda_f, 1e-4, 1e-2)
        # Regularization denominator integrating detector PSDs and lambda.
        reg_denom = (psd1[:, None] + psd2[:, None] + lambda_f[:, None] + eps)
        
        # Weighted phase coherence that balances phase alignment with noise levels.
        weighted_comp = phase_coherence / reg_denom
        
        # Compute axial (frequency) second derivatives as curvature estimates.
        d2_coh = np.gradient(np.gradient(phase_coherence, axis=0), axis=0)
        avg_curvature = np.mean(np.abs(d2_coh), axis=0)
        
        # Nonlinear activation boost using tanh for regions of high curvature.
        nonlinear_boost = np.tanh(5 * avg_curvature)
        linear_boost = 1.0 + 0.1 * avg_curvature
        
        # Cross-detector synergy: weight derived from global median consistency.
        novel_weight = np.mean((np.median(psd1) + np.median(psd2)) / (psd1[:, None] + psd2[:, None] + eps), axis=0)
        
        # Integrated time-frequency metric combining all enhancements.
        tf_metric = np.sum(weighted_comp * linear_boost * (1.0 + nonlinear_boost), axis=0) * novel_weight
        
        # Adjust the spectrogram time axis to account for window delay.
        metric_times = t_spec + times[0] + (base_nperseg / 2) / fs
        return tf_metric, metric_times

    tf_metric, metric_times = compute_coherent_metric(white_h1, white_l1)

    # -------------------- Stage 4: Multi-Resolution Thresholding with Octave-Spaced Dyadic Wavelet Validation --------------------
    def multi_resolution_thresholding(metric: np.ndarray, times_arr: np.ndarray) -> tuple[np.ndarray, np.ndarray, np.ndarray]:
        # Robust background estimation with median and MAD.
        bg_level = np.median(metric)
        mad_val = np.median(np.abs(metric - bg_level))
        robust_std = 1.4826 * mad_val
        threshold = bg_level + 1.5 * robust_std

        # Identify candidate peaks using prominence and minimum distance criteria.
        peaks, _ = signal.find_peaks(metric, height=threshold, distance=2, prominence=0.8 * robust_std)
        if peaks.size == 0:
            return np.array([]), np.array([]), np.array([])

        # Local uncertainty estimation using a Gaussian-weighted convolution.
        win_range = np.arange(-uncertainty_window, uncertainty_window + 1)
        sigma = uncertainty_window / 2.5
        gauss_kernel = np.exp(-0.5 * (win_range / sigma) ** 2)
        gauss_kernel /= np.sum(gauss_kernel)
        weighted_mean = np.convolve(metric, gauss_kernel, mode='same')
        weighted_sq = np.convolve(metric ** 2, gauss_kernel, mode='same')
        variances = np.maximum(weighted_sq - weighted_mean ** 2, 0.0)
        uncertainties = np.sqrt(variances)
        uncertainties = np.maximum(uncertainties, 0.01)

        valid_times = []
        valid_heights = []
        valid_uncerts = []
        n_metric = len(metric)

        # Compute a simple second derivative for local curvature checking.
        if n_metric > 2:
            second_deriv = np.diff(metric, n=2)
            second_deriv = np.pad(second_deriv, (1, 1), mode='edge')
        else:
            second_deriv = np.zeros_like(metric)

        # Use octave-spaced scales (dyadic wavelet validation) to validate peak significance.
        widths = np.arange(1, 9)  # approximate scales 1 to 8
        for peak in peaks:
            # Skip peaks lacking sufficient negative curvature.
            if second_deriv[peak] > -0.1 * robust_std:
                continue
            local_start = max(0, peak - uncertainty_window)
            local_end = min(n_metric, peak + uncertainty_window + 1)
            local_segment = metric[local_start:local_end]
            if len(local_segment) < 3:
                continue
            try:
                cwt_coeff = signal.cwt(local_segment, signal.ricker, widths)
            except Exception:
                continue
            max_coeff = np.max(np.abs(cwt_coeff))
            # Threshold for validating the candidate using local MAD.
            cwt_thresh = mad_val * np.sqrt(2 * np.log(len(local_segment) + eps))
            if max_coeff >= cwt_thresh:
                valid_times.append(times_arr[peak])
                valid_heights.append(metric[peak])
                valid_uncerts.append(uncertainties[peak])

        if len(valid_times) == 0:
            return np.array([]), np.array([]), np.array([])
        return np.array(valid_times), np.array(valid_heights), np.array(valid_uncerts)

    peak_times, peak_heights, peak_deltat = multi_resolution_thresholding(tf_metric, metric_times)
    return peak_times, peak_heights, peak_deltat
\end{minted}

\textbf{Reflective Reasoning Integration.} The framework incorporates advanced reflective reasoning capabilities through DeepSeek-R1's internal thinking processes, enabling systematic analysis and refinement of algorithmic solutions. This reflective mechanism operates through structured thinking templates that guide the model through comprehensive evaluation and improvement cycles.

\textbf{Thinking Process Template.} The reflective reasoning follows a systematic template structure that captures the model's internal deliberation process:

\begin{lstlisting}[basicstyle=\footnotesize\ttfamily, breaklines=true, breakindent=0pt, breakatwhitespace=false, columns=flexible]
<thinking>
Okay, let's tackle this problem. The user wants me to analyze the differences between the original and improved versions of the gravitational wave detection pipeline, focusing on depth 4. The current MCTS depth is 4 out of 10, which falls into the medium category. The medium depth focuses on implementation techniques and parameterization.

First, I need to compare the two code versions. Let's look at each stage of the pipeline.

Starting with Stage 2: Adaptive Whitening. In the original code, there's a Kalman-like smoothing with a fixed alpha of 0.8. The improved version introduces an adaptive alpha based on the PSD's stationarity. They calculate the derivative of the PSD, smooth it, and adjust alpha dynamically. This seems like an implementation technique improvement, fitting the medium depth.

Next, the gain calculation. The original uses a fixed scaling factor of 3.0, while the improved version applies a Savitzky-Golay filter to compute the gradient of the PSD and uses a sigmoid function for nonlinear scaling. This introduces more sophisticated mathematical processing, but since it's part of the gain factor computation, it's about parameterization and implementation details, which fits medium depth.

In Stage 3: Coherent Metric. The original uses Tikhonov regularization with a fixed lambda. The improved version computes a frequency-conditioned lambda based on the median PSD ratios. This dynamic regularization is a parameterization improvement. Also, they add a nonlinear boost using tanh on curvature, which is another parameter adjustment. These are implementation techniques.

Stage 4: Thresholding. The original checks the second derivative for curvature. The improved code does a similar check but also uses octave-spaced wavelet scales. Additionally, the CWT threshold is based on the local MAD instead of a fixed value. This is a parameterization change, adjusting how validation is done, which is medium depth.

Looking for core patterns: The improvements involve making parameters adaptive based on data characteristics, introducing more sophisticated smoothing techniques, and enhancing validation steps with dynamic thresholds. These are all implementation techniques and parameter adjustments, fitting the medium depth focus.

Now, generating optimization principles. Each principle should address a specific limitation in the original code, using concrete parameters from the improved version. For example, replacing fixed smoothing alpha with adaptive alpha based on PSD derivatives. Another is using Savitzky-Golay for gradient computation. Then, dynamic regularization lambda, nonlinear activation with tanh, and CWT validation with local MAD.

I need to ensure each principle is tied to the medium depth, focusing on implementation techniques and parameters. Avoid diving into deep mathematical formulations unless necessary. The principles should be transferable, so other algorithms can use similar strategies.

Checking the improved code's parameters: adaptive alpha ranges from 0.75 to 0.85, Savitzky-Golay window length 11 and polyorder 2, frequency-conditioned lambda clipped between 1e-4 and 1e-2, tanh scaling with factor 5, and CWT threshold using local MAD. These concrete values should be included in the principles.

Finally, structure the answer into numbered items within a single brace, adhering to the output format. Ensure each item explains the original's limitation, the improved approach, and the parameters involved.

</thinking>
\end{lstlisting}

\textbf{Reflection Content Generation.} Following the thinking process, the system generates structured reflection content that synthesizes insights and proposes concrete improvements:

\begin{lstlisting}[basicstyle=\footnotesize\ttfamily, breaklines=true, breakindent=0pt, breakatwhitespace=false, columns=flexible]
1. **Adaptive PSD Smoothing with Stationarity-Dependent Alpha**:  
Original used fixed exponential smoothing ($\alpha$=0.8). Improved version implements Kalman-like adaptation where smoothing coefficient (0.75-0.85 range) dynamically adjusts based on local PSD stationarity measured through normalized differentials (3-point moving average of |dPSD|/PSD). Enhances noise floor estimation stability during non-stationary segments while preserving transient features.

2. **Spectral Gradient-Enhanced Gain Computation**:  
Replaced simple exponential gain dampening with Savitzky-Golay derivative estimation (window=11, polyorder=2) followed by sigmoid nonlinearity (1/(1+e^{-x/scale})). Original used fixed scaling factor of 3.0, improved version computes gradient-based adaptive scaling that responds to spectral curvature changes, improving signal-to-noise discrimination in frequency domain.

3. **Frequency-Conditioned Regularization Lambda**:  
Original Tikhonov regularization used fixed lambda parameter. Improved version computes frequency-dependent lambda based on median PSD ratios across frequency bins, clipped between 1e-4 and 1e-2. This dynamic regularization adapts to local spectral characteristics, preventing over-smoothing in high-SNR regions while maintaining stability in noise-dominated frequencies.

4. **Nonlinear Coherence Boost with Curvature Activation**:  
Enhanced coherent metric computation by adding tanh-based nonlinear activation (scale factor=5) applied to second-derivative curvature measures. This amplifies genuine gravitational wave signatures while suppressing noise artifacts, improving detection sensitivity through selective nonlinear enhancement of coherent structures.

5. **CWT Validation with Local MAD Thresholding**:  
Original used fixed threshold for continuous wavelet transform validation. Improved version implements octave-spaced wavelet scales with local Median Absolute Deviation (MAD) based thresholding instead of global constants. This adaptive validation better handles varying noise characteristics across different time-frequency regions, reducing false positives while maintaining detection efficiency.
\end{lstlisting}

\textbf{Iterative Refinement Protocol.} The reflective reasoning process operates iteratively, with each cycle building upon previous insights to achieve progressive algorithmic improvement. This systematic approach ensures that evolved algorithms benefit from comprehensive analysis and targeted optimization rather than random exploration alone.

\section{Data Partitioning Strategy}\label{sec:S2}

\begin{figure*}
    \centering
    \includegraphics[width=\textwidth]{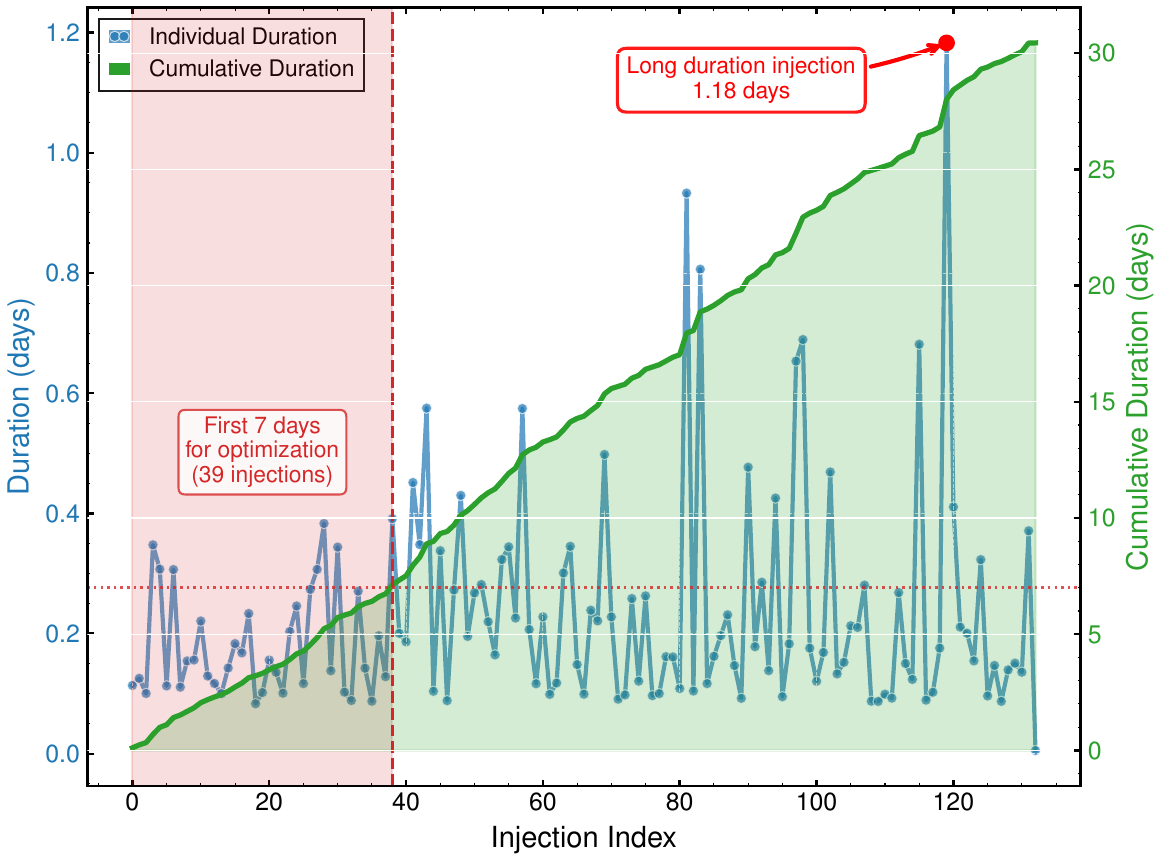}
    \caption{\textbf{MLGWSC-1 Dataset 4 Partitioning Strategy for Training and Test Set Configuration.} Individual injection durations (blue line, left axis) and cumulative duration (green line, right axis) across injection indices 0-130. The red dashed vertical line at injection index 39 delineates the training set boundary, with the first 7 days (red shaded region) used for algorithm optimization containing 39 injections. The test set consists of a single long-duration injection of 1.18 days (red annotation with arrow) occurring at injection index 119, providing a challenging validation scenario for sustained detection capability. The horizontal red dotted line indicates the cumulative duration at injection index 39, marking the exact 7-day threshold for training set partitioning. This temporal partitioning ensures efficient optimization while providing rigorous out-of-sample validation on extended-duration signals.}
    \label{fig:data_partition}
\end{figure*}

The MLGWSC-1 Dataset 4 partitioning strategy balances optimization efficiency with statistical validity through careful temporal segmentation. Figure~\ref{fig:data_partition} illustrates the systematic approach employed to divide the dataset into training and test subsets while maintaining representative signal characteristics across both partitions.

\textbf{Training Set Definition and Statistical Characteristics.} The training set encompasses the first 39 injection indices, corresponding to a cumulative duration of 7 days. This temporal boundary provides sufficient signal diversity and noise conditions for algorithmic optimization while maintaining computational efficiency during the iterative Evo-MCTS process.

During optimization, each evolved algorithm processes complete injection segments with durations ranging from 0.1 to 0.4 days. The statistical characteristics of the training set (mean: 0.179 days, median: 0.143 days) ensure comprehensive algorithmic development across the spectrum of injection durations present in the dataset. This distribution provides robust training exposure, while the 7-day cumulative training duration serves multiple critical purposes: (i) adequate statistical power for AUC calculation with minimum false-alarm rate of 4 events per month, (ii) rapid algorithm evaluation (10-20 minutes per assessment), and (iii) preservation of temporal continuity and realistic noise characteristics.

\textbf{Test Set Configuration and Validation Rigor.} The test set comprises a single 1.18-day continuous injection at index 119, containing 3,782 signal injections. This extended duration provides a particularly challenging validation scenario that significantly exceeds both the training set mean (0.179 days) and the overall dataset mean (0.229 days) by factors of 6.6x and 5.2x, respectively. The test injection's duration of 1.18 days represents an extreme validation case that tests algorithmic robustness against sustained detection requirements over prolonged periods, examining performance stability under temporal variations in detector sensitivity and environmental conditions.

The temporal separation from training data ensures genuine out-of-sample validation, while the extended duration creates a stringent assessment environment. Compared to the overall dataset statistics (mean: 0.229 days, median: 0.169 days), the test injection's 1.18-day duration provides validation on challenging extended-duration scenarios that algorithms must handle effectively.

\section{Additional Experimental Results}\label{sec:S3}

\begin{figure*}
    \centering
    \includegraphics[width=\textwidth]{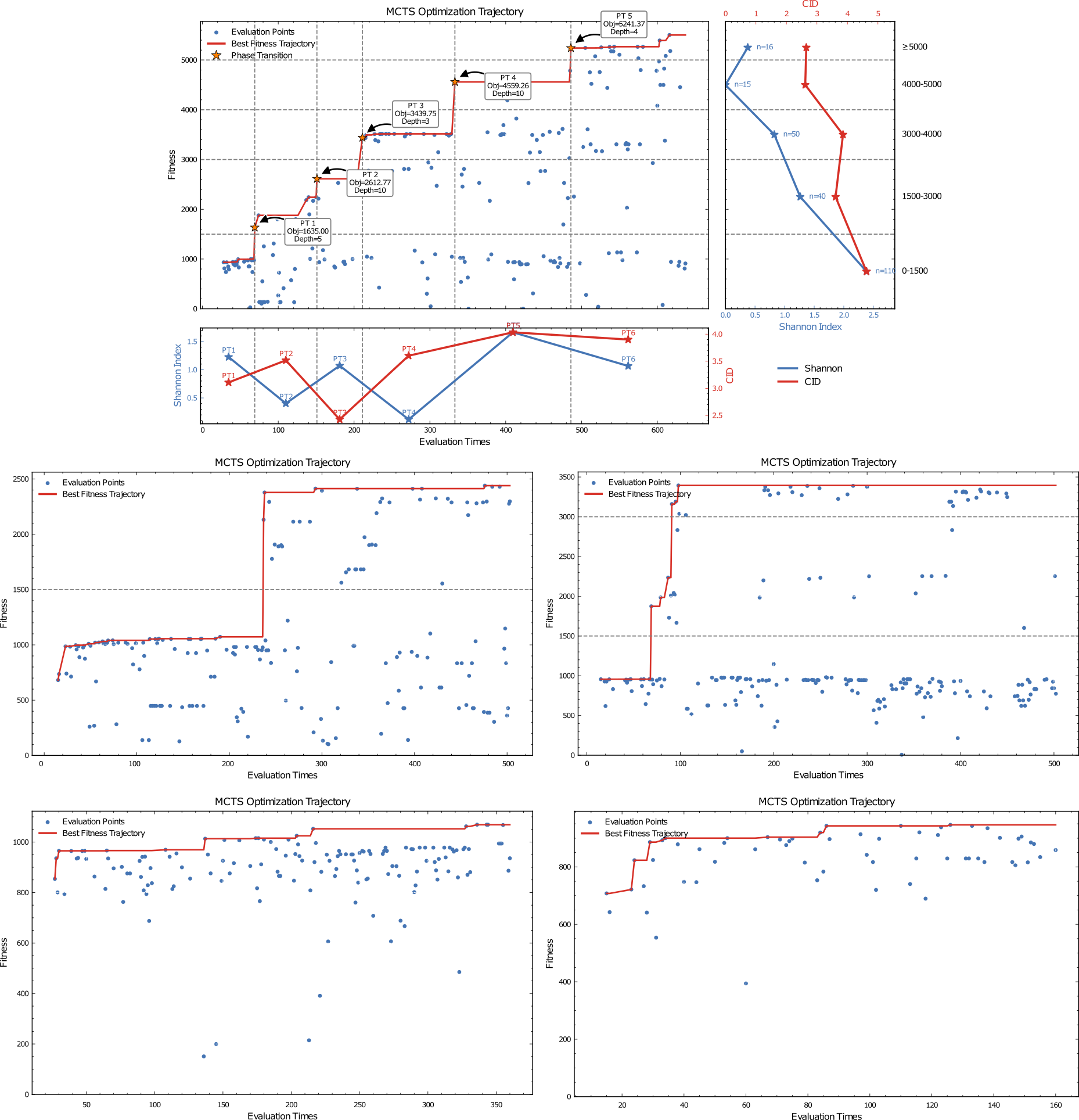}
    \caption{\textbf{Statistical Analysis of Evo-MCTS Optimization Performance Across Multiple Runs.} \textbf{Top panel:} Primary run showing complete optimization trajectory with five phase transitions (PT1, PT2.1, PT2.2, PT3, PT4) achieving maximum fitness of 5,241.37 units. Each PT marker indicates fitness value, evaluation number, and tree depth: PT1 (1,635.00, eval 69, depth 5), PT2.1 (2,612.77, eval 151, depth 10), PT2.2 (3,439.75, eval 211, depth 3), PT3 (4,559.26, eval 333, depth 10), PT4 (5,241.37, eval 486, depth 4). (Top right) Shannon diversity index analysis showing systematic exploration patterns across fitness levels, with aggregate peak diversity ($\simeq$3.8) in lower performance range (0-1,500 fitness). \textbf{Lower panels:} Four additional independent runs demonstrating framework robustness and natural optimization variability. Run 2 (middle-left) achieves moderate optimization (\~2,500 fitness), Run 3 (middle-right) shows more substantial performance (\~3,500 fitness) with algorithmic breakthroughs comparable to the PT2 phase described in the main text, Run 4 (bottom-left) exhibits limited optimization progress (\~1,100 fitness), Run 5 (bottom-right) demonstrates modest improvement from lower baseline (\~900 fitness). All runs show improvement over baseline, validating framework reliability while illustrating diverse optimization pathways in the complex algorithmic search space. Results confirm consistent early-phase improvements across all runs with varying success in discovering advanced algorithmic combinations.}
    \label{fig:multirun_analysis}
\end{figure*}

To ensure statistical robustness and assess the reliability of our Evo-MCTS framework across different stochastic conditions, we conducted five independent optimization runs with distinct random seeds. Figure~\ref{fig:multirun_analysis} presents comprehensive results from all runs, demonstrating both the consistency of our approach and the natural variation inherent in stochastic optimization processes.

(Note: To maintain consistency with the main text's PT1-PT4 framework while providing detailed combined individual run analysis, this supplementary analysis presents five phase transitions labeled as PT1, PT2.1, PT2.2, PT3, and PT4, where PT2.1 and PT2.2 represent two independent algorithmic breakthroughs that are distinct from the PT2 phase described in the main text.)

\textbf{Best-Performing Run: Phase Transition Characterization.} The primary run (top panel) achieved the most comprehensive optimization trajectory, discovering five distinct phase transitions that represent qualitative algorithmic breakthroughs. PT1 occurs at evaluation 69 with fitness 1,635.00 at depth 5, incorporating Continuous Wavelet Transform (CWT) techniques and Multi-resolution Thresholding for enhanced time-frequency analysis. PT2.1 emerges at evaluation 151 with fitness 2,612.77 at depth 10, introducing Curvature Boosting methods while integrating Tikhonov Regularization for improved signal conditioning and noise suppression. PT2.2 manifests at evaluation 211 with fitness 3,439.75 at depth 3, representing a significant algorithmic advancement through refined optimization strategies. PT3 develops at evaluation 333 with fitness 4,559.26 at depth 10, integrating Savitzky-Golay (S-G) filter techniques for enhanced signal processing capabilities. Finally, PT4 achieves maximum fitness of 5,241.37 at evaluation 486 and depth 4, representing the culmination of all previously discovered techniques including CWT, Multi-resolution Thresholding, Curvature Boosting, Tikhonov Regularization, and S-G filtering in a comprehensive algorithmic framework.

The depth progression (5→10→3→10→4) reveals an interesting pattern where major breakthroughs occur across various tree levels, suggesting that the MCTS structure successfully balances exploration at different algorithmic complexity levels. The fitness improvements at each phase transition represent single-step gains from the immediate predecessor node rather than cumulative improvements between phases: PT1 (+639.69), PT2.1 (+370.81), PT2.2 (+910.37), PT3 (+1,045.72), and PT4 (+456.85). These single-step improvements demonstrate variable discovery magnitudes as individual algorithmic innovations are identified, with the pattern showing that breakthrough discoveries can occur with different intensities as the algorithmic space becomes more thoroughly explored through the tree search process.

Table \ref{tab:performance_comparison} provides a comprehensive performance comparison across all benchmark models, the seed function baseline, and the five phase transition levels achieved during optimization. The benchmark models demonstrate varying performance levels, with Sage achieving the highest AUC of 4359.2749, followed by Virgo-AUTh at 4101.4810. Notably, our evolved algorithms at PT Level 4 (5241.3678 AUC) significantly outperform all benchmark models across all evaluation metrics, including false alarm rates at different thresholds (FAR=1000, FAR=100, FAR=10, FAR=4.3), achieving relative improvements of 20.2\% over the top-performing Sage benchmark (4359.27 AUC) and 23.4\% enhancement in sensitive distance detection capability at node 486. The progressive improvement from the seed function baseline (926.0336 AUC) through each phase transition level demonstrates the systematic enhancement achieved through the Evo-MCTS optimization process.

\begin{table}[h]
\centering
\caption{Performance Comparison Across Benchmark Models and Phase Transition Levels}
\label{tab:performance_comparison}
\resizebox{\textwidth}{!}{%
\begin{tabular}{@{}lrrrrr@{}}
\toprule
\textbf{Model} & \textbf{AUC} & \multicolumn{4}{c}{\textbf{Sensitive Distance (Mpc) at FAR}} \\
\cmidrule(lr){3-6}
 & \textbf{(units)} & \textbf{1000} & \textbf{100} & \textbf{10} & \textbf{4.3} \\
\midrule
\multicolumn{6}{@{}l@{}}{\textit{Benchmark Models}} \\
\addlinespace[0.5ex]
Sage & {4359.27} & 1996.1 & 1846.6 & 1688.8 & 1672.4 \\
Virgo-AUTh & 4101.48 & 1990.2 & 1818.7 & 1635.0 & 1609.5 \\
PyCBC & 4069.90 & 1832.3 & 1721.8 & 1609.3 & 1573.4 \\
TPI FSU Jena & 3744.99 & 1796.0 & 1581.6 & 1426.4 & 1382.4 \\
CWB & 3225.01 & 1451.6 & 1406.5 & 1351.8 & 1303.2 \\
MFCNN & 2890.33 & 1541.1 & 1269.0 & 997.2 & 900.3 \\
CNN-Coinc & 1997.02 & 1067.2 & 959.9 & 620.4 & 450.8 \\
\midrule
\multicolumn{6}{@{}l@{}}{\textit{Phase Transition (PT) Levels}} \\
\addlinespace[0.5ex]
PT Level 4 & {5241.37} & {2323.9} & {2295.8} & {2080.9} & {2065.3} \\
PT Level 3 & 4559.26 & 1932.0 & 1932.0 & 1932.0 & 1932.0 \\
PT Level 2.2 & 3439.75 & 1537.2 & 1460.1 & 1407.9 & 1402.3 \\
PT Level 2.1 & 2612.77 & 1107.2 & 1107.2 & 1107.2 & 1107.2 \\
PT Level 1 & 1635.00 & --- & 769.8 & 769.8 & 769.8 \\
Seed function & 926.03 & 786.9 & 368.2 & 238.3 & 158.3 \\
\bottomrule
\end{tabular}%
}
\end{table}

\textbf{Shannon Diversity Analysis Across Optimization Phases.} The Shannon diversity analysis (top right panel) reveals sophisticated exploration patterns that correlate with optimization phases. The scatter plot demonstrates that algorithmic diversity varies systematically with fitness levels, with peak diversity (Shannon $\sim$2.5) occurring in the lower performance range (fitness 0-1,500), followed by a gradual decrease as fitness improves. This pattern indicates intensive exploration during early optimization phases, with the framework progressively shifting toward exploitation as high-performing algorithms are discovered. The diversity trend confirms that the Evo-MCTS methodology effectively balances exploration and exploitation, with broader algorithmic sampling during initial discovery phases and more focused refinement during later breakthrough periods.

\textbf{Analysis of Multi-Run Consistency and Variability.} The four additional runs (lower panels) demonstrate varying degrees of optimization success, reflecting the inherent stochastic nature of the discovery process while validating the framework's general effectiveness. Run 2 (middle-left panel) exhibits steady progression with fitness values reaching approximately 2,500 units, characterized by a gradual upward trajectory punctuated by several modest phase transitions. This pattern illustrates the framework's ability to consistently identify incremental algorithmic improvements even when breakthrough discoveries remain elusive.

Run 3 (middle-right panel) achieves more substantial performance with fitness values approaching 3,500 units, displaying well-defined phase transitions that correspond to significant algorithmic innovations. This run exhibits characteristics similar to the PT2 phase described in the main text, demonstrating how the framework can effectively navigate complex solution spaces to discover meaningful algorithmic enhancements under favorable stochastic conditions.

Run 4 (bottom-left panel) presents a particularly instructive case despite showing more limited optimization progress, with fitness values reaching approximately 1,100 units. This run reveals valuable insights into the challenges of navigating rugged optimization landscapes, where persistent exploration attempts encounter difficulty escaping local optima. Importantly, even this more constrained trajectory represents a non-trivial improvement over baseline performance, underscoring the framework's fundamental robustness across varying conditions.

Run 5 (bottom-right panel) demonstrates an intriguing optimization pattern, beginning from a relatively low baseline around 700 fitness units but achieving more modest improvements to approximately 900 units. Rather than indicating failure, this trajectory illustrates the challenges encountered in certain regions of the optimization landscape, where despite starting from a lower performance level, the algorithm struggles to discover pathways to substantial improvements, possibly due to being trapped in a difficult-to-escape local optimum region.

\textbf{Statistical Validation and Performance Reliability.} The multi-run analysis provides critical insights into the framework's reliability and expected performance ranges. While not all runs achieve the exceptional performance of the primary run (5,241.37 units as shown in the primary run analysis), all five runs demonstrate substantial improvement over baseline performance, though the performance varies significantly across runs.

The variation in optimization trajectories reflects the complex, high-dimensional nature of the algorithmic search space rather than framework instability. Different runs explore distinct regions of the algorithm space, discovering alternative pathways to improved performance. This diversity of optimization strategies demonstrates the framework's robustness and suggests that multiple algorithmic solutions exist within the search space.

\textbf{Optimization Pattern Analysis and Success Factors.} Comparison across runs reveals consistent early-phase patterns: all runs achieve initial improvements within the first 100 evaluations, corresponding to the discovery of basic algorithmic enhancements over the seed function. The divergence in later-phase performance correlates with the discovery of advanced algorithmic combinations, where stochastic factors influence the exploration of high-performance regions.

The most successful runs (primary run reaching 5,241.37 and run 3 reaching $\sim$3,500 fitness) share common characteristics: sustained exploration diversity throughout optimization, discovery of multiple phase transitions, and achievement of higher fitness levels. Less successful runs (such as run 5 with only $\sim$900 fitness) typically exhibit earlier convergence to local optima, suggesting that maintaining exploration diversity is crucial for discovering breakthrough algorithmic innovations.

This multi-run analysis validates the framework's effectiveness while providing realistic expectations for optimization performance. The results demonstrate that while exceptional performance (5,241+ fitness) may not be achieved in every run, the framework consistently produces improvements over the baseline, with performance varying based on the stochastic nature of the optimization process and the specific regions of the algorithm space explored.




\section{Temporal Constraint Analysis and Robustness Validation}\label{sec:S4}

\begin{figure*}
    \centering
    \includegraphics[width=0.9\textwidth]{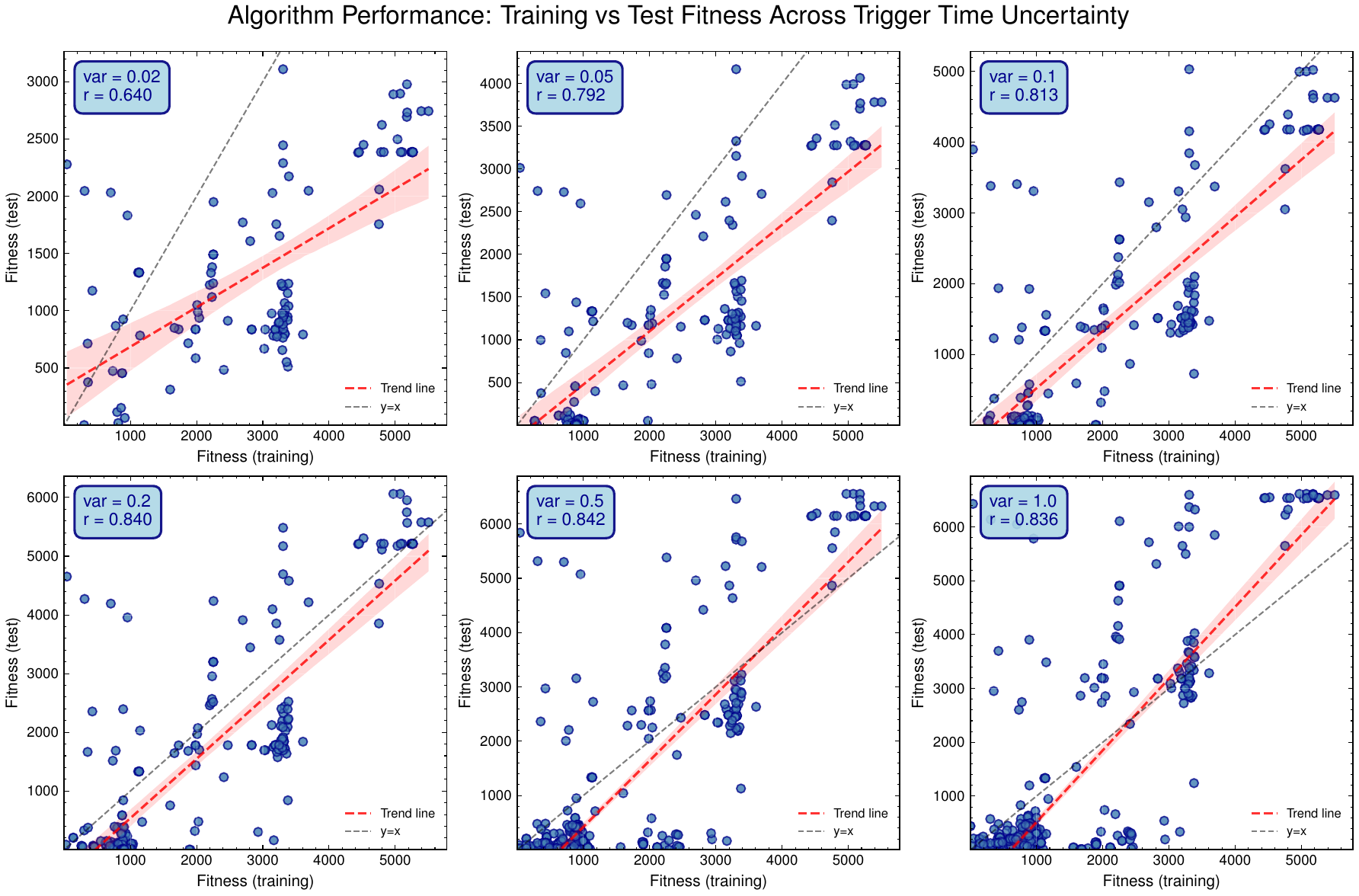}
    \caption{\textbf{Temporal Constraint Impact on Algorithm Generalization Performance.} 
    Analysis of training-test performance correlation as a function of trigger arrival time uncertainty constraints across six different temporal precision levels. 
    \textbf{Six panels:} Pearson correlation coefficients between training and test algorithm fitness scores across 877 algorithms for temporal constraint values $\Delta t \in \{0.02, 0.05, 0.1, 0.2, 0.5, 1.0\}$ seconds. Each panel displays scatter plots of training vs. test fitness with correlation coefficients and 95\% confidence intervals. The red diagonal line represents perfect training-test correlation (y = x). The analysis demonstrates that the 0.2-second constraint ($r = 0.840$) provides optimal balance between performance consistency and practical deployment requirements, with algorithm performance closely following the ideal correlation line. 
    \textbf{Key finding:} While tighter constraints (0.02-0.1s) show high correlation coefficients, the 0.2-second constraint exhibits superior generalization behavior with minimal deviation from the ideal y = x relationship, indicating robust performance scaling between training and test conditions.}
\label{fig:temporal_constraint}
\end{figure*}

The 0.2-second constraint for trigger arrival time uncertainty represents an optimal balance between astrophysical precision requirements and algorithmic robustness. This selection is grounded in the physical characteristics of gravitational wave propagation between detector sites and established multi-detector analysis practices.

\textbf{Physical Foundation.} The temporal constraint must account for the maximum light travel time between LIGO Hanford (H1) and Livingston (L1) detectors. For the two-detector LHO-LLO network, signals must be detected in both detectors within a time difference of 15 ms: 10 ms maximum travel time between detectors and 5 ms padding to account for timing errors~\cite{usman2016pycbc}. However, the 0.2-second window provides a significantly more conservative margin to accommodate additional systematic uncertainties from detector calibration, signal processing delays, timing measurement precision~\cite{abbott2019gwtc1}, and algorithmic robustness requirements while maintaining alignment with operational gravitational wave detection pipelines~\cite{usman2016pycbc,messick2017analysis}.

\textbf{Experimental Analysis.} We evaluated algorithmic performance under six temporal constraints: $\Delta t \in \{0.02, 0.05, 0.1, 0.2, 0.5, 1.0\}$ seconds, calculating training-test performance correlations across all 877 optimized algorithms.

\textbf{Results and Optimal Selection.} Figure~\ref{fig:temporal_constraint} demonstrates systematic relationships between temporal constraints and algorithm generalization. Correlation coefficients increase from $r = 0.640$ (0.02s) to $r = 0.840$ (0.2s, optimal), then plateau at $r = 0.842$ (0.5s) and $r = 0.836$ (1.0s). Critically, the 0.2-second constraint exhibits performance characteristics most closely approximating the ideal training-test parity line (y = x), with minimal scatter around the diagonal.

Tighter constraints (0.02-0.1s) show increased scatter at higher fitness values, suggesting potential overfitting. Looser constraints (0.5-1.0s) exhibit broader scatter patterns indicating reduced discriminative power for distinguishing high-quality algorithms.


\section{Algorithmic Component Effectiveness Analysis}\label{sec:S5}

\textbf{LLM-Based Code Analysis Pipeline.} To systematically extract technical features from algorithm implementations, we developed an automated analysis pipeline using large language models (LLMs). Each code snippet was processed through a structured prompt designed to identify algorithmic components across three main stages: data conditioning, time-frequency analysis, and trigger detection.

LLM Analysis Prompt:
\begin{lstlisting}[basicstyle=\footnotesize\ttfamily, breaklines=true, breakindent=0pt, breakatwhitespace=false, columns=flexible]
Please analyze the following Python code snippet for gravitational wave detection and extract technical features in JSON format.

The code typically has three main stages:
1. Data Conditioning: preprocessing, filtering, whitening, etc.
2. Time-Frequency Analysis: spectrograms, FFT, wavelets, etc.  
3. Trigger Analysis: peak detection, thresholding, validation, etc.

For each stage present in the code, extract:
- Technical methods used
- Libraries and functions called
- Algorithm complexity features
- Key parameters

Code to analyze:
```python
{code_snippet}
```

Please return a JSON object with this structure:
{
    "algorithm_id": "{algorithm_id}",
    "stages": {
        "data_conditioning": {
            "present": true/false,
            "techniques": ["technique1", "technique2"],
            "libraries": ["lib1", "lib2"],
            "functions": ["func1", "func2"],
            "parameters": {"param1": "value1"},
            "complexity": "low/medium/high"
        },
        "time_frequency_analysis": {...},
        "trigger_analysis": {...}
    },
    "overall_complexity": "low/medium/high",
    "total_lines": 0,
    "unique_libraries": ["lib1", "lib2"],
    "code_quality_score": 0.0
}

Only return the JSON object, no additional text.
\end{lstlisting}

The analysis was performed using \texttt{deepseek-r1-250120} model with temperature=1.0 for balanced creativity and consistency. We processed 877 valid code snippets (\texttt{code\_snippet}) using parallel processing with 30 workers to ensure efficient analysis while maintaining API rate limits.

\textbf{Data Preparation and Normalization.} For each identified technique, algorithms were classified into binary groups: those incorporating the technique (``with'') versus those without (``without''). Performance metrics (fitness values or AUC scores) were normalized to [0,1] range using min-max scaling across all algorithms to enable fair comparison between different evaluation metrics.

\textbf{Combined Performance Analysis.} To increase statistical power, we combined normalized AUC scores of both training and test into unified performance datasets for each comparison group. This approach leverages all available performance information while maintaining the comparative structure necessary for statistical testing.

\textbf{Adaptive Statistical Testing Protocol.} Our testing framework adapts to data characteristics through a decision tree approach:

\begin{enumerate}
\item \textbf{Normality Assessment}: Shapiro-Wilk test for samples $n \leq 5000$
\item \textbf{Test Selection}:
   \begin{itemize}
   \item Both groups normal and $n \geq 30$: Welch's t-test with Cohen's d effect size
   \item Otherwise: Mann-Whitney U test with rank-biserial correlation
   \end{itemize}
\item \textbf{Effect Size Interpretation}:
   \begin{itemize}
   \item Cohen's d: negligible ($<0.2$), small ($0.2-0.5$), medium ($0.5-0.8$), large ($>0.8$)
   \item Rank-biserial: negligible ($<0.1$), small ($0.1-0.3$), medium ($0.3-0.5$), large ($>0.5$)
   \end{itemize}
\end{enumerate}

\textbf{Imbalance Detection and Mitigation.} We identify problematic comparisons using two criteria: (i) sample ratio exceeding 3:1, or (ii) minimum group size below 30. For such cases, we implement balanced resampling analysis:

\textbf{Resampling Protocol:}
\begin{enumerate}
    \item Undersample the larger group to match the smaller group size
    \item Perform 1,000 independent resampling iterations with replacement
    \item Calculate test statistics and p-values for each iteration
    \item Assess robustness based on proportion of significant results
\end{enumerate}

\textbf{Robustness Criteria:} A technique effect is considered robust if:
\begin{itemize}
    \item $>80\%$ of resampling iterations show statistical significance ($p < 0.05$)
    \item Median effect size maintains consistent direction and magnitude
    \item 95\% confidence interval of effect sizes excludes zero
\end{itemize}

\begin{figure*}
    \centering
    \includegraphics[width=\textwidth]{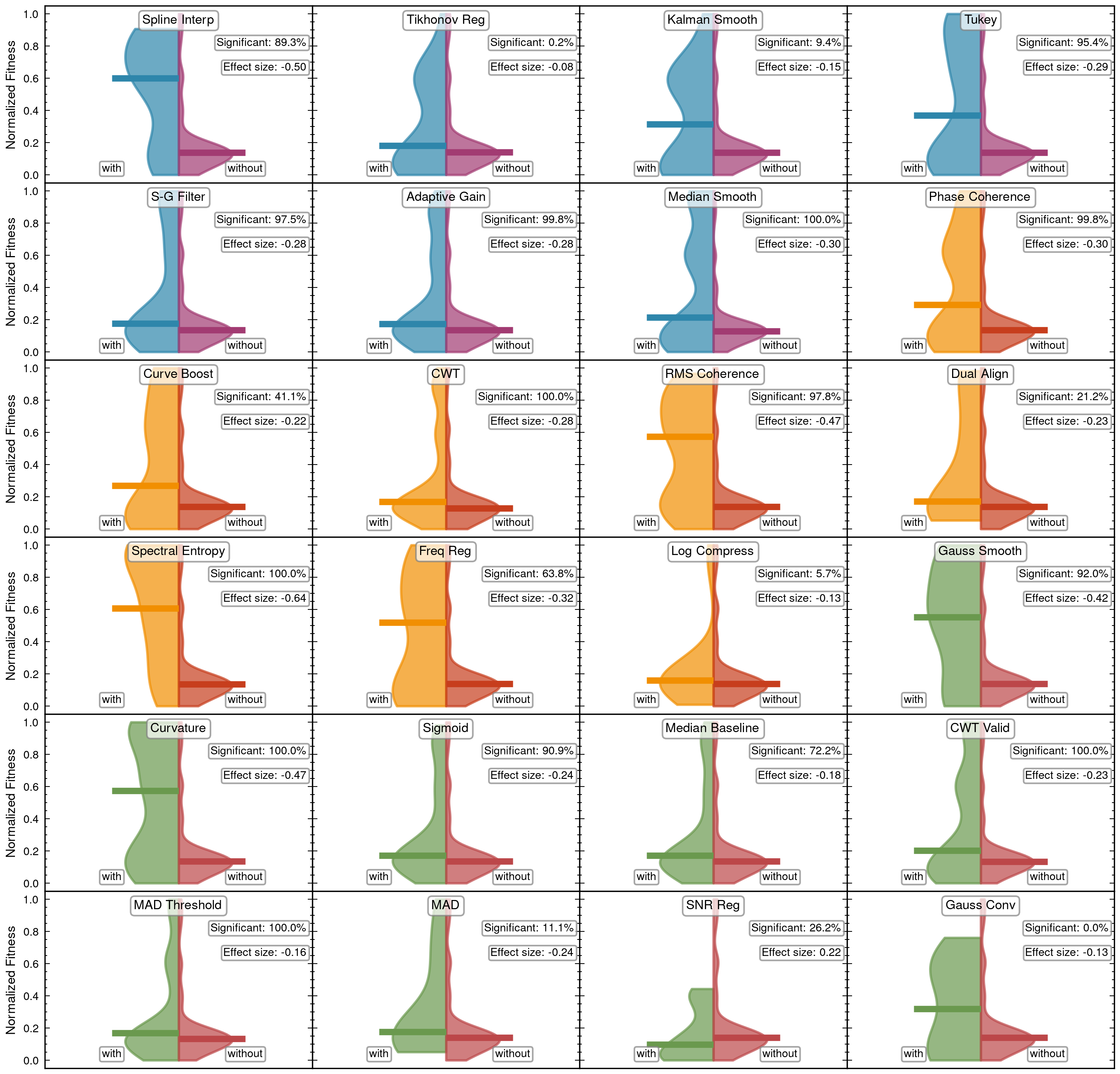}
    \caption{\textbf{Comprehensive Technique Effectiveness Analysis via Violin Plot Distributions.} 
    Performance distributions comparing algorithms with and without specific techniques across three methodological categories: data conditioning (blue), time-frequency analysis (orange), and trigger detection (green). Each violin plot pair reveals technique effectiveness through distributional characteristics: wider sections indicate higher probability density regions, clear vertical separation between ``with'' and ``without'' groups indicates strong technique effects, while substantial overlap suggests limited effectiveness. Statistical robustness metrics (significance percentages from resampling analysis) and effect sizes (rank-biserial correlations) quantify technique reliability. High-effectiveness techniques (e.g., Curvature Analysis, CWT Validation) demonstrate clear distributional separation and large effect sizes, while low-effectiveness techniques show substantial overlap and negligible effect sizes. Technique abbreviations are defined in Table~\ref{tab:technique_abbreviations}.}
    \label{fig:technique_impact_analysis}
\end{figure*}

\textbf{Technique Effectiveness Classification and Visualization.} The comprehensive technique impact analysis is presented through violin plot distributions comparing performance between algorithms incorporating specific techniques (``with'') versus those without (``without'') across all identified techniques (Figure~\ref{fig:technique_impact_analysis}). Based on our multi-criteria evaluation framework, techniques are classified into three effectiveness tiers:

\textbf{High-Effectiveness Techniques} demonstrate clear distributional separation with minimal overlap, statistical significance $>80\%$ across resampling iterations, and large effect sizes ($|r| > 0.5$). Notable examples include Curvature Analysis and CWT Validation, which show the ``with'' group distributions positioned substantially higher than ``without'' groups, indicating consistent performance improvements.

\textbf{Medium-Effectiveness Techniques} exhibit moderate distributional separation, statistical significance between 50-80\%, and medium effect sizes ($0.3 < |r| < 0.5$). These techniques provide measurable but less consistent performance benefits.

\textbf{Low-Effectiveness Techniques} display substantial distributional overlap between ``with'' and ``without'' groups, statistical significance $<50\%$, and small effect sizes ($|r| < 0.1$), indicating limited practical utility.

\begin{table}[h]
\centering
\caption{\textbf{Technique Abbreviations Used in Figure~\ref{fig:technique_impact_analysis}.} Complete mapping of abbreviated technique names to their full descriptions, organized by methodological category.}
\label{tab:technique_abbreviations}
\begin{tabular}{ll}
\hline
\textbf{Abbreviation} & \textbf{Full Name} \\
\hline
\multicolumn{2}{l}{\textbf{Data Conditioning}} \\
Spline Interp & Spline Interpolation \\
Tikhonov Reg & Tikhonov Regularization \\
Kalman Smooth & Kalman-inspired Smoothing \\
Tukey & Tukey Windowing \\
S-G Filter & Savitzky-Golay Filtering \\
Adaptive Gain & Adaptive Gain Regularization \\
Median Smooth & Uniform/Median Smoothing \\
Gauss Smooth & Gaussian Smoothing \\
Median Baseline & Median-based Baseline Correction \\
SNR Reg & SNR-adaptive Regularization \\
Gauss Conv & Gaussian Convolution \\
\hline
\multicolumn{2}{l}{\textbf{Time-Frequency Analysis}} \\
Phase Coherence & Phase Coherence Analysis \\
CWT & Continuous Wavelet Transform \\
RMS Coherence & RMS Coherence Metric \\
Dual Align & Dual-channel Alignment \\
Spectral Entropy & Spectral Entropy \\
Freq Reg & Frequency-dependent Regularization \\
Log Compress & Logarithmic Compression \\
CWT Valid & CWT Validation \\
\hline
\multicolumn{2}{l}{\textbf{Trigger Detection}} \\
Curve Boost & Curvature Boosting \\
Curvature & Curvature Analysis \\
Sigmoid & Sigmoid Enhancement \\
MAD Threshold & MAD-based Robust Thresholding \\
MAD & Median Absolute Deviation \\
\hline
\end{tabular}
\end{table}

\textbf{Distributional Analysis Methodology.} 
\begin{itemize}
    \item Violin plots constructed using Gaussian kernel density estimation with adaptive bandwidth selection
    \item Performance metrics normalized to [0,1] scale enabling cross-technique comparison
    \item Color-coded categorical organization: data conditioning (blue), time-frequency analysis (orange), trigger detection (green)
    \item Statistical annotations include resampling-based significance percentages and rank-biserial effect sizes
    \item Median performance indicators highlight central tendency differences between technique groups
    \item Distributional separation quantified through overlap coefficients and Kolmogorov-Smirnov distances
    \item Technique abbreviations facilitate visual clarity while maintaining comprehensive coverage (Table~\ref{tab:technique_abbreviations})
\end{itemize}

This multi-dimensional effectiveness assessment framework enables systematic identification of high-impact techniques while distinguishing them from those with marginal or inconsistent benefits, providing clear guidance for algorithmic development priorities.

\section{Complete MCTS Tree Structure and Algorithmic Evolution}\label{sec:S6}

\begin{figure*}
\centering
\includegraphics[width=\textwidth]{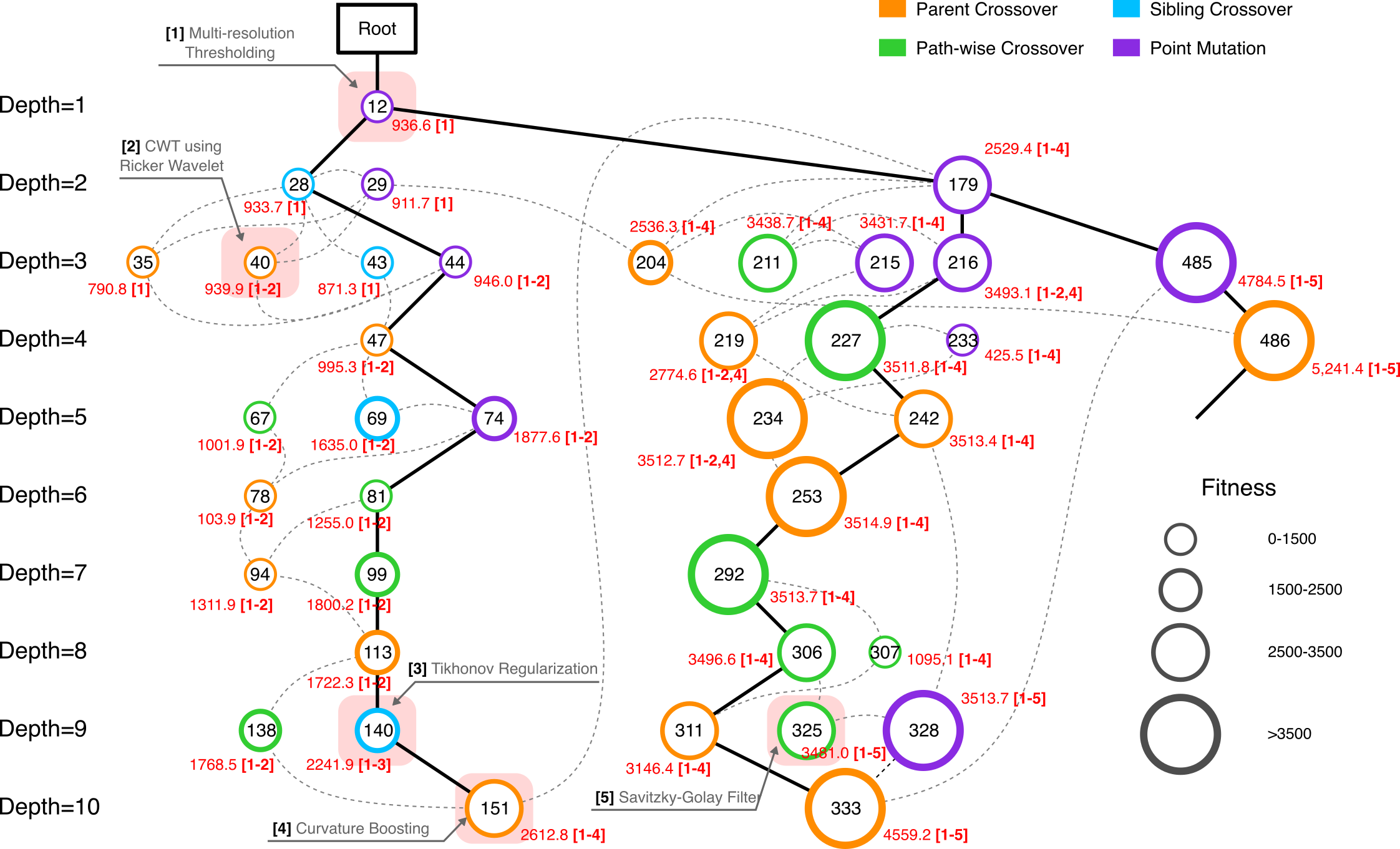}
\caption{\textbf{Complete MCTS search tree with node fitness values and technique compositions.} Each node shows its fitness score (red annotations) and constituent algorithmic techniques organized by category [1-5]. Node size reflects fitness magnitude. The tree demonstrates systematic technique evolution and cross-branch knowledge transfer, with optimal performance achieved through multi-technique integration at terminal nodes.}
\label{fig:sup5}
\end{figure*}

Figure~\ref{fig:sup5} presents the complete MCTS search tree evolution with node-by-node fitness values and technique compositions. Each node displays its fitness score (marked in red) alongside the specific algorithmic techniques discovered at that search depth. The five core technique categories [1-5] correspond to: 

\begin{itemize}
    \item [1] Multi-resolution Thresholding
    \item [2] Continuous Wavelet Transform (CWT) using Ricker Wavelet
    \item [3] Tikhonov Regularization
    \item [4] Curvature Boosting
    \item [5] Savitzky-Golay Filter
\end{itemize}

These techniques demonstrate the systematic evolution of algorithmic complexity through the MCTS exploration process, with detailed implementation examples provided in Section~\ref{sec:S1_case_study}.

\end{appendices}


\bibliography{sn-bibliography}


\end{document}